\let\prismorigsloppy\sloppy
\def\sloppy{%
    \tolerance 9999%
    \emergencystretch 3em%
    \hfuzz 1000pt%
    \vfuzz\hfuzz}
\documentclass[research]{fcs}
\let\sloppy\prismorigsloppy

\usepackage{amsmath,amsfonts,bm}

\def\eqref#1{equation~\ref{#1}}

\def\1{\bm{1}}

\DeclareMathAlphabet{\mathsfit}{\encodingdefault}{\sfdefault}{m}{sl}
\SetMathAlphabet{\mathsfit}{bold}{\encodingdefault}{\sfdefault}{bx}{n}

\usepackage[utf8]{inputenc}
\usepackage[T1]{fontenc}
\usepackage{url}
\usepackage{booktabs}
\usepackage{amsfonts}
\usepackage{nicefrac}
\usepackage{microtype}
\usepackage{placeins}

\usepackage[table, svgnames, dvipsnames]{xcolor}
\usepackage{makecell, cellspace, caption}
\usepackage{subcaption}
\usepackage{wrapfig}
\usepackage{graphicx}
\usepackage{cancel}

\usepackage{algorithm}
\usepackage{algorithmic}

\usepackage{amsmath}
\usepackage{amssymb}
\usepackage{mathtools}
\usepackage{amsthm}
\usepackage{bm}
\usepackage{multirow}
\usepackage{stmaryrd}
\usepackage{bbm}
\usepackage{longtable}
\makeatletter
\def\Hy@Warning#1{}
\makeatother

\theoremstyle{plain}
\theoremstyle{definition}

\theoremstyle{remark}

\definecolor{peach_red}{HTML}{BD4146}
\definecolor{deep_burgundy}{HTML}{551F33}
\definecolor{light_purple}{HTML}{B192B4}
\definecolor{light_coral_pink}{HTML}{F9C8C4}

\title{EchoAlign: Bridging Generative and Discriminative Learning under Noisy Labels}

\author[1]{Yuxiang Zheng}
\author*[2]{Zhongyi Han}
\author[2]{Yilong Yin}

\address[2]{School of Software, Shandong University, Jinan 250100, China}
\address[1]{Sydney AI Centre, The University of Sydney, Sydney, NSW 2050, Australia}

\fcssetup{
    received       = {October 03, 2025},
    accepted       = {March 17, 2026},
    corr-email     = {zhongyi.han@sdu.edu.cn},
}

\begin{abstract}
    Noisy labels severely hinder the accuracy and generalization of machine learning models, especially when caused by ambiguous instance features that complicate reliable annotation. Existing approaches, such as transition-matrix-based label correction, struggle to capture complex relationships between instances and noisy labels, limiting their effectiveness in such scenarios. We present EchoAlign, a framework that bridges generative and discriminative learning under noisy labels. Instead of correcting labels, EchoAlign treats noisy labels ($\tilde{Y}$) as accurate and modifies corresponding instances ($X$) to align with them. The framework integrates two components: (1) EchoMod employs controllable generative models to adjust instance features while preserving key instance-level structural cues (e.g., shape and edges) and avoiding excessive distortion; and (2) EchoSelect addresses distribution shifts by retaining a reliable subset of original instances, guided by feature similarity between original and modified samples. This generative-discriminative interplay enables robust learning even in highly noisy settings. Experiments on three benchmark datasets show that EchoAlign outperforms state-of-the-art methods in most evaluated settings. Under 30\% instance-dependent noise, EchoSelect retains nearly twice as many correctly labeled samples as competing approaches while maintaining 99\% selection accuracy, highlighting the robustness and effectiveness of EchoAlign.
\end{abstract}

\begin{document}

    \section{Introduction}
    \label{Introduction}

    The rapid advancement of neural networks has underscored the significance of learning from noisy labels (LNL)~\cite{Tan2019EfficientNet, Dosovitskiy2021AnImageWorth, Stiennon20HumanFeedback, chen2023understanding}. Although web crawling and crowdsourcing provide cost-effective means for collecting large datasets, they often introduce noisy labels that hinder model generalization~\cite{Yu2018WebCrawling, li2017webvision, Welinder2010crowdsourcing, zhang2017improving, natarajan2013learning, gu2023instance}. Recent studies have highlighted that label noise in pretraining data adversely affects out-of-distribution generalization in downstream tasks for foundation models \cite{chen2023understanding,chen2024catastrophic}. Noisy labels are generally categorized as random, class-dependent, or instance-dependent, with the latter two posing particular challenges due to ambiguous instance features, making it difficult to distinguish mislabeled examples from true class instances~\cite{menon2018learning,xia2020part,yao2023better,bai2023subclass}.

    Prior research has primarily approached LNL through either noise-modeling-free or noise-modeling frameworks. Noise-modeling-free techniques, such as filtering out high-loss examples~\cite{han2018co,yu2019does,wang2019co}, are limited to selecting clean samples and do not address the potential for correcting incorrect labels, thereby discarding valuable supervisory information. In contrast, noise-modeling approaches explicitly consider the label-noise generation process~\cite{scott2013classification,scott2015rate,goldberger2016training}, often employing a transition matrix to relate noisy labels to their clean counterparts~\cite{Berthon2021ConfidenceScores}. Theoretically, an optimal classifier can be trained with sufficient noisy data and an accurate transition matrix~\cite{reed2014training,liu2015classification}. However, estimating this matrix is inherently ill-posed due to uncertainty and variability in noisy data~\cite{xia2019anchor,cheng2020learning}. Moreover, these models often rely on additional assumptions, such as the exact nature of the noise, which are challenging to validate and may not hold in real-world datasets, leading to suboptimal performance~\cite{xia2020part, yao2023causality, liu2023identifiability}. Traditional label correction methods are particularly limited when dealing with label noise caused by ambiguous features. For example, in datasets collected through web crawling, an image labeled as ``dog'' might actually depict a cartoon or a product featuring a dog, which makes label correction challenging and often impractical.

    \begin{figure}[ht]
        \centering
        \includegraphics[width=0.45\textwidth]{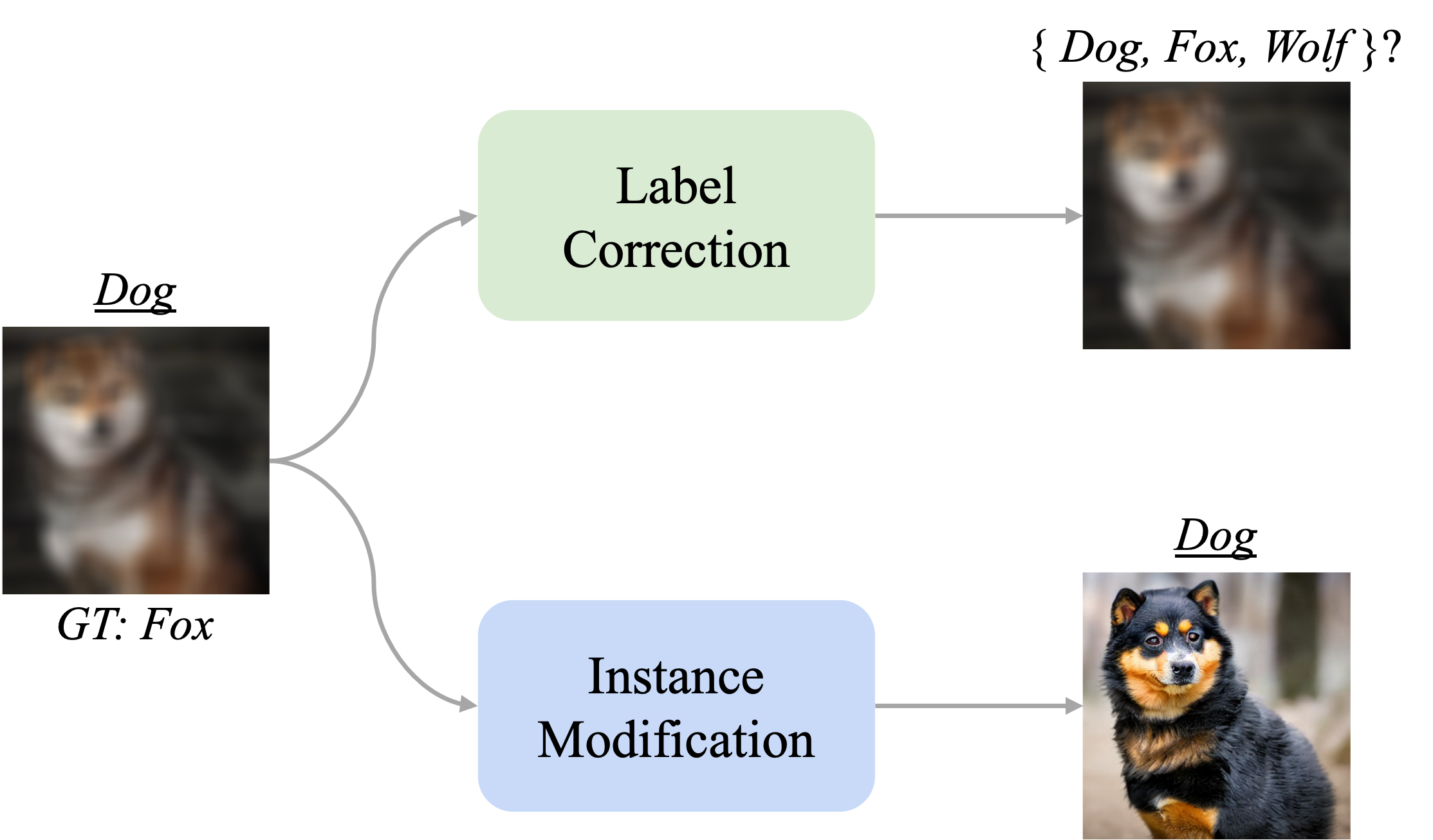}
        \caption{Instance modification effectively aligns instances with their labels, while label correction struggles with ambiguous cases.}
        \label{fig1}
    \end{figure}

    In this paper, we introduce a novel perspective on handling label noise by employing \emph{instance modification} rather than correcting labels. Instead of attempting to correct noisy labels, we adjust instances to better align with their labels, even if those labels are incorrect. This innovative approach, illustrated in Figure~\ref{fig1}, directly addresses the root cause of label noise. Leveraging causal learning principles~\cite{neuberg2003causality,peters2017elements,yao2021instance}, we model instance-dependent label noise from a causal perspective, as depicted in the causal graph of Figure~\ref{causal_graph}. Specifically, we consider how different factors, such as instance characteristics and latent variables, contribute to the generation of noisy labels, enabling us to better understand and address the root causes of label noise. In crowdsourcing scenarios, for instance, ambiguous or blurred instances are more prone to labeling errors. Instead of attempting to infer the `true' label, modifying the instance to make it more distinguishable can be more effective. For example, in medical imaging, if a tumor is labeled as malignant but its visual features are too subtle, enhancing the image to highlight relevant features can assist both models and humans in identifying it more accurately~\cite{covid2022}. Similarly, in sentiment analysis, modifying ambiguous sentences to be more explicit can better align them with their intended sentiment labels, thereby reducing ambiguity and improving classification accuracy.

    \begin{figure*}[ht]
        \centering
        \begin{subfigure}[b]{0.8\textwidth}
            \centering
            \includegraphics[width=\linewidth]{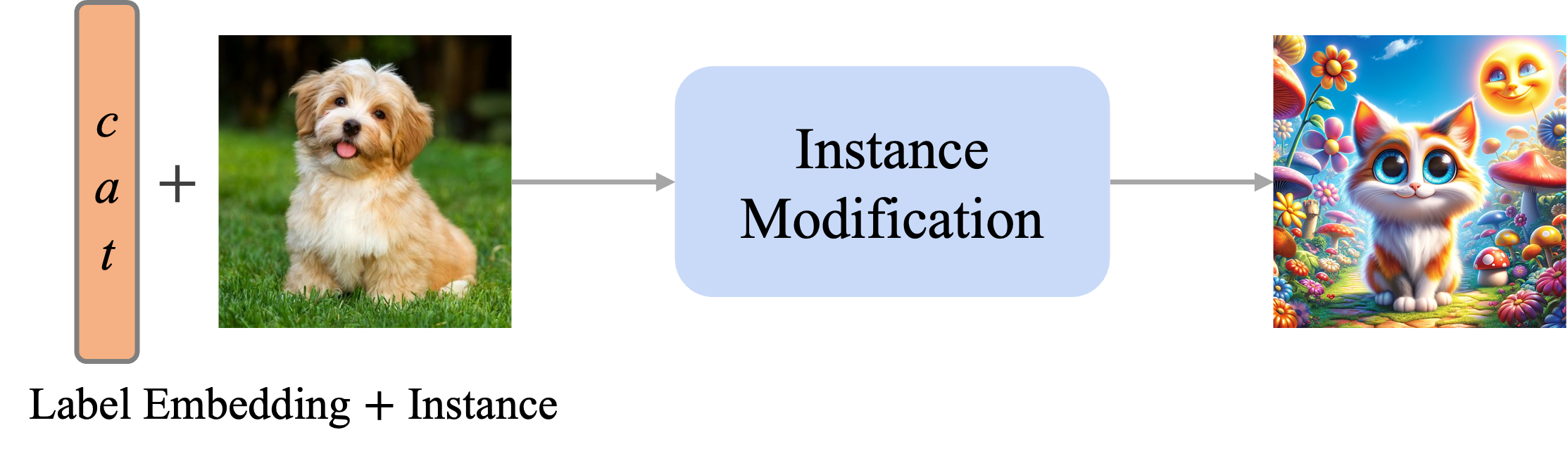}
            \caption{Characteristic Shift.}
            \label{shift}
        \end{subfigure}

        \vspace{0.5cm}

        \begin{subfigure}{0.48\textwidth}
            \centering
            \includegraphics[width=\linewidth]{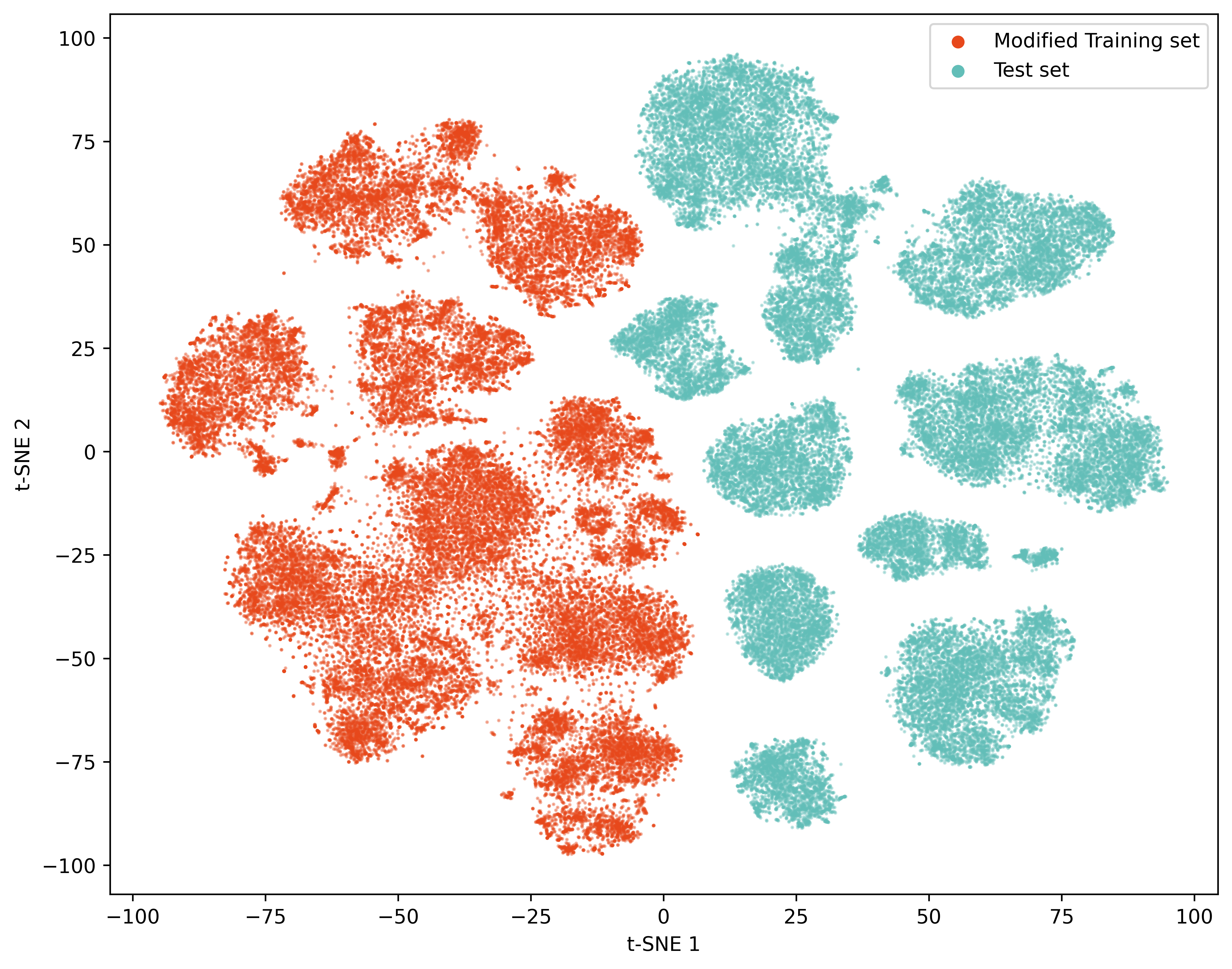}
            \caption{T-SNE visualization of CIFAR-10 instance representations using \(X\).}
            \label{distribution_before}
        \end{subfigure}\hfill
        \begin{subfigure}{0.48\textwidth}
            \centering
            \includegraphics[width=\linewidth]{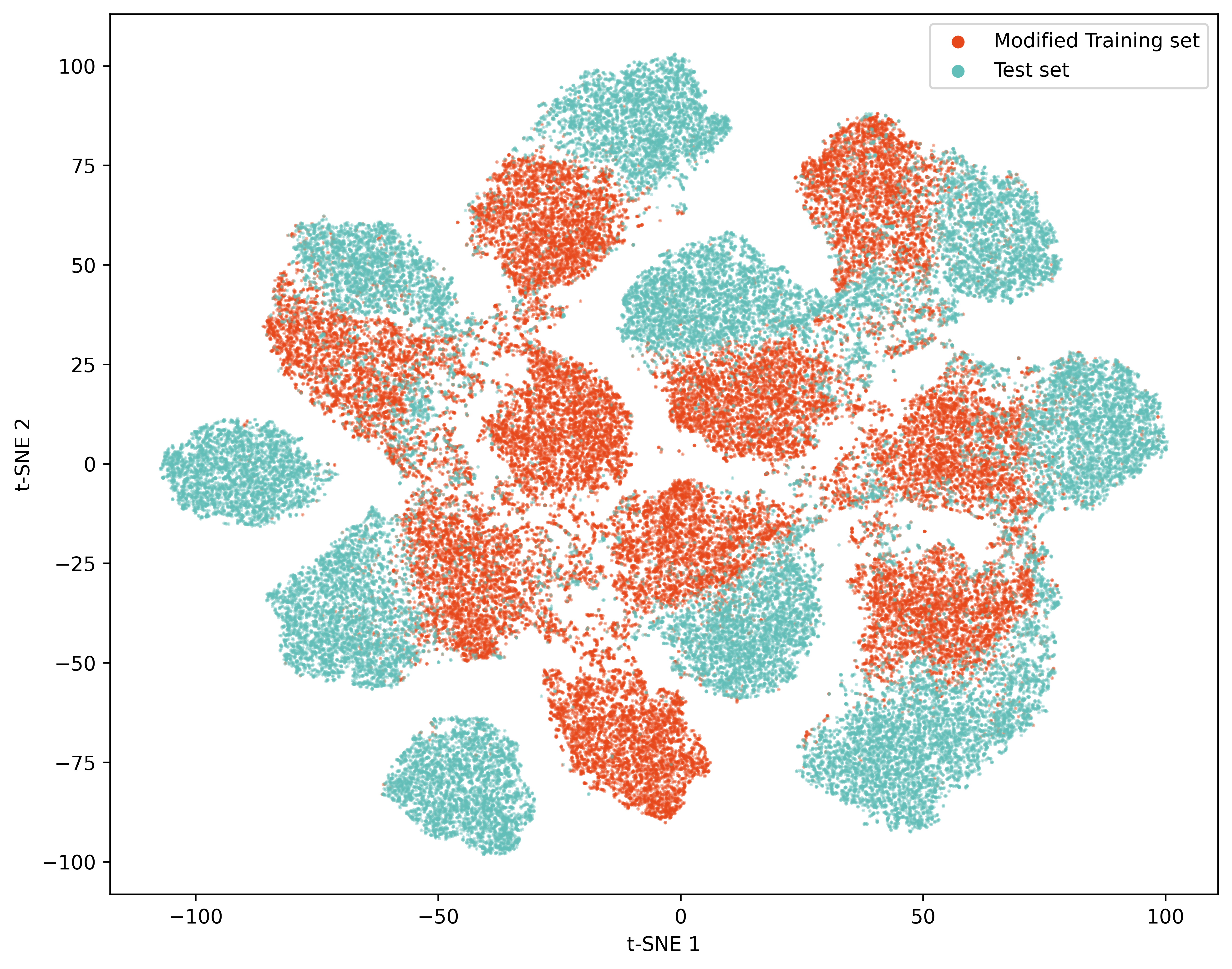}
            \caption{T-SNE visualization of CIFAR-10 instance representations using \(X'\)}
            \label{distribution_after}
        \end{subfigure}

        \caption{(Top) Main challenge 1: Characteristic Shift. (Bottom) Main challenge 2: Distribution Shift.}
    \end{figure*}

    However, instance modification presents challenges at both the instance and dataset levels. At the instance level, a key challenge is modifying instances while preserving their essential characteristics, such as shape, texture, or color patterns. These features are crucial for distinguishing related categories. Excessive alterations can distort these defining features, leading to a phenomenon we refer to as the characteristic shift (Figure~\ref{shift}). For example, when a wolf is mislabeled as a dog, transforming the wolf image into a typical dog might eliminate important shared features such as body shape and texture, which are critical for distinguishing between wolves and dogs. At the dataset level, instance modification may introduce a distribution shift (Figures~\ref{distribution_before} and \ref{distribution_after}), where the statistical distribution of modified instances deviates from that of the original instances, potentially affecting model generalization to real-world data~\cite{han2022towards,han2022learning}. Empirical observations, as visualized through T-SNE projections, reveal that modified instances may occupy a distinct feature space from their unaltered counterparts, altering the training dynamics. Addressing these shifts requires a balanced framework that preserves essential characteristics while effectively managing distribution differences between original and modified instances to ensure robustness in real-world applications.

    To address these challenges, we propose a simple yet effective framework, EchoAlign (\S~\ref{method}). EchoAlign consists of two key components: EchoMod and EchoSelect. EchoMod modifies instances using controllable generative models, ensuring alignment with noisy labels while preserving intrinsic characteristics. EchoSelect mitigates covariate shifts by selecting original instances with correct labels, maintaining a balanced distribution between original and modified data. This selection is guided by a novel insight: after instance modification, the cosine feature similarity between original and modified images reveals distinctions between correctly and incorrectly labeled samples. EchoSelect uses this similarity metric to curate a reliable training set, improving robustness and accuracy in both supervised and self-supervised training.

    Our key contributions and findings are summarized as follows:
    \begin{enumerate}
        \item We introduce a transformative shift in addressing label noise by modifying instances to align with noisy labels instead of correcting them, supported by theoretical analysis  (\S~\ref{Analysis});
        \item  We present EchoAlign, a framework featuring EchoMod for controlled instance modification and EchoSelect for strategic sample selection (\S~\ref{method});
        \item We demonstrate consistent improvements of instance modification with EchoAlign on \allowbreak CIFAR-10/100 under three common synthetic noise types and on CIFAR-10N with real human annotations, outperforming strong baselines under the evaluated settings (\S~\ref{experiments}).
    \end{enumerate}

    \section{Related Work}
    \textbf{Learning with Noisy Labels} \quad
    Research in this domain has predominantly followed two paths:
    \textbf{(1)} Noise-modeling-free methods: These methods primarily rely on the memorization effects observed in deep neural networks (DNNs), which tend to learn simpler (clean) examples before memorizing more complex (noisy) ones \cite{Arpit2017Look,Wu2020TopoFilter,Kim2021FINE}. Techniques include early stopping \cite{han2018co, Nguyen2020SELF, Liu2020ELR, xia2021CDR, Lu2022NoiseAttention, Bai2021PES}, pseudo-labeling \cite{tanaka2018joint}, and leveraging Gaussian Mixture Models in a semi-supervised learning context \cite{Li2020DivideMix}.
    \textbf{(2)} Noise-modeling methods: These approaches focus on estimating a noise transition matrix, modeling how clean labels can become corrupted into noisy observations. However, accurately modeling this noise process is particularly challenging when relying solely on noisy data \cite{xia2019anchor,cheng2020learning}. Many existing studies depend on assumptions that may not hold in real-world datasets \cite{xia2020part, yao2023causality, liu2023identifiability}. Consequently, these methods often struggle to effectively handle structured noise patterns, such as subclass-dominant label noise \cite{bai2023subclass}.

    \textbf{Generative Models} \quad
    Recent advances in generative models, including variational auto-encoders, generative adversarial networks, and diffusion models, have transformed applications with their exceptional sample generation capabilities \cite{Du2023SequentialRW,Wang2023ConditionalDD,Franceschi2023UnifyingGA}. Diffusion models, known for their superior output control, are particularly effective at denoising signals \cite{zhang2023adding,kingma2021variational}. While these models hold promise for noisy label scenarios, existing approaches like Dynamics-Enhanced Generative Models (DyGen) \cite{Zhuang2023DyGenLF} and Label-Retrieval-Augmented Diffusion Models \cite{Chen2023LabelRetrievalAugmentedDM} still focus primarily on enhancing predictions or retrieving latent clean labels.
    Our work takes a fundamentally different approach. We leverage controllable generative models, treating noisy labels as correct and aligning instances with these labels, thus bypassing the challenges of traditional noise modeling and focusing on improving the quality of training data. Controllable generative models, such as ControlNet \cite{zhang2023adding} and iPromptDiff \cite{chen2023improving}, enable precise control over the generated outputs. Unlike traditional generative models which generate images from random noise, controllable generative models use control information (\emph{e.g.}, text descriptions, class labels, or reference images) as input \cite{bose2022controllable}, guiding the generation process to ensure that the outputs align with the desired characteristics.

    \section{Analysis}
    \label{Analysis}

    \textbf{Problem Definition}
    In addressing the challenges posed by learning from noisy labels (LNL), we formally define the problem and introduce the concept of instance modification within a mathematical framework. Let $\mathcal{X}$ represent the input space of instances and $\mathcal{Y}$ the space of labels. In the traditional LNL setting, each instance $X \in \mathcal{X}$ is associated with a noisy label $\tilde{Y} \in \mathcal{Y}$, which may differ from the true label $Y \in \mathcal{Y}$. The goal is to learn a mapping $f: \mathcal{X} \rightarrow \mathcal{Y}$ that predicts the true label $Y$ as accurately as possible, despite the presence of noisy labels. Instance modification diverges from the conventional approach of directly correcting noisy labels $\tilde{Y}$ to match the true labels $Y$. Instead, we propose adjusting each instance $X$ to better align with its given noisy label $\tilde{Y}$. Mathematically, this involves transforming each instance $X$ into a modified instance $X^{\prime}$, such that $f(X^{\prime})$ aligns more closely with $\tilde{Y}$, leveraging the inherent information contained within the noisy label itself.

\begin{figure}[ht]
        \centering
        \includegraphics[width=0.4\textwidth]{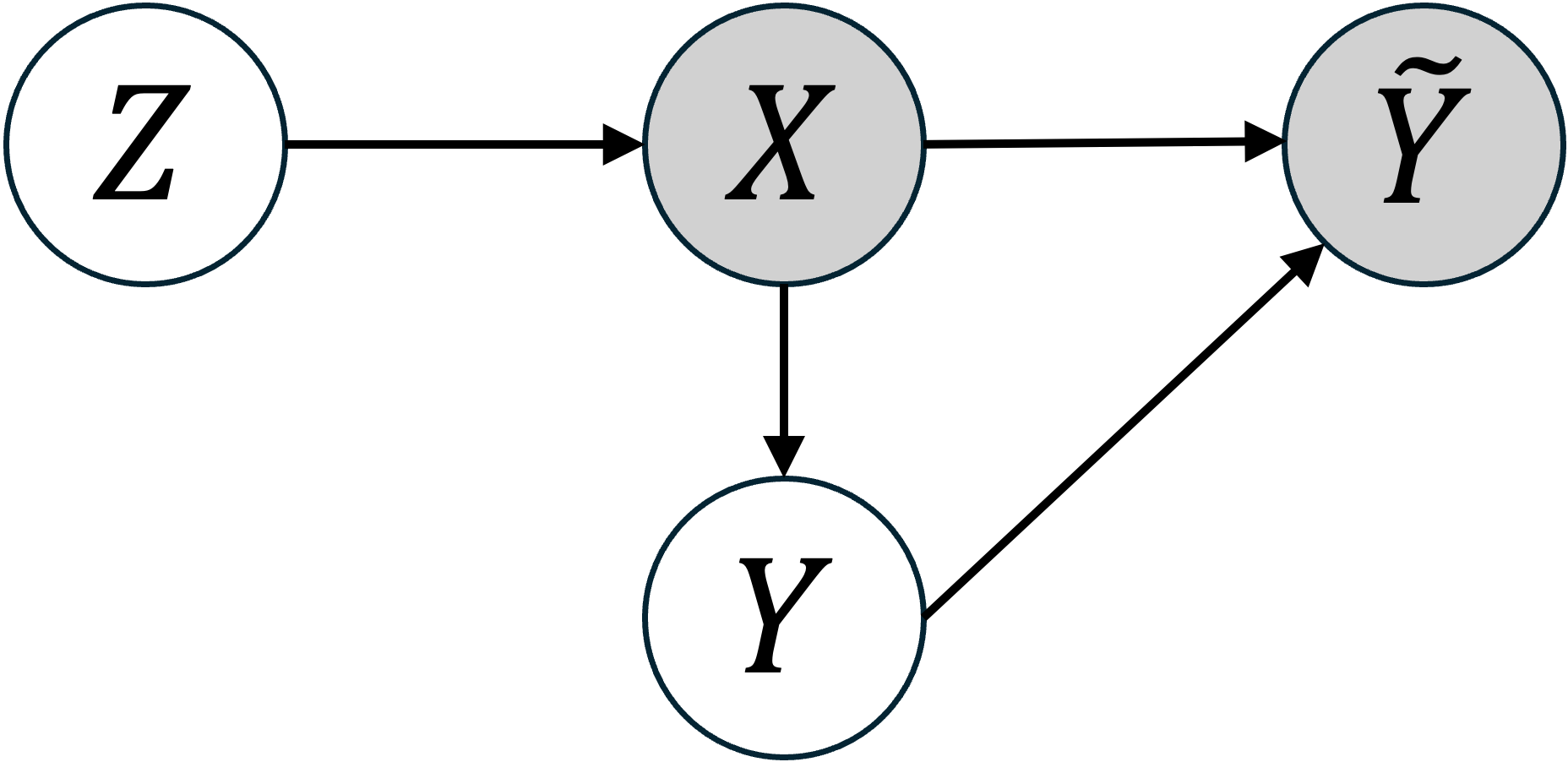}
        \caption{A graphical causal model, revealing a data generative process with instance-dependent label noise.}
        \label{causal_graph}
    \end{figure}

    \textbf{Theoretical Analysis} \quad
    According to the causal learning framework~\cite{liu2023identifiability,yao2021instance}, the noise can often be represented as a function of both the instance features and external factors, encapsulated by latent variables $Z$. We assume that the causal relations (commonly occurring in crowdsourcing scenarios) are represented by the causal graph as illustrated in Figure~\ref{causal_graph}, where $Z$ represents latent variables that affect the instance $X$ and the noisy label $\tilde{Y}$, inducing instance-dependent label noise. Instance modification aims to transform $X$ into $X'$ such that the modified instance $X^{\prime}$ better aligns with $\tilde{Y}$ under the assumption that $\tilde{Y}$ contains partial information about the true label $Y$. Accordingly, we can deduce the effectiveness of instance modification as follows.

    \begin{theorem}[Effectiveness of Instance Modification]
        Assume that the noisy labels are generated by a stochastic process influenced by latent variables $Z$, where $\tilde{Y} = h(Y, Z)$ and $Y$ are the true labels. Let $T$ be a transformation such that $X' = T(X, \tilde{Y}; \theta)$, where $\theta$ is chosen to optimize the alignment of $X'$ with $ \tilde{Y}$. Then, under this transformation, the predictive performance of a model trained on $(X', \tilde{Y})$ is no worse, and can be improved compared to training on $(X,\tilde{Y})$, in terms of:
        \begin{enumerate}
            \item \textbf{Alignment}: The mutual information between $X'$ and $\tilde{Y}$, $ I(X'; \tilde{Y}) $, is no smaller than $ I(X; \tilde{Y}) $, indicating better alignment of modified instances with their noisy labels.
            \item \textbf{Error Reduction}: Under squared loss, let \(f_{X'}^{*}(x')=\mathbb{E}[Y\mid X'=x']\) denote the population-optimal predictor on the modified instances. If the modification does not remove predictive information about \(Y\) and the induced distribution shift is controlled, then the achievable expected prediction error using \(X'\), \( \mathbb{E}_{X', Y}[(Y - f_{X'}^{*}(X'))^2] \), is no larger than that using \(X\).
            \item \textbf{Estimation Stability}: In a linear regression illustration, if the modified design is no worse conditioned in the relevant subspace (e.g., \(\lambda_{\min}(X'^T X') \ge \lambda_{\min}(X^T X)\)) and the modification is non-expansive in norm (e.g., \(\|x'\|_2 \le \|x\|_2\)), then the predictive variance using \(X'\) is reduced compared to using \(X\), leading to more stable predictions.
            \item \textbf{Generalization}: Modifications in \(X'\) can lead to better generalization. In particular, under a non-expansive modification (e.g., \(\|x_i'\|_2 \le \|x_i\|_2\)), the complexity of common hypothesis classes on \(X'\) is no larger than that on \(X\), which yields a no-worse (and potentially tighter) generalization bound.
        \end{enumerate}
        These statements follow under the assumptions on the noise model $h$ and the transformation $T$ in our causal formulation. \textbf{More details and proofs can be found in the Appendix A.}
        \label{theorem1}
    \end{theorem}

    Theorem~\ref{theorem1} provides a stylized sufficient-condition analysis that motivates why instance modification can be beneficial. It suggests improved label alignment, no larger Bayes risk under squared loss, more stable estimation in a linear illustration, and controlled generalization under the stated assumptions. These insights provide several key motivations for the design of our method. First, the improvements in alignment highlight the importance of modifying instances to embed noisy label information directly. This motivates the use of controllable generative models in EchoAlign, which can effectively incorporate label information into the instance features. Second, ensuring a minimal distribution difference between $X$ and $X'$ is crucial. EchoMod generates $X'$ with small distribution differences from $X$, while EchoSelect retains clean samples to control distribution differences, ensuring better generalization on test data. Third, the improvement in estimation stability indicates that using modified features can result in more consistent and reliable model predictions, motivating a focus on preserving the essential characteristics of the data during transformation to reduce variability and enhance both statistical stability and robustness in model performance.

\begin{figure}[ht]
        \centering
        \includegraphics[width=0.45\textwidth]{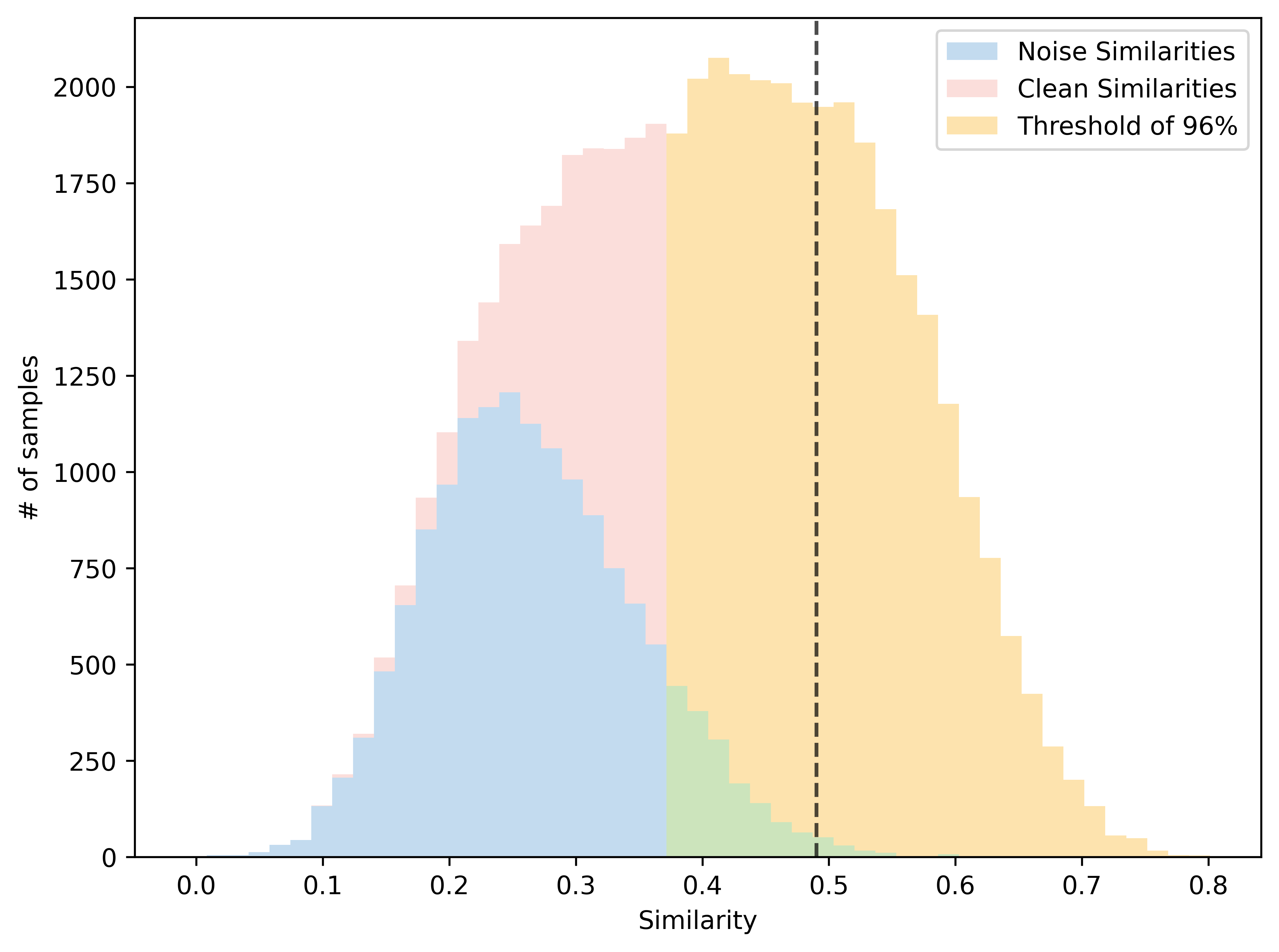}
        \caption{The feature similarity between the original and modified instances is a valuable metric for sample selection after instance modification.}
        \label{feature_similarity}
    \end{figure}

    \textbf{Analyzing Feature Similarity Distributions} \quad
    In this study, we address the challenges of instance modification, which can induce distribution shifts between the training and test sets. Preserving clean original instances is crucial to mitigating these shifts. Existing sample selection methods (\emph{e.g.}, small loss \cite{han2018co}) often falter under complex label noise conditions, such as instance-dependent noise, necessitating a more precise selection strategy. To this end, we find an interesting phenomenon: Clean samples generally exhibit higher similarity between features of original and modified images, indicating minimal semantic and label changes after modification, whereas samples suspected to be noisy tend to yield lower similarity, which is consistent with larger semantic and label adjustments. Utilizing the feature similarity distributions between original and modified instances emerges as a robust tool for enhancing sample selection accuracy. These distinctions are visually represented in Figure~\ref{feature_similarity}.
    The similarity is computed using the CLIP ViT-B-32 feature extractor~\cite{radford2021learning} on the CIFAR-10 dataset with 30\% instance-dependent noise. We use ControlNet~\cite{zhang2023adding} to modify instances. The black dashed line indicates the sample threshold achievable by the previous best method at 96\% accuracy~\cite{Yang2022bayeslabel}. In contrast, EchoSelect, at 96\% accuracy, can retain the samples in the yellow section. In environments with 30\% instance-dependent noise, EchoSelect retains nearly twice as many samples at 99\% accuracy. Statistical validation using the Kolmogorov-Smirnov test confirmed significant differences in the distributions (p-value < 0.001), demonstrating the utility of feature similarity as a robust metric for identifying clean samples within noisy datasets.

    \begin{figure}[ht]
        \centering
        \begin{subfigure}{0.44\textwidth}
            \centering
            \includegraphics[width=\linewidth]{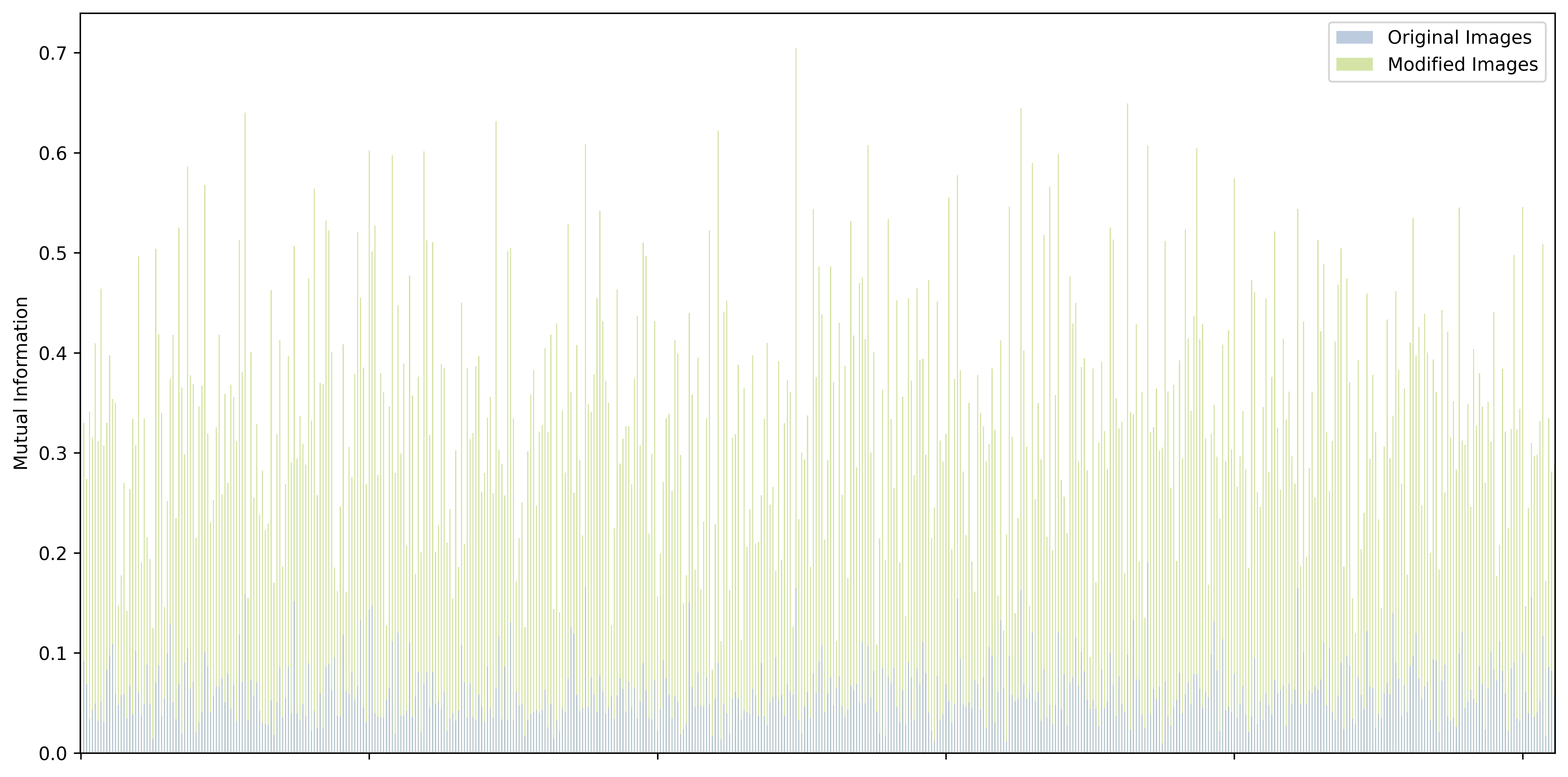}
            \caption{}
            \label{theorem:f6a}
        \end{subfigure}
        \hfill
        \begin{subfigure}{0.48\textwidth}
            \centering
            \includegraphics[width=\linewidth]{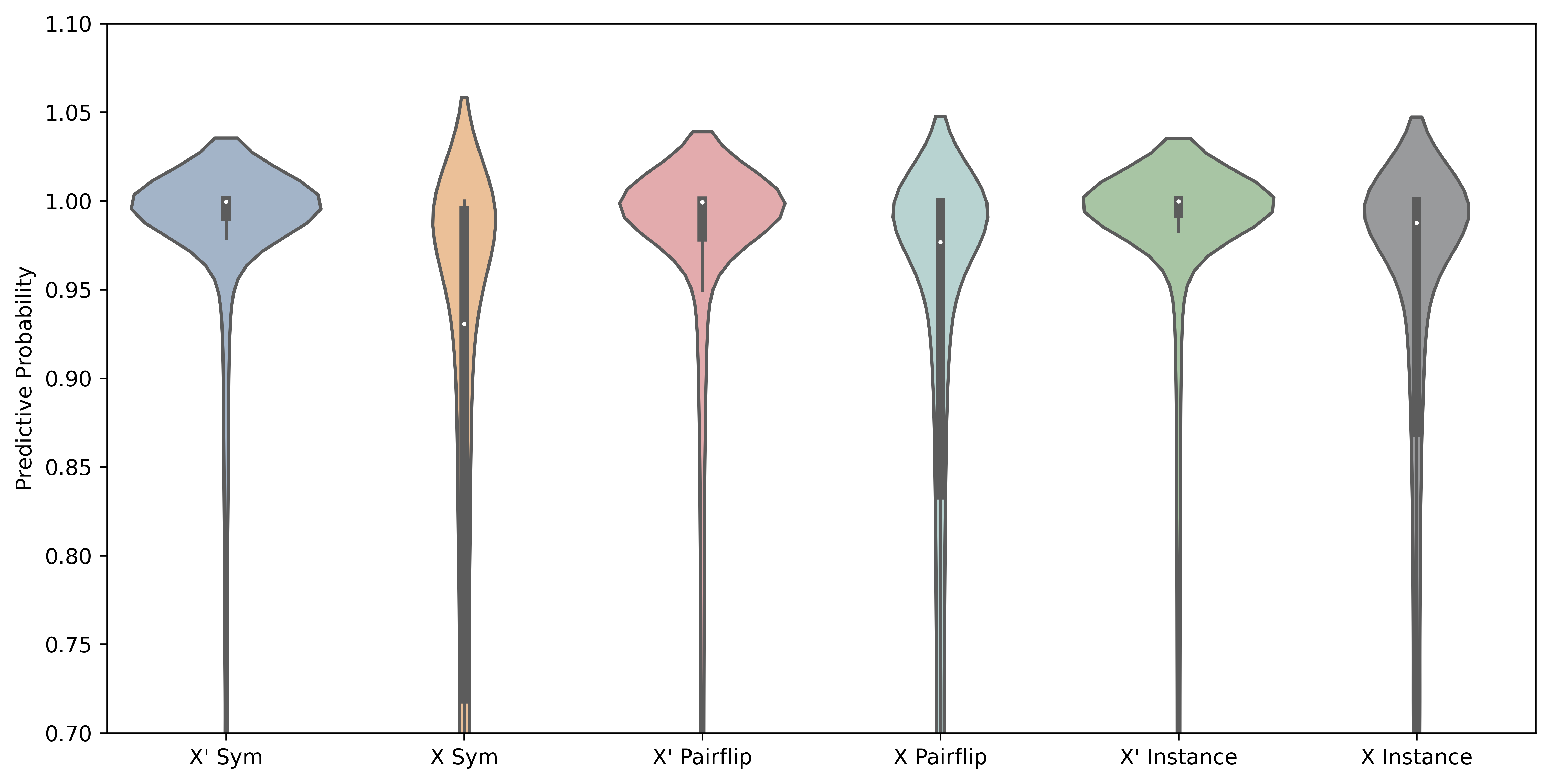}
            \caption{}
            \label{theorem:f6b}
        \end{subfigure}
        \caption{(a) illustrates the mutual information between the labels of 50,000 original samples and their corresponding 50,000 modified samples under 50\% instance-dependent noise on CIFAR-10. (b) shows the distribution of the predictive probability of the estimator \(f\) using \(X'\) and \(X\).}
    \end{figure}

\begin{figure*}[ht]
        \centering
        \begin{subfigure}{0.32\textwidth}
            \centering
            \includegraphics[width=\linewidth]{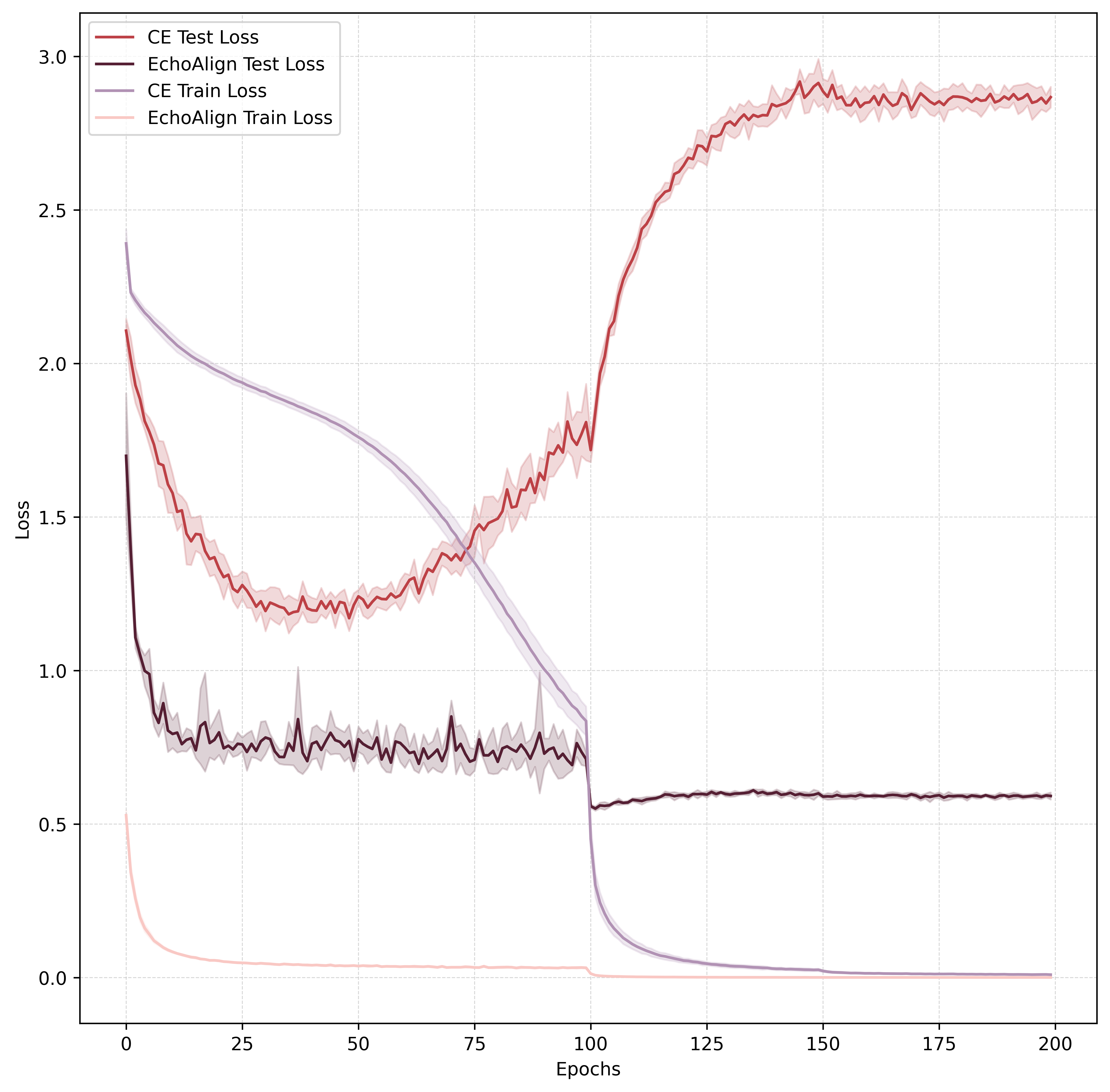}
            \caption{50\% symmetric noise}
        \end{subfigure}\hfill
        \begin{subfigure}{0.32\textwidth}
            \centering
            \includegraphics[width=\linewidth]{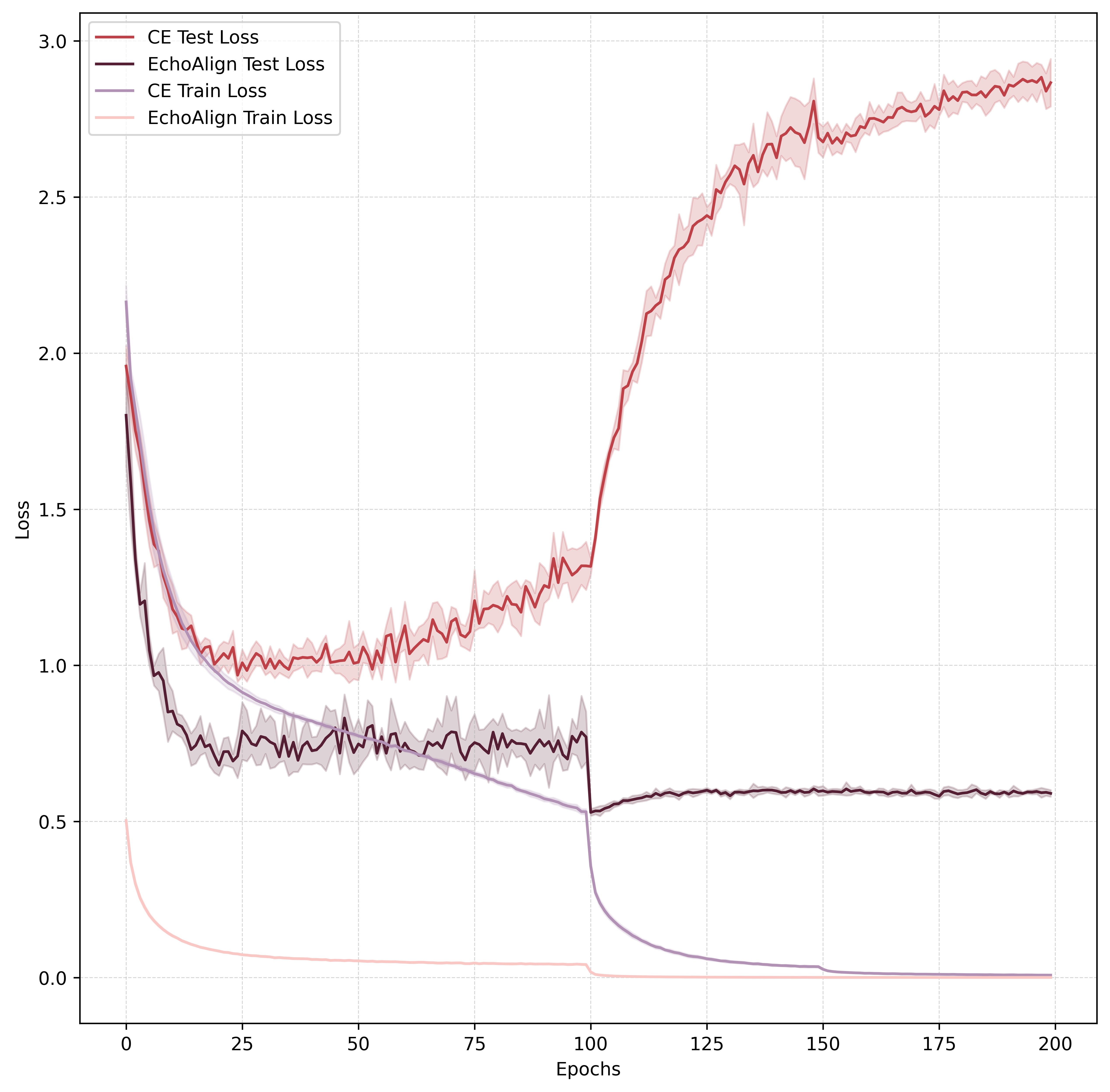}
            \caption{45\% pair-flip noise}
        \end{subfigure}\hfill
        \begin{subfigure}{0.32\textwidth}
            \centering
            \includegraphics[width=\linewidth]{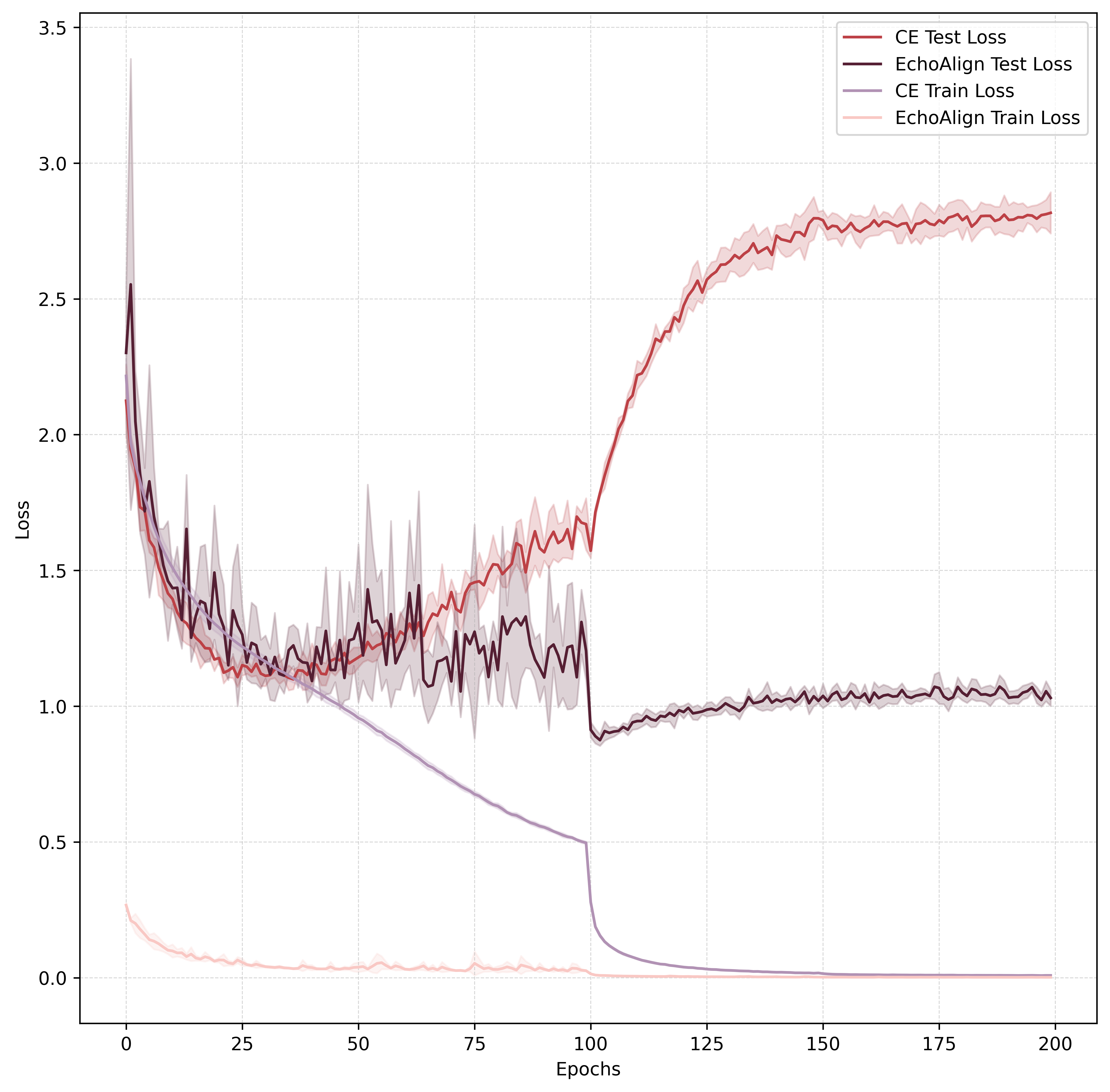}
            \caption{50\% instance-dependent noise}
        \end{subfigure}\hfill
        \caption{Figures (a), (b), and (c) respectively illustrate the differences in training and testing losses between EchoAlign and the CE model under 50\% symmetric noise, 45\% pair-flip noise, and 50\% instance-dependent noise conditions on the CIFAR-10. The \textcolor{peach_red}{bright peach red} and \textcolor{deep_burgundy}{deep burgundy} lines represent the performance of CE and EchoAlign on the test set, respectively, while the \textcolor{light_purple}{light purple} and \textcolor{light_coral_pink}{light coral pink} lines denote their performance on the training set.}
        \label{loss}
    \end{figure*}

    \textbf{Empirical Validation of Theoretical Analysis} \quad
    To empirically illustrate the behavior predicted by Theorem~\ref{theorem1}, we undertook specific experiments to demonstrate its efficacy. The theorem posits that by applying an appropriate transformation \(T\), the alignment between the instances \(X\) and the noisy labels \(\tilde{Y}\) can be optimized, thereby increasing their mutual information. On the CIFAR-10 dataset, we calculated the mutual information between 50,000 images and their labels. As observed in Figure~\ref{theorem:f6a}, the mutual information \(I(X' ; \tilde{Y})\) between the transformed instances \(X'\) and the noisy label \(\tilde{Y}\) is significantly higher than the mutual information \(I(X ; \tilde{Y})\) between the original instances \(X\) and \(\tilde{Y}\). Figure~\ref{theorem:f6b} also supports the third point of our theorem, i.e., the estimator trained on X' has lower variance than the one trained on X, which illustrates the higher stability and robustness of our method. Furthermore, concerning prediction error, Figure~\ref{loss} displays the training and testing results under different noise types. The results show that using modified samples results in significantly lower errors, both in the training and testing sets.

    \section{EchoAlign}
    \label{method}

    The EchoAlign framework tackles the challenge of noisy labels in supervised learning. It comprises two primary components: \textbf{(1)} EchoMod modifies instances using controllable generative models, ensuring alignment with noisy labels while preserving intrinsic characteristics. \textbf{(2)} EchoSelect selects original instances with correct labels, maintaining a balanced distribution between original and modified data.

    \begin{figure*}[t]
        \centering
        \includegraphics[width=0.88\linewidth]{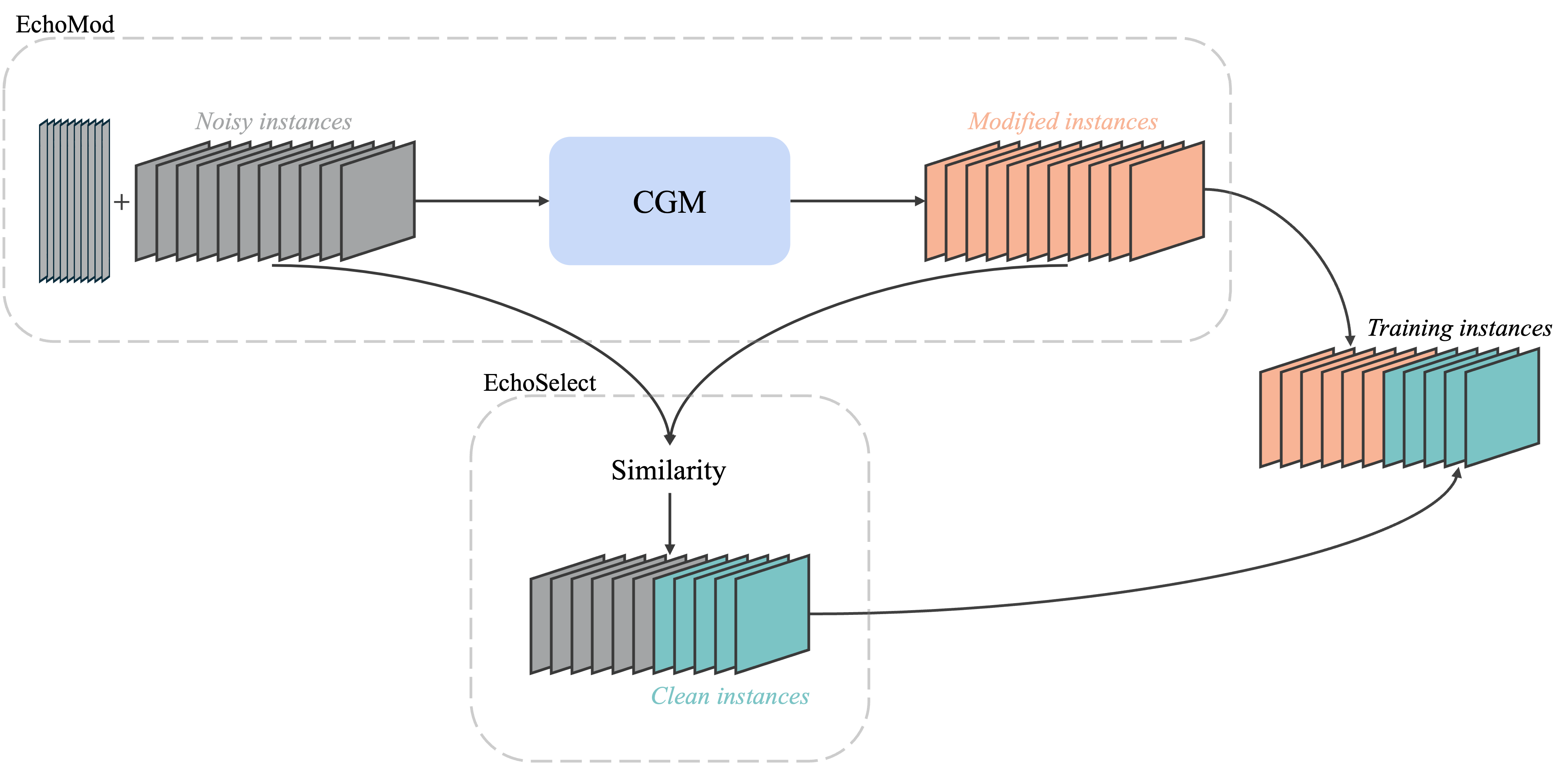}
        \caption{The framework of EchoAlign.}
        \label{framework}
        \vspace{-2mm}
    \end{figure*}

    Figure~\ref{framework} illustrates the framework of EchoAlign. Specifically, EchoMod utilizes controllable generative models (CGMs) to perform instance modifications based on noisy labels, while EchoSelect applies feature similarity evaluation to strategically filter instances, preserving original instances that are likely clean and adopting modified instances to ensure label alignment. The integration of these modules effectively addresses the characteristic and distribution shifts introduced by label noise.

    \subsection{EchoMod: Instance Modification}

    \textbf{Motivation} \quad
    When labels are noisy, they do not reflect the true characteristics of the corresponding data instances. This discrepancy hinders a model's ability to learn meaningful patterns. EchoMod addresses this by transforming data instances to be consistent with their noisy labels. This controlled modification helps the model extract relevant information even when labels contain noise.

    \textbf{Mechanism} \quad
    EchoMod leverages a pre-trained controllable generative model (\emph{e.g.}, a controllable diffusion-based model) to modify data instances. The primary goal is to enhance the alignment between instances and their potentially noisy labels. This alignment is achieved by carefully guiding the generative model's process. First, the controllable generative model has undergone prior training on a large dataset. This pre-training has equipped the model with a deep understanding of the patterns and structures inherent in the data domain. Second, EchoMod provides both the original instance ($X$) and the noisy label ($\tilde{Y}$) as inputs to the generative model. This dual conditioning shapes the output, encouraging the model to produce a modified instance ($X'$) that closely aligns with the noisy label while still preserving essential characteristics of the original data. Striking this balance between label alignment and preventing excessive distortion is crucial.

    \textbf{Effectiveness and Flexibility} \quad
    Handling label noise in noisy label learning is a well-recognized challenge. Previous works have primarily focused on the utilization and optimization of internal data. Our approach introduces a novel perspective by integrating external knowledge to enhance model robustness. This integration does not compromise fairness, as our framework is compatible with different controllable generative models. In this paper, we instantiate EchoMod with ControlNet as a representative implementation, while a systematic comparison across CGMs is beyond the scope of this work. This approach is particularly advantageous when dealing with ambiguous data. The inherent ambiguity in the data can lead to low confidence in direct discrimination. Instead, by modifying the input data through controllable generative models, we can better resolve discrepancies between instances and labels while preserving the meaningful characteristics of the original data. This not only improves the model's discriminative capability but also enhances overall performance and reliability.

    \textbf{Flexible Generalization with Minimal Tuning} \quad
    In most cases, EchoMod can leverage a pre-trained controllable generative model without fine-tuning during the alignment process. This preserves the model's general prior and suggests potential applicability beyond the evaluated benchmarks, and cross-domain robustness is an important direction that requires dedicated evaluation. While EchoMod can be effective without fine-tuning, additional performance gains might be realized by tailoring the controllable generative model to highly specialized tasks or data distributions. In such cases, fine-tuning could lead to better alignment between instances and noisy labels, especially in specialized applications such as medical imaging or scientific data.

    \textbf{Visualization and Comparison of Generation Examples} \quad
    To further illustrate the significance of controllable generative models (CGM) and their advantages over non-controllable generative models (NGM), we provide additional generation examples in Table~\ref{gm:results}. Note that the ``Original Instance'' is paired with the observed noisy label and may be semantically mismatched by design. Specifically, we compare the modifications produced by two representative controllable generative models (ControlNet and UniControl) against two advanced non-controllable generative models (GPT-4 and Gemini).

    As depicted, CGMs successfully maintain intrinsic instance characteristics while aligning instances effectively with their noisy labels. In contrast, NGMs struggle to preserve crucial features, often leading to semantic misalignments or unrealistic modifications. For instance, when converting a hoodie to a T-shirt, CGMs effectively adjust clothing style while preserving facial and body features, whereas NGMs drastically distort or remove essential details.

    These examples provide qualitative evidence consistent with the motivation in (\S~\ref{Analysis}), demonstrating that CGMs substantially enhance instance-label alignment, reduce information distortion, and facilitate stable predictions in noisy label scenarios. Thus, controllable generative models appear to better preserve task-relevant structure while aligning instances with noisy labels.

    \begin{table*}[ht]
        \centering
        \caption{Qualitative generation examples under noisy labels. The \emph{Noisy Label} column shows the observed noisy label $\tilde{y}$, and the \emph{Original Instance} may be semantically mismatched by design. Each model modifies the original instance given $\tilde{y}$.}
        \footnotesize
        \resizebox{\textwidth}{!}{
            \begin{tabular}{cccccc}
                \toprule
                & & \multicolumn{2}{c}{CGM} & \multicolumn{2}{c}{NGM} \\
                \cmidrule(r){3-6}
                Noisy Label & Original Instance & ControlNet & UniControl & GPT-4 & Gemini \\
                Cat & \includegraphics[width=0.14\textwidth]{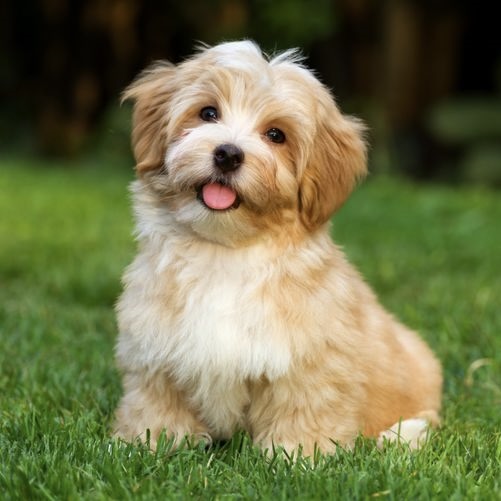} &
                \includegraphics[width=0.14\textwidth]{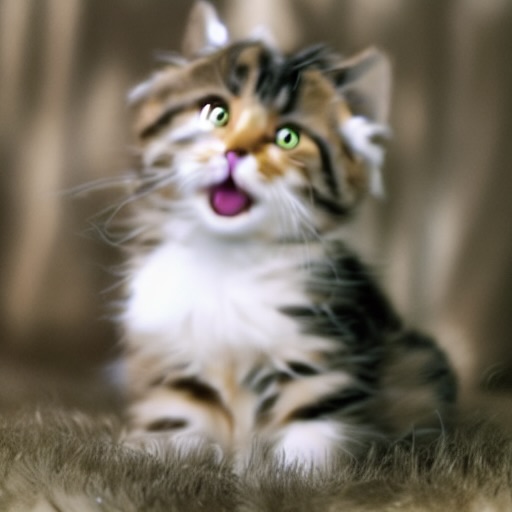} &
                \includegraphics[width=0.14\textwidth]{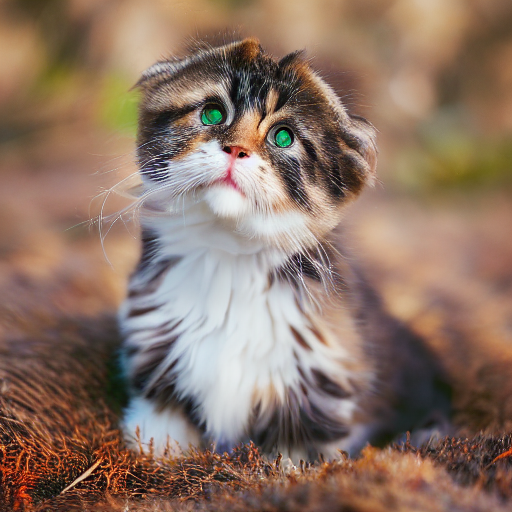} &
                \includegraphics[width=0.14\textwidth]{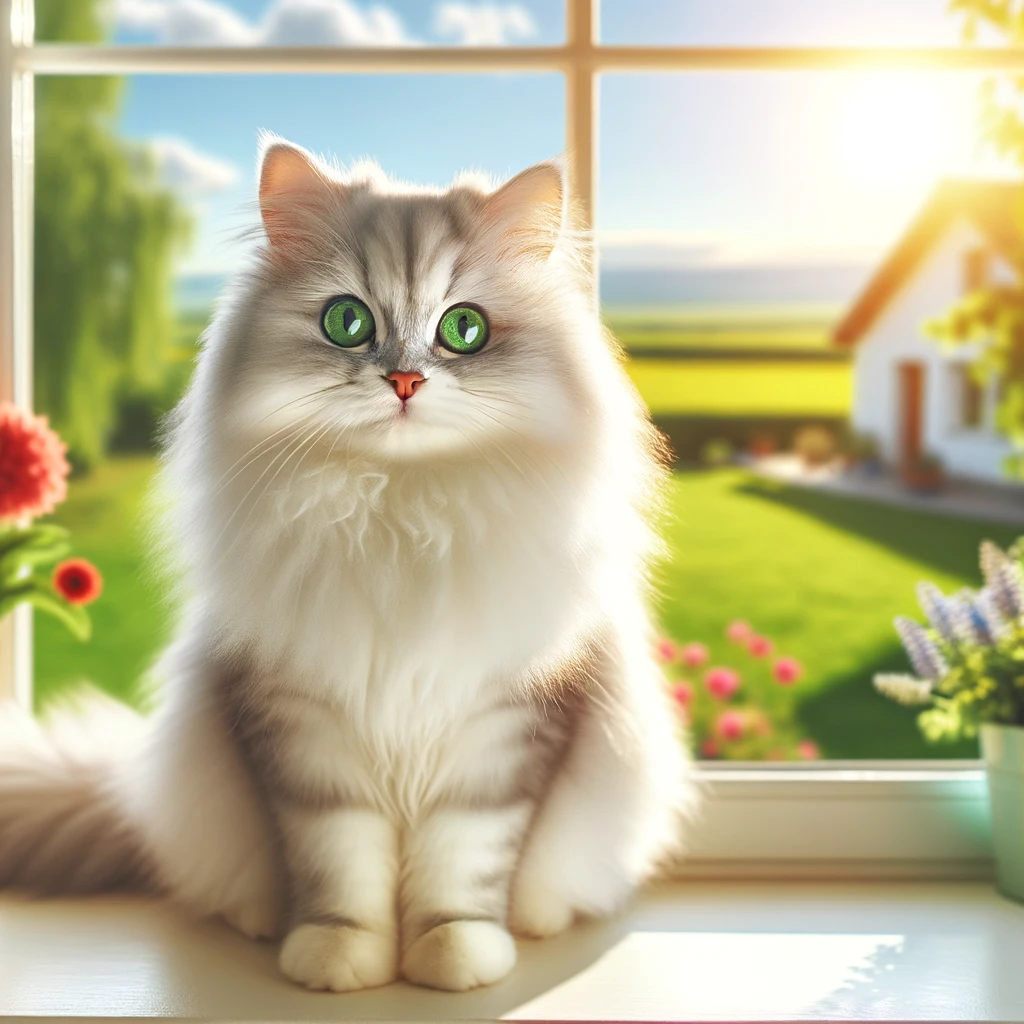} &
                \includegraphics[width=0.14\textwidth]{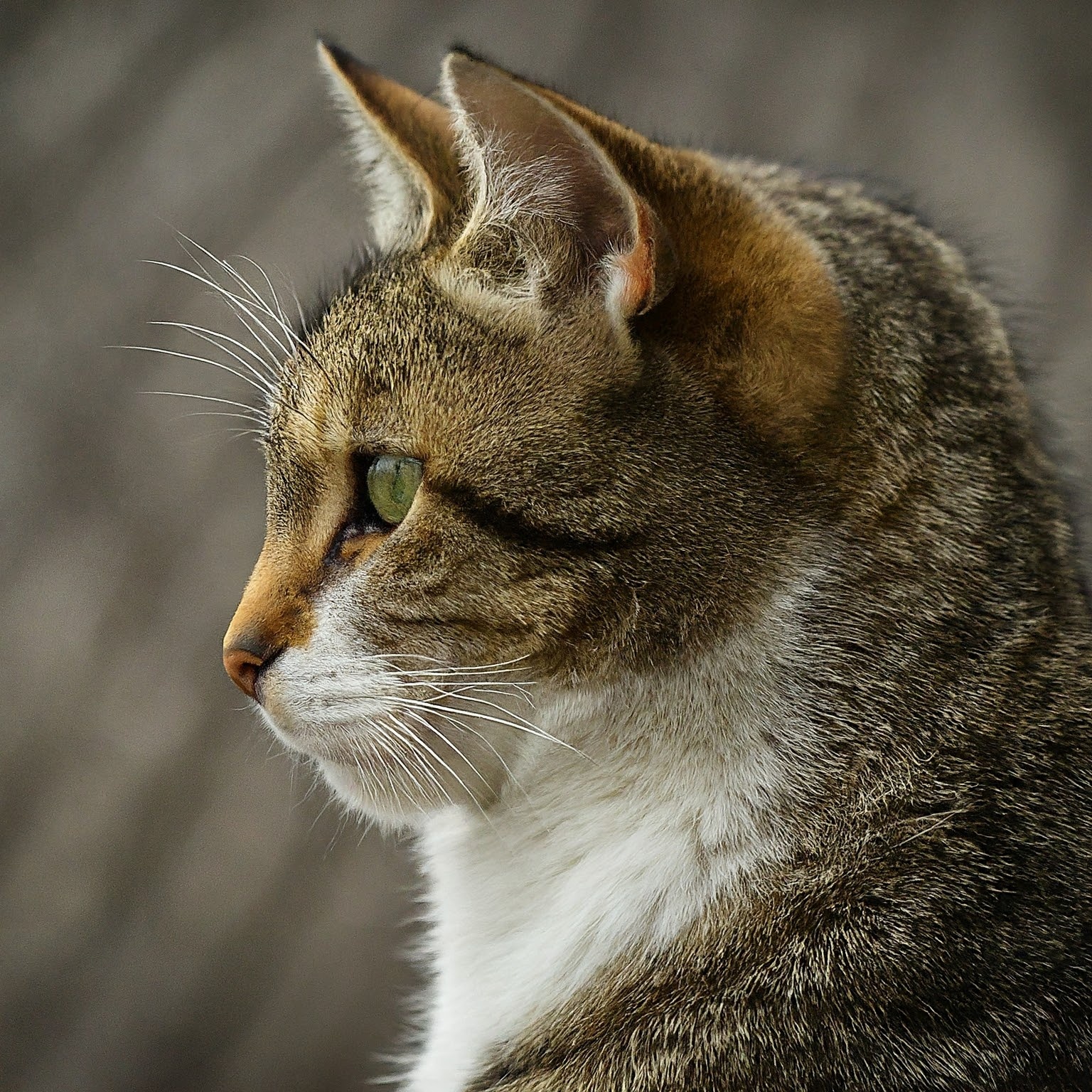} \\
                Magpie & \includegraphics[width=0.14\textwidth]{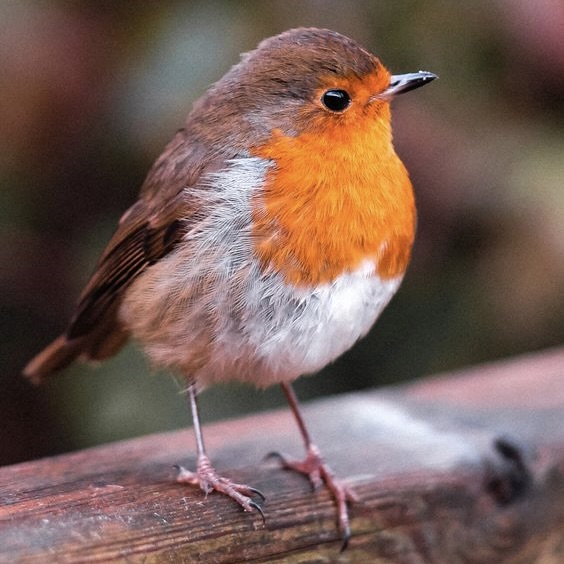} &
                \includegraphics[width=0.14\textwidth]{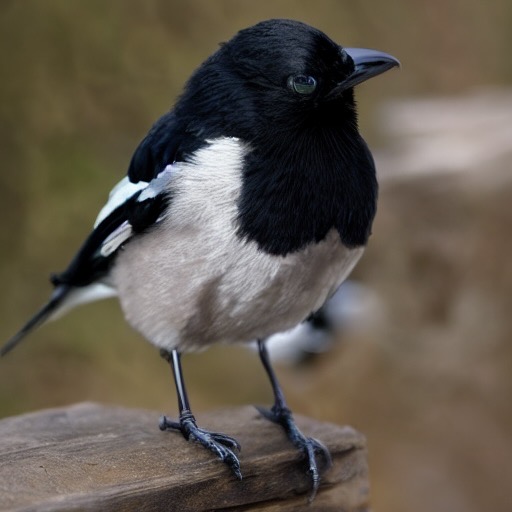} &
                \includegraphics[width=0.14\textwidth]{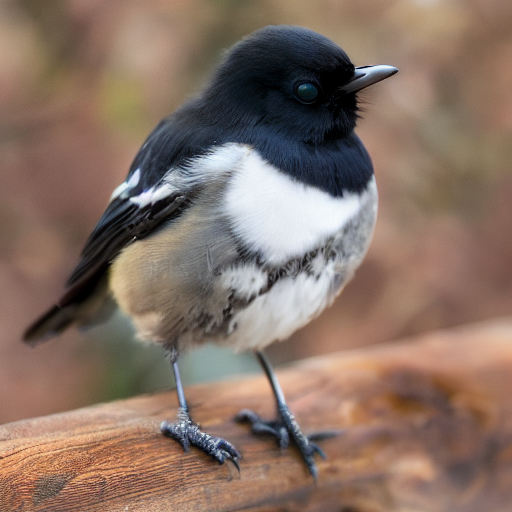} &
                \includegraphics[width=0.14\textwidth]{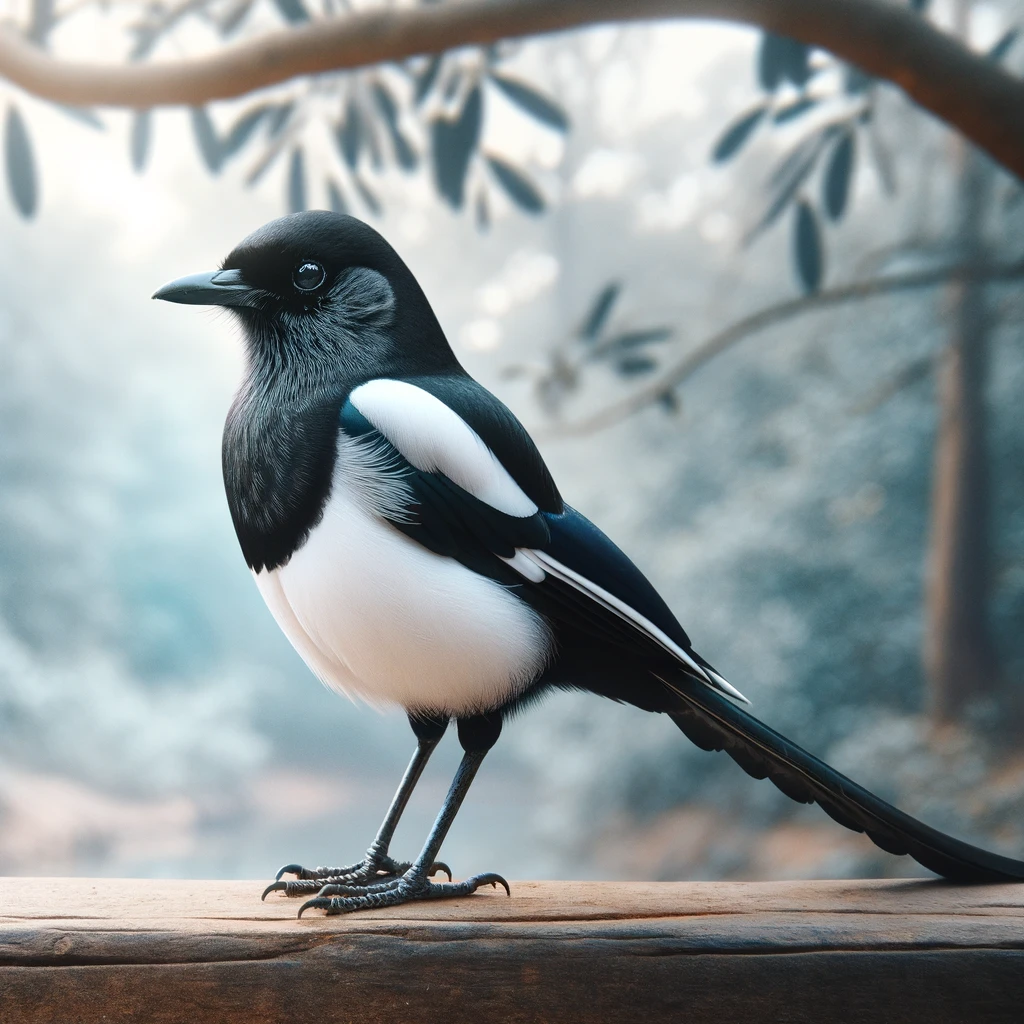} &
                \includegraphics[width=0.14\textwidth]{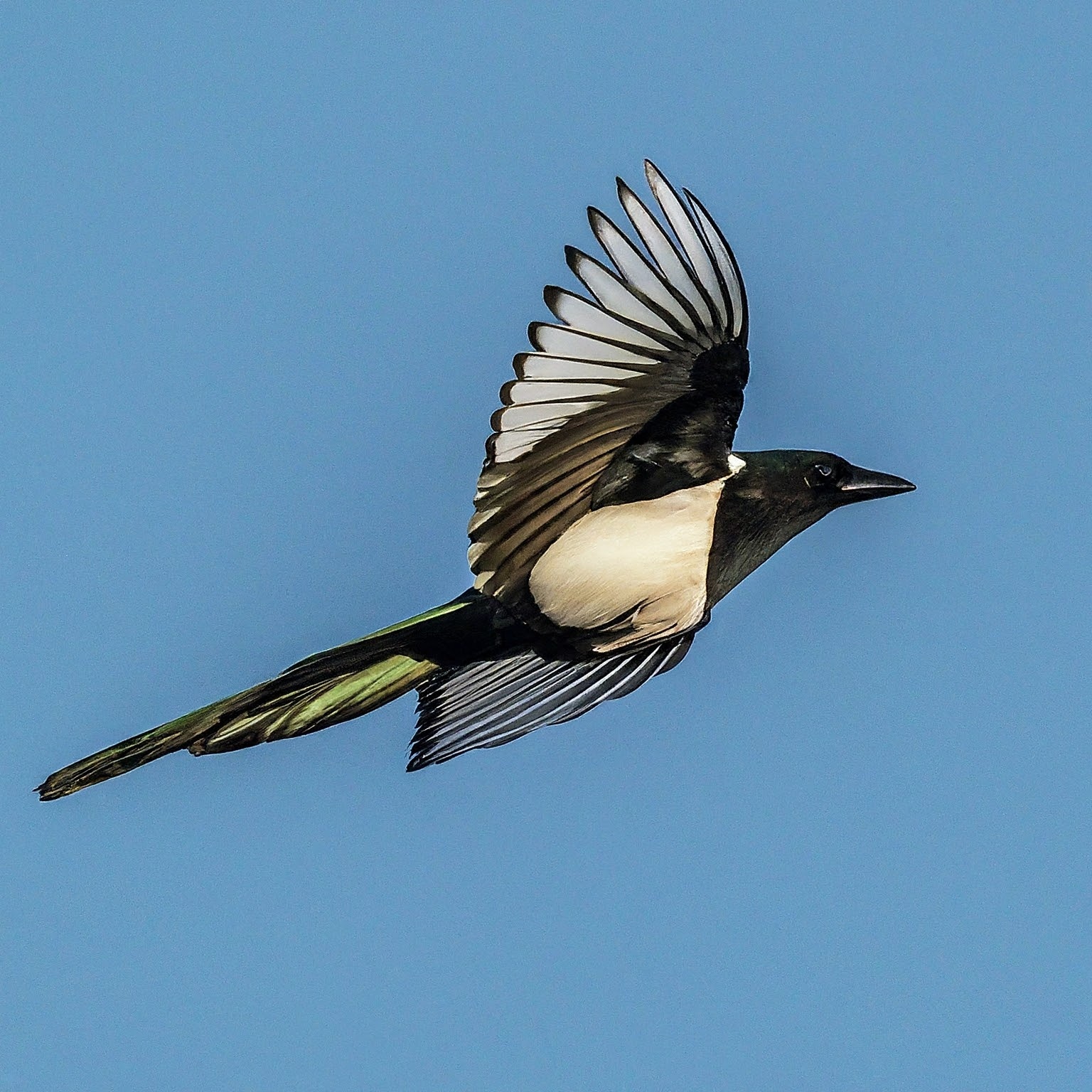} \\
                T-shirt & \includegraphics[width=0.14\textwidth]{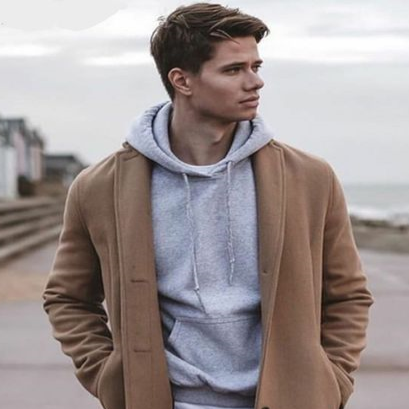} &
                \includegraphics[width=0.14\textwidth]{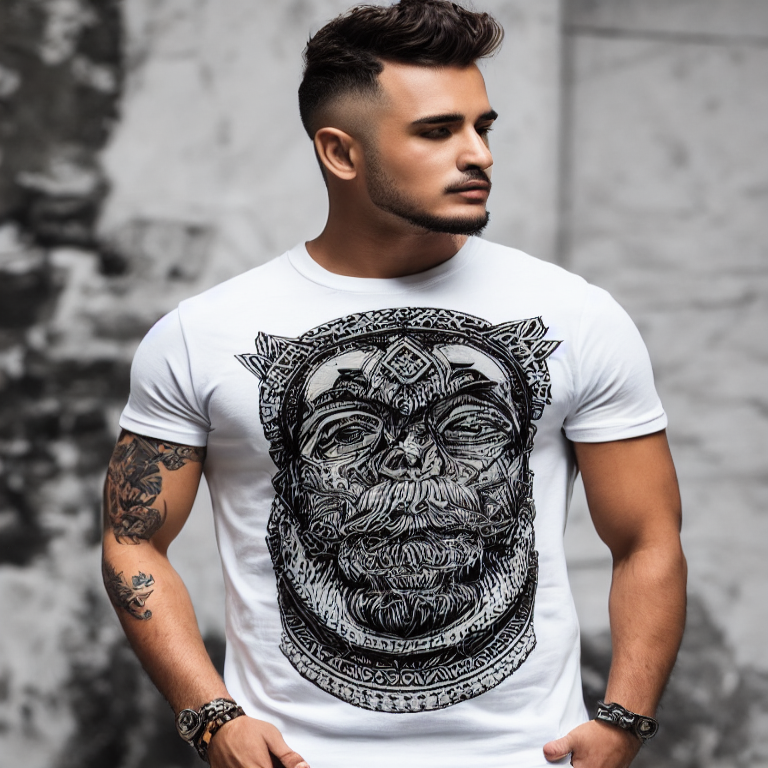} &
                \includegraphics[width=0.14\textwidth]{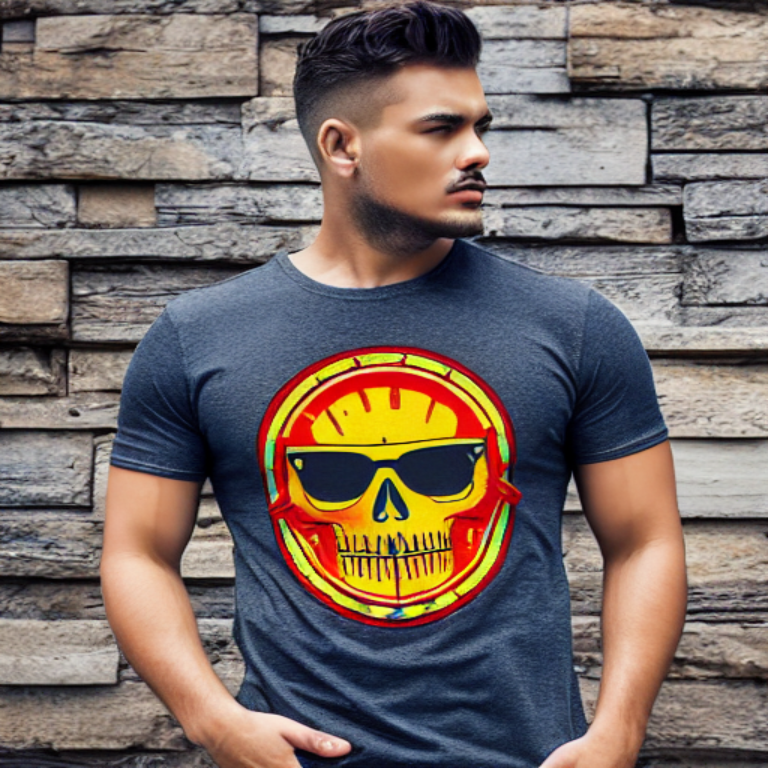} &
                \includegraphics[width=0.14\textwidth]{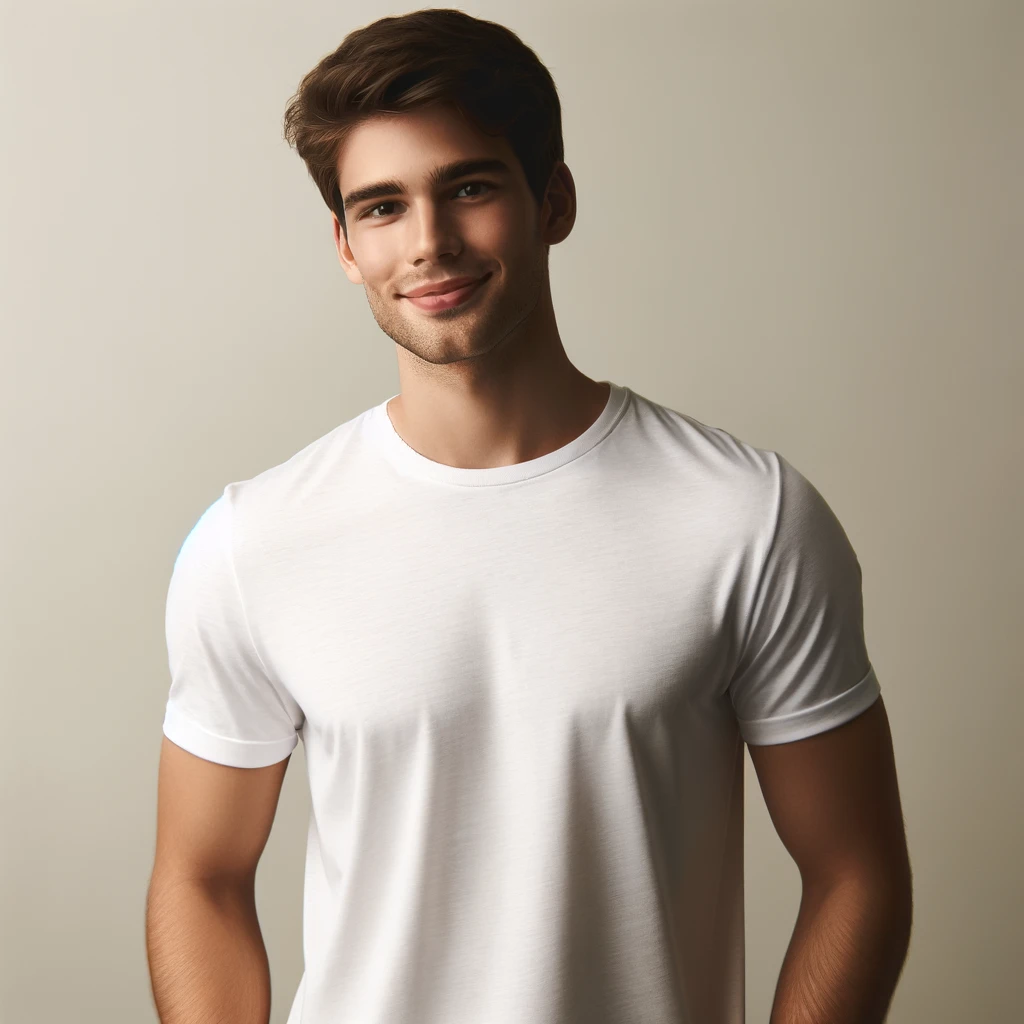} &
                \includegraphics[width=0.14\textwidth]{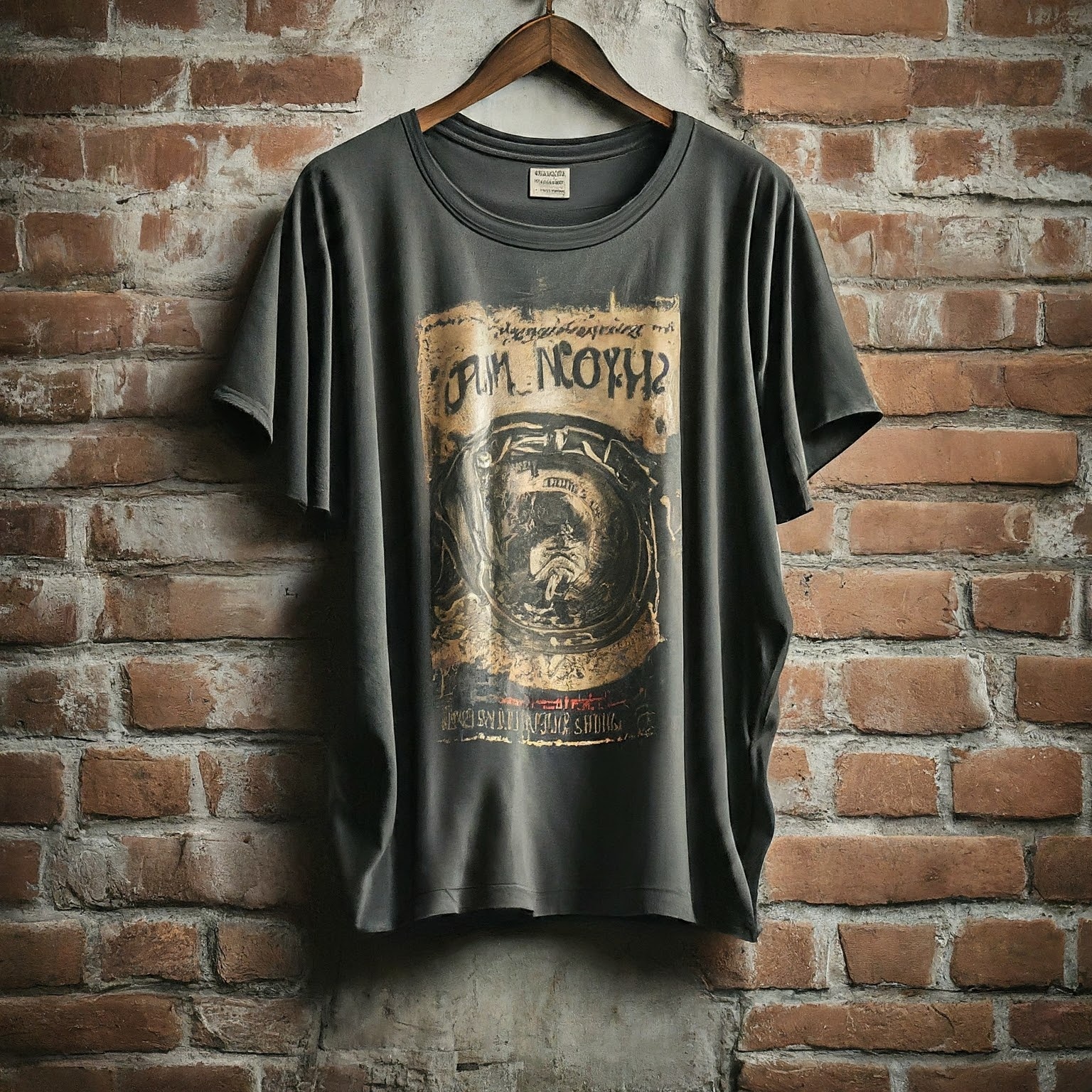} \\
                Fabric bag & \includegraphics[width=0.14\textwidth]{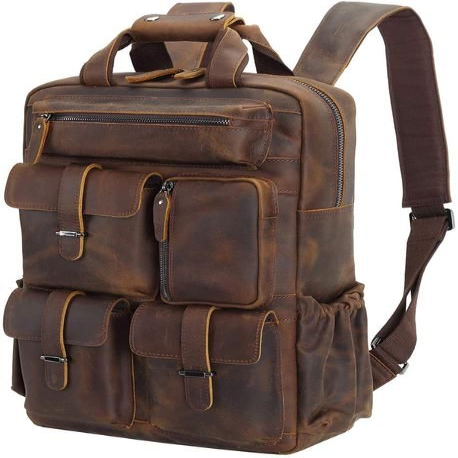} &
                \includegraphics[width=0.14\textwidth]{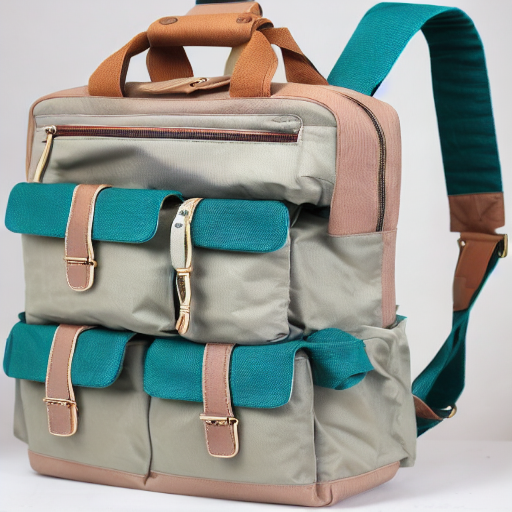} &
                \includegraphics[width=0.14\textwidth]{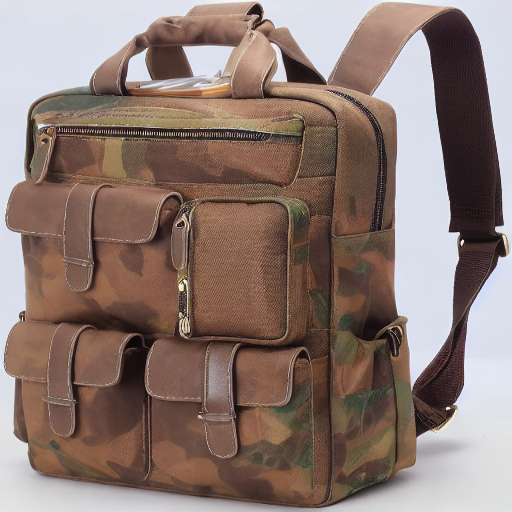} &
                \includegraphics[width=0.14\textwidth]{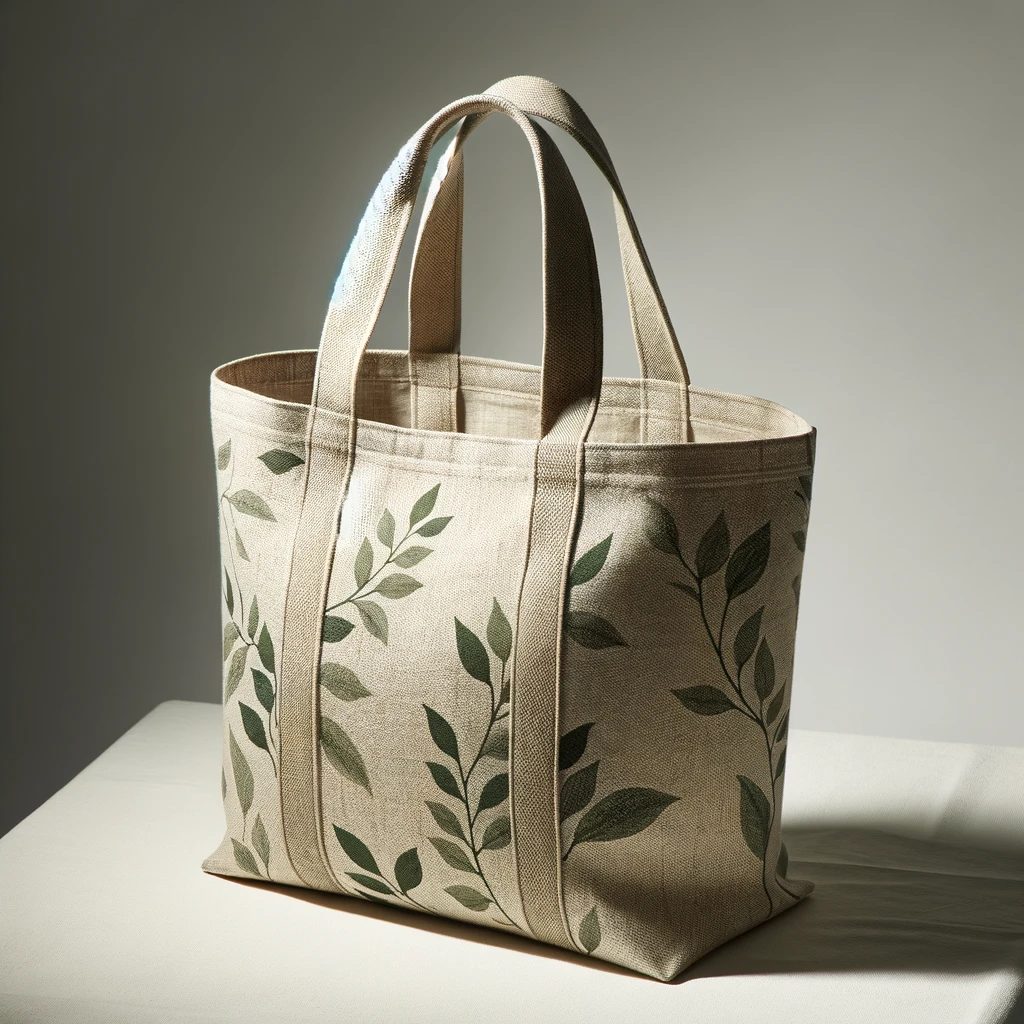} &
                \includegraphics[width=0.14\textwidth]{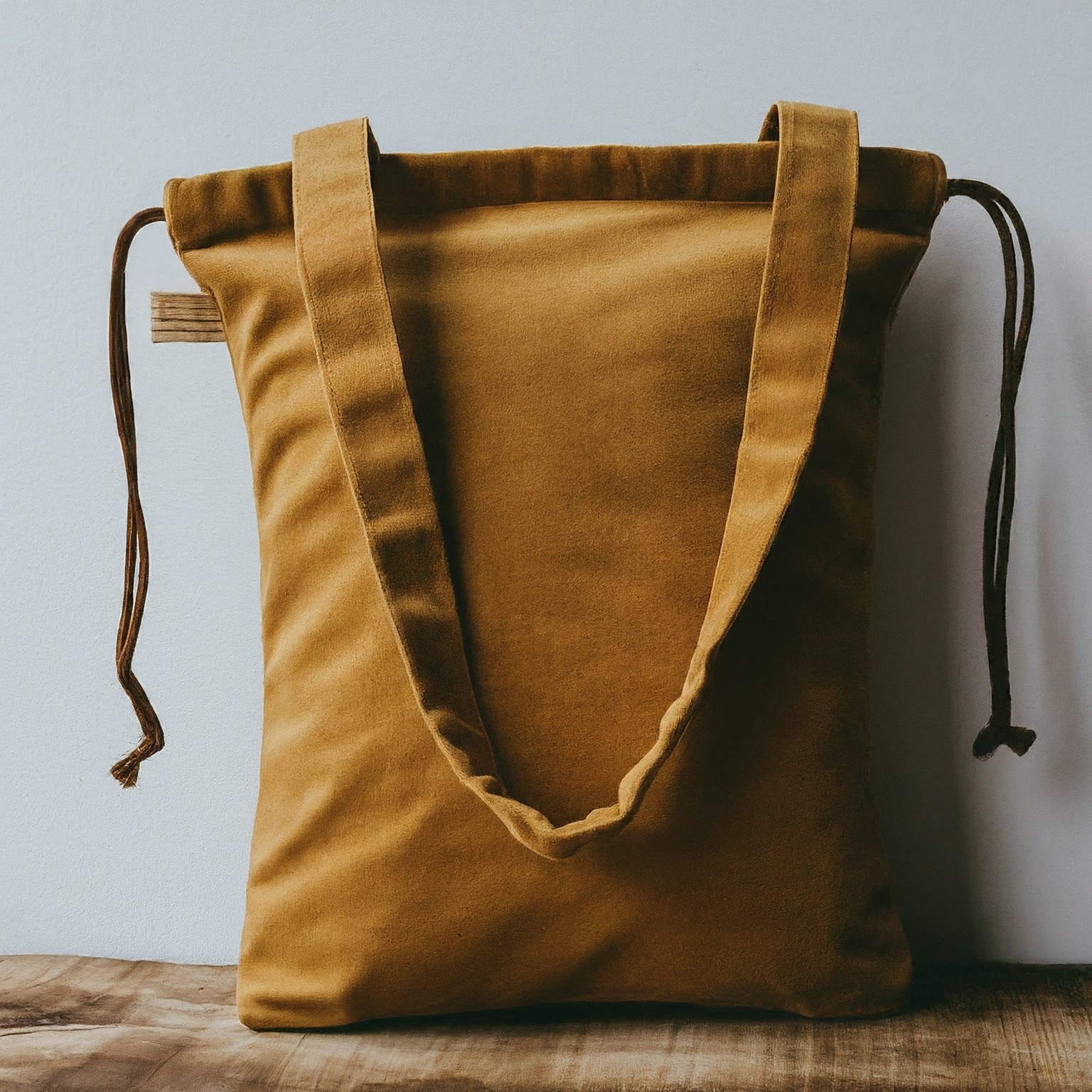} \\
                Dress shoes & \includegraphics[width=0.14\textwidth]{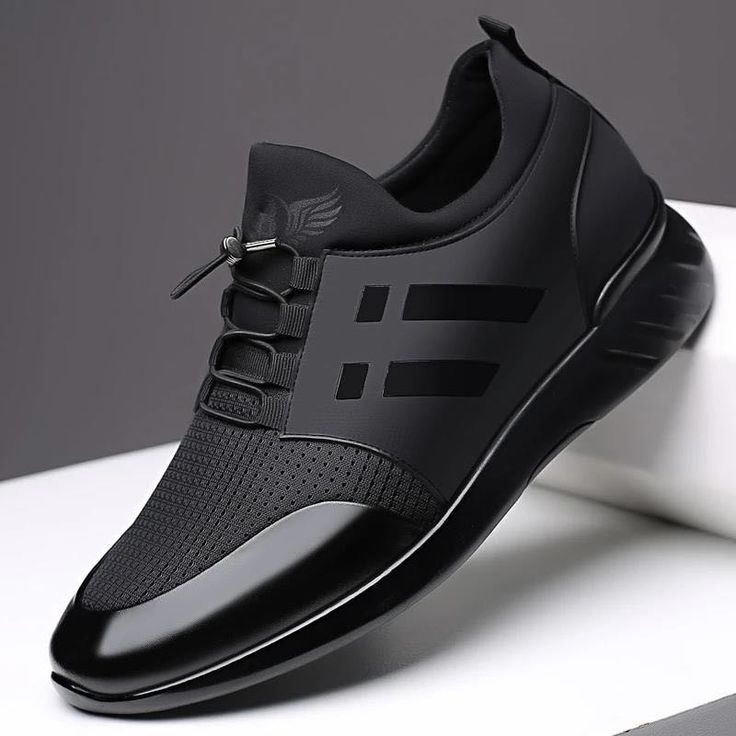} &
                \includegraphics[width=0.14\textwidth]{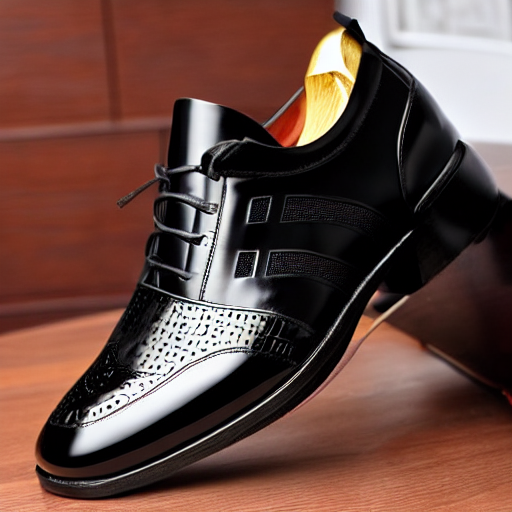} &
                \includegraphics[width=0.14\textwidth]{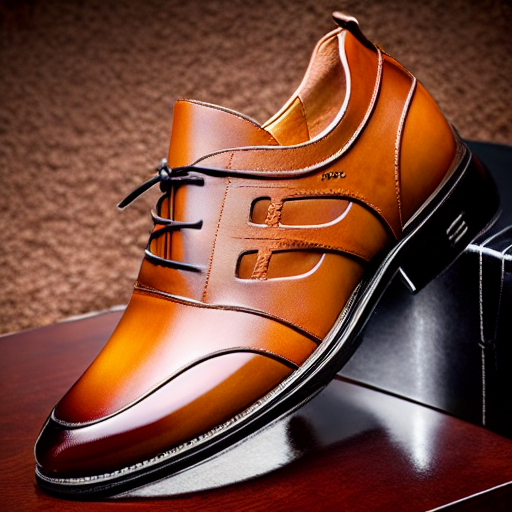} &
                \includegraphics[width=0.14\textwidth]{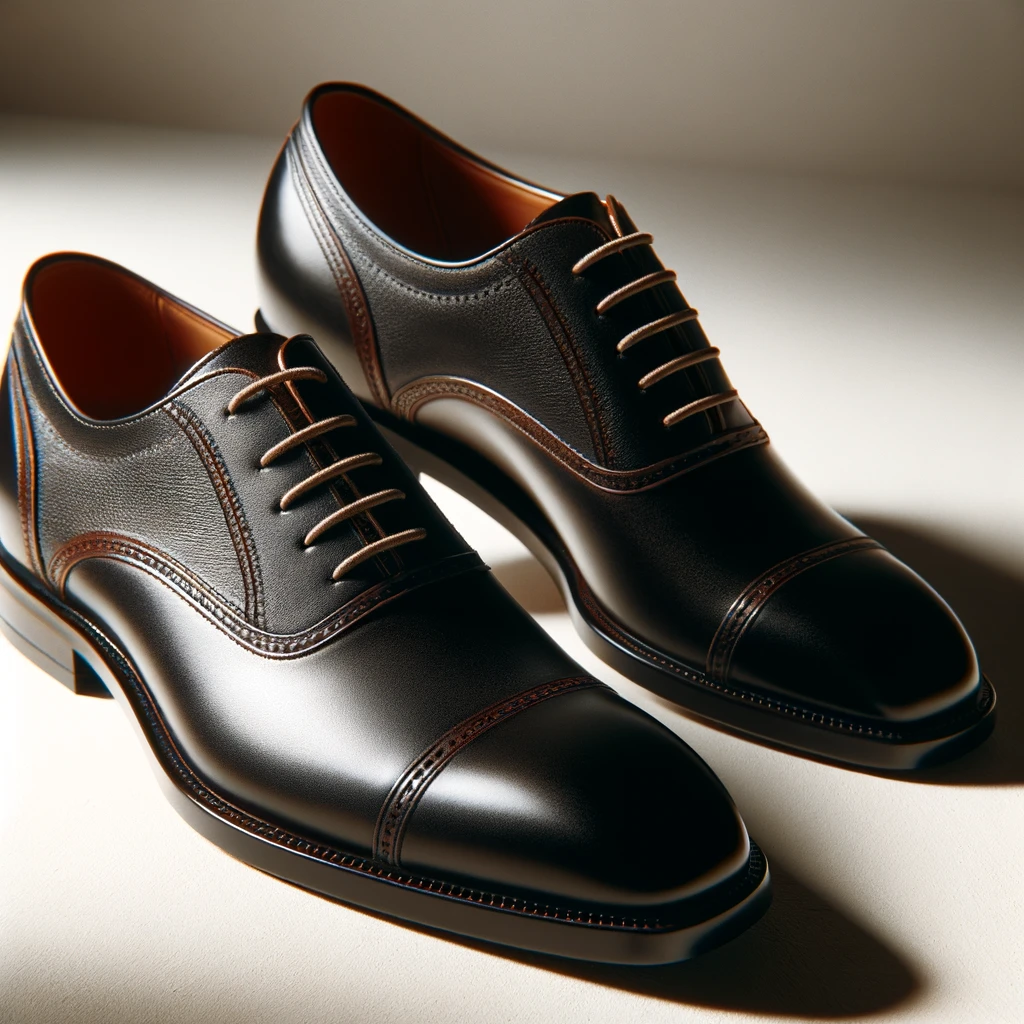} &
                \includegraphics[width=0.14\textwidth]{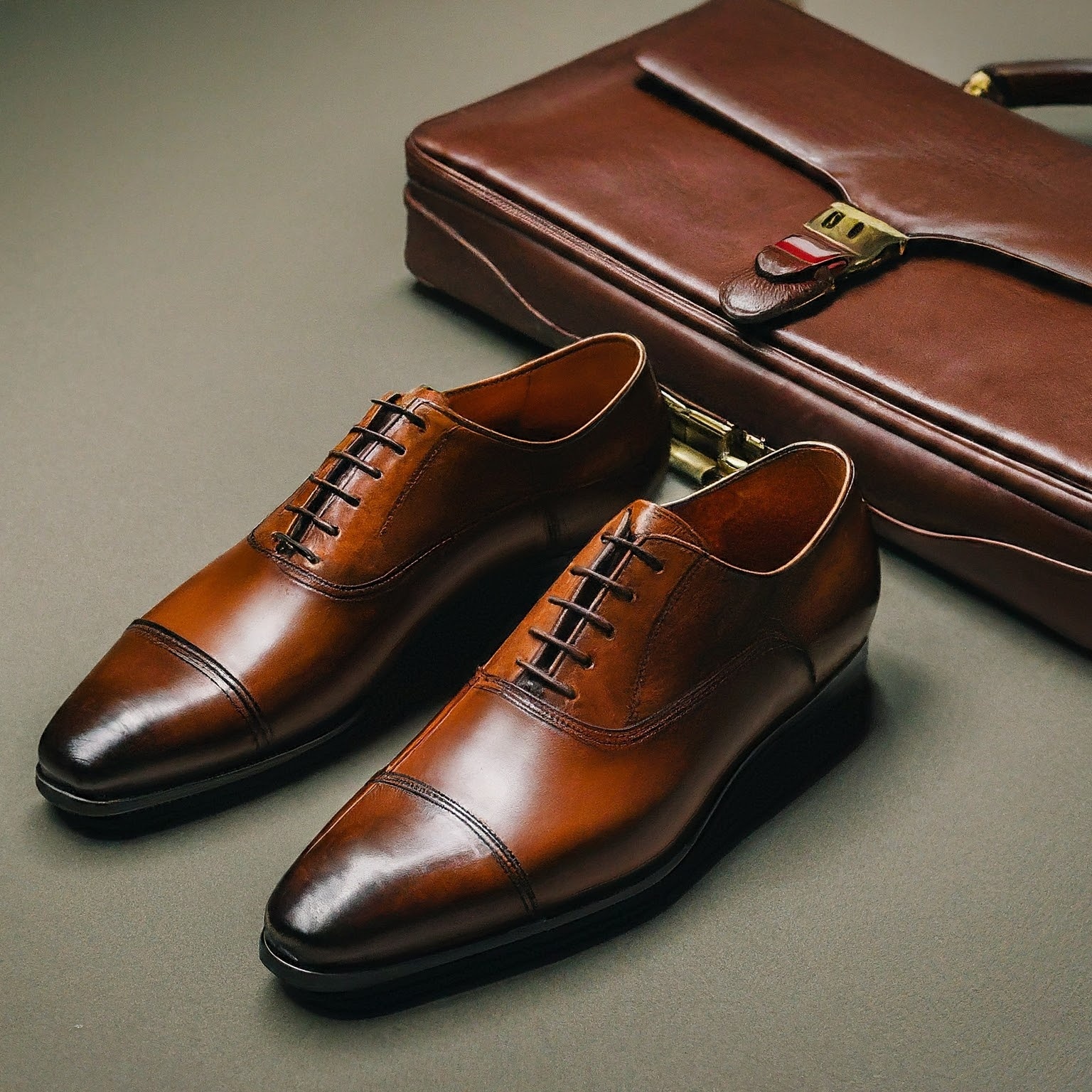}
            \end{tabular}
        }
        \label{gm:results}
    \end{table*}

    \subsection{EchoSelect: Instance Selection}

    \textbf{Motivation} \quad
    While EchoMod improves the alignment between instances and noisy labels, some modified instances may still exhibit inconsistencies. Additionally, the instance modification process can introduce distribution shifts between the modified training data and the true test distribution. EchoSelect safeguards against these issues by identifying and retaining only the most reliable instances after modification. This filtering enhances model robustness, reduces the impact of noisy data, and mitigates distribution shifts introduced by instance modification.

    \textbf{Mechanism} \quad
    EchoSelect employs a metric to assess the similarity between modified instances and a reference representation of clean data. We use cosine similarity between feature vectors extracted using a suitable feature extractor (\emph{e.g.}, the image encoder of CLIP~\cite{radford2021learning}):
    \begin{equation}
        S(X', X) = \frac{\bm{z}(X') \cdot \bm{z}(X)}{\|\bm{z}(X')\| \|\bm{z}(X)\|}\,,
        \label{eq:distance_metric}
    \end{equation}
    where $X'$ and $X$ are modified and original instances, and $\bm{z}$ denotes the feature extractor.

    \textbf{Selection Process} \quad
    EchoSelect calculates similarity for all modified instances by comparing them to their original counterparts. To mitigate distribution shifts, priority is given to preserving original instances whenever possible. For each original input, the refined training set contains exactly one selected sample: if $S(X', X) \geq \tau$, we retain the original instance; otherwise, we replace it with the modified instance. Accordingly, the refined set has the same cardinality as the original noisy training set. In Table~\ref{table:selection} and Figure~\ref{parameter_sensitivity}, the reported selection counts refer only to the number of retained original instances, which varies with $\tau$. The threshold $\tau$ therefore controls the trade-off between preserving the original data distribution and adopting modified instances for stronger label alignment. Section~\ref{analyses} characterizes how $\tau$ affects selection precision and coverage, and the resulting curves reveal a stable operating region under the evaluated settings.

    \subsection{EchoAlign: Optimized Combination}
    \begin{algorithm}[ht]
        \caption{EchoAlign Framework}
        \label{alg:echoalign}
        \begin{algorithmic}[1]
            \REQUIRE Pre-trained controllable generative model $f_{\theta}$, Noisy dataset $(X, \tilde{Y})$, Threshold $\tau$, Feature Extractor
            \ENSURE Refined training dataset
            \STATE Generate modified instances: $X' \leftarrow f_{\theta}(X,\tilde{Y})$
            \STATE Compute similarity: $S(X', X)$ using \eqref{eq:distance_metric}

            \STATE \# Construct a refined dataset with one selected sample per original input
            \STATE \textbf{Part 1: Original Instances}
            \STATE Retain original instances where $S(X', X) \geq \tau$
            \STATE \textbf{Part 2: Modified Instances}
            \STATE Replace the remaining instances with their modified versions where $S(X', X) < \tau$

            \STATE Combine Part 1 and Part 2 to form the refined dataset of the same size as the original training set
            \STATE Return the refined dataset
        \end{algorithmic}
    \end{algorithm}
    The integration of EchoMod and EchoSelect enables the creation of a refined training dataset that is aligned with noisy labels and filtered for quality. This optimized dataset is better suited for robust learning in the presence of large label noise. Since the refined training dataset can be further used to train a supervised or self-supervised model for LNL, EchoAlign can be combined with advanced LNL methods to further mitigate the impact of label noise. The integration of EchoMod and EchoSelect is encapsulated in Algorithm~\ref{alg:echoalign}, which details the steps for modifying instances and selecting the optimal subset.

    \section{Experiments}
    \label{experiments}

    To evaluate the robustness and effectiveness of our proposed method, we conducted a comprehensive set of experiments across multiple datasets and baseline comparisons. The detailed implementation settings, including model configurations, hyperparameters, and data preprocessing, are provided in Section \ref{experiment_setup}.

    \subsection{Experiment Setup}
    \label{experiment_setup}

    \textbf{Dataset} \quad
    Our experiments are conducted on two synthetic datasets: CIFAR-10 and CIFAR-100 \cite{krizhevsky2009CIFAR}, and a real-world dataset: CIFAR-10N \cite{wei2022learning}. CIFAR-10 and CIFAR-100 each contain 50,000 training and 10,000 testing images, with a size of 32$\times$32, covering 10 and 100 classes respectively. CIFAR-10N utilizes the same training images from CIFAR-10 but with labels re-annotated by humans. Although CIFAR-10 is a clean dataset, inherent ambiguity in many images leads to prevalent label noise, as even humans struggle to provide consistent labels, a phenomenon reflected in CIFAR-10N. Following previous research protocols \cite{Bai2021PES, xia2019anchor, xia2023regularly}, we corrupted these synthetic datasets using three types of label noise. Specifically, symmetric noise randomly alters a proportion of labels to different classes to simulate random errors; pair flip noise changes labels to adjacent classes with a certain probability; and instance-dependent noise modifies labels based on image features to related incorrect classes. Due to the inherent ambiguity in CIFAR-10N images, correcting label noise has limited impact on performance, making it a more practical choice over Clothing1M \cite{Xiao2015Clothing}. A detailed runtime analysis is provided in Section \ref{runtime}, demonstrating that the runtime is reasonable across different datasets and can be further optimized using model acceleration techniques. For CIFAR-10N, we use four noisy label sets: ``Random i=1, 2, 3'', each representing labels provided by one of three independent annotators; and ``Worst'', which selects the noisiest label when incorrect annotations are present.

    \textbf{Baseline} \quad
    We compare EchoAlign against various paradigms of baselines for addressing label noise. Under the robust loss function paradigm, we include APL \cite{ma2020normalized}, PCE \cite{menon2019can}, AUL \cite{zhou2023asymmetric}, and CELC \cite{wei2023logitclip}; under the loss correction paradigm, we adopt T-Revision \cite{xia2019anchor} and Identifiability \cite{liu2023identifiability}; under the label correction paradigm, we select Joint \cite{tanaka2018joint}; and under the sample selection paradigm, we employ Co-teaching \cite{han2018co}, SIGUA \cite{pmlr-v119-han20c}, and Co-Dis \cite{xia2023combating}. We compare these methods against a simple cross-entropy (CE) loss baseline. Following the fair baseline design proposed by \cite{xia2023regularly}, we do not compare with methods such as MixUp \cite{Zhang2017MixUp}, DivideMix \cite{Li2020DivideMix}, and M-correction \cite{ICML2019_UnsupervisedLabelNoise}, as these involve semi-supervised learning, making such comparisons unfair due to inconsistent settings.

    \textbf{Implementation Details} \quad
    All experiments were conducted on an NVIDIA V100 GPU using PyTorch. The model architectures and parameter settings were kept consistent with previous studies \cite{Bai2021PES}. The experiments were configured with a learning rate of 0.1, using the Stochastic Gradient Descent (SGD) optimizer with a momentum of 0.9, and a weight decay set to \(1 \times 10^{-4}\). We applied 30\% and 50\% symmetric noise and 45\% pair flip noise on the CIFAR-10 and CIFAR-100 datasets to assess model performance. The CIFAR-10 dataset utilized the standard ResNet-18 \cite{He2016ResNet} architecture, while CIFAR-100 used ResNet-34. For the CIFAR-10N dataset, the same ResNet-18 model was used. Prior to training, ControlNet was utilized as our reference model in the controllable generation model module. This choice was strategic: we instantiate EchoMod with ControlNet, a widely-used controllable diffusion model, and keep the generator and prompting settings fixed across all experiments. By avoiding generator-specific tuning and using a standard instantiation, the reported gains are less likely to be an artefact of a particularly strong or heavily engineered generator. The results reported in this paper correspond to the ControlNet instantiation, with all generation settings fixed as summarized in Appendix~B. We employed the Canny edge detector as a simple preprocessor to extract features from the instances, using labels as textual controls with the prompt "a photo of \{label\}". No additional or negative prompts were used, and the sampling process was limited to 20 steps. All experiments were repeated three times with different random seeds, and results are reported as averages with standard deviations.

    \textbf{Data preprocessing} \quad For all datasets, including CIFAR-10, CIFAR-100, and CIFAR-10N, we adopted a unified data augmentation strategy. Specifically, we first applied 4-pixel padding, followed by random cropping to \(32 \times 32\) pixels. We then applied random horizontal flipping and normalization.

    \textbf{Hyperparameter settings} \quad For the ControlNet controllable generation model, we used the simplest Canny preprocessor with both the low threshold and high threshold set to 75. The prompt used was “a photo of {label}” without any additional prompts or negative prompts. The feature maps output by the preprocessor and the generated images were both set to \(512 \times 512\) pixels. The diffusion process consisted of 20 steps. For reproducibility, we report the $\tau$ values used in our experiments (e.g., $\tau=0.4$ for 30\% noise and $\tau=0.52$ for 45\%/50\% noise), selected via the lightweight calibration procedure described in the Sensitivity Analysis. The hyperparameters for the training are detailed in Table \ref{table:hyperparameter}.

    \begin{table*}[!t]
        \centering
        \footnotesize
        \caption{Training hyperparameters for CIFAR-10/CIFAR-10N and CIFAR-100.}
        \begin{tabular}{ccc}
            \toprule
            & CIFAR-10/CIFAR-10N & CIFAR-100       \\
            \midrule
            architecture      & ResNet-18          & ResNet-34       \\
            optimizer         & SGD                & SGD             \\
            loss function     & CE                 & CE              \\
            learning rate(lr) & 0.1                & 0.1             \\
            lr decay          & 100th and 150th    & 100th and 150th \\
            weight decay      & \(10^{-4}\)        & \(10^{-4}\)     \\
            momentum          & 0.9                & 0.9             \\
            batch size        & 128                & 128             \\
            training samples  & 45,000             & 45,000          \\
            training epochs   & 200                & 200             \\
            \bottomrule
        \end{tabular}
        \label{table:hyperparameter}
    \end{table*}

    \begin{table*}[!t]
        \centering
        \caption{Comparison of test accuracy (\%) with state-of-the-art methods on synthetic datasets CIFAR-10 and CIFAR-100. The best three results are bolded and the best one is underlined.}
        \resizebox{\textwidth}{!}{
            \begin{tabular}{ccccccc}
                \toprule
                & & \multicolumn{2}{c}{Symmetric} & Pair-flip & \multicolumn{2}{c}{Instance} \\
                \cmidrule(r){3-7}
                Datasets                    & Methods         & 30\%                                & 50\%                    & 45\%                    & 30\%                                & 50\%                    \\
                \midrule
                \multirow{13}{*}{CIFAR-10}  & CE              & \(73.17\pm1.13\)                    & \(52.59\pm0.70\)        & \(51.49\pm0.42\)& \(71.56\pm0.19\)& \(49.20\pm0.42\)\\
                & APL             & \(85.54\pm0.51\)                    & \(78.36\pm0.47\)        & \(80.84\pm0.72\)        & \(77.57\pm0.15\)                    & \(39.45\pm6.51\)        \\
                & PCE             & \(86.12\pm0.85\)                    & \(74.03\pm4.96\)        & \(65.08\pm3.41\)        & \(85.64\pm0.72\)                    & \(\bm{64.82\pm4.13}\)   \\
                & AUL             & \(88.09\pm0.78\)                    & \(82.81\pm1.16\)        & \(56.80\pm2.69\)        & \(86.35\pm0.90\)                    & \(60.75\pm3.77\)        \\
                & CELC            & \(82.51\pm0.22\)                    & \(\bm{85.08\pm3.95}\)   & \(\bm{85.72\pm4.52}\)   & \(86.67\pm1.47\)& \(61.85\pm4.98\)\\
                & T-Revision      & \(88.39\pm0.38\)                    & \(83.40\pm0.65\)        & \(83.61\pm1.06\)        & \(\bm{89.07\pm0.35}\)& \(\bm{66.93\pm4.14}\)\\
                & Identifiability & \(87.12\pm1.69\)                    & \(83.43\pm2.11\)        & \(83.65\pm2.46\)        & \(80.47\pm1.54\)& \(55.25\pm3.78\)\\
                & Joint           & \(\bm{89.34\pm0.52}\)               & \(85.06\pm0.29\)        & \(80.52\pm1.90\)        & \(\bm{88.41\pm1.02}\)                    & \(64.12\pm3.89\)        \\
                & Co-teaching     & \(88.93\pm0.56\)                    & \(74.02\pm0.04\)        & \(84.19\pm0.68\)        & \(87.07\pm0.35\)& \(60.09\pm3.31\)\\
                & SIGUA           & \(83.19\pm1.26\)                    & \(77.92\pm3.11\)        & \(70.39\pm1.94\)        & \(82.90\pm2.00\)                    & \(30.95\pm9.70\)        \\
                & Co-Dis          & \(\bm{89.20\pm0.13}\)               & \(\bm{85.36\pm0.94}\)   & \(\bm{85.02\pm1.33}\)   & \(87.13\pm0.25\)& \(62.77\pm3.90\)\\
                \cmidrule(r){2-7}
                & \textbf{Ours} & \(\underline{\bm{90.98 \pm 0.20}}\) & \(\underline{\bm{87.95 \pm 0.12}}\) & \(\underline{\bm{87.42 \pm 0.11}}\) & \
                \(\underline{\bm{89.18 \pm 0.20}}\) & \(\underline{\bm{77.81 \pm 0.30}}\)\\
                \midrule
                \midrule
                \multirow{13}{*}{CIFAR-100} & CE              & \(50.99 \pm 1.29\)                  & \(34.5 \pm 0.96\)       & \(37.03 \pm 0.41\) & \(50.33 \pm 2.14\) & \(34.70 \pm 1.45\) \\
                & APL             & \(55.78 \pm 0.91\)                  & \(46.96 \pm 0.81\)      & \(49.55 \pm 1.05\)      & \(43.30 \pm 1.57\) & \(29.01 \pm 0.09\) \\
                & PCE             & \(58.84 \pm 1.32\)                  & \(42.63 \pm 2.02\)      & \(41.05 \pm 2.83\)      & \(55.72 \pm 1.96\) & \(38.72 \pm 3.01\) \\
                & AUL             & \(\underline{\bm{69.89 \pm 0.21}}\) & \(\bm{60.00 \pm 0.40}\) & \(39.37 \pm 1.61\) & \(\underline{\bm{67.75 \pm 1.84}}\) & \(40.27 \pm 1.76\) \\
                & CELC            & \(\bm{67.96 \pm 1.88}\)             & \(\bm{60.71 \pm 2.39}\) & \(\bm{52.53 \pm 3.17}\) & \(\bm{66.25 \pm 1.93}\) & \(\bm{47.52 \pm 3.93}\) \\
                & T-Revision      & \(62.97 \pm 0.46\)                  & \(43.60 \pm 0.94\)      & \(49.33 \pm 1.10\)      & \(56.46 \pm 1.45\) & \(40.78 \pm 1.75\) \\
                & Identifiability & \(50.53 \pm 1.52\)                  & \(34.87 \pm 2.36\)      & \(38.16 \pm 2.68\)      & \(52.48 \pm 1.93\) & \(36.72 \pm 3.10\) \\
                & Joint           & \(63.69 \pm 0.84\)                  & \(55.62 \pm 1.68\)      & \(49.77 \pm 1.15\)      & \(64.15 \pm 1.11\) & \(\bm{45.47 \pm 2.73}\) \\
                & Co-teaching     & \(59.49 \pm 0.36\)                  & \(52.19 \pm 1.42\)      & \(47.53 \pm 1.39\)      & \(56.71 \pm 1.26\) & \(42.09 \pm 1.73\) \\
                & SIGUA           & \(54.22 \pm 0.90\)                  & \(50.64 \pm 3.92\)      & \(39.92 \pm 2.33\)      & \(53.19 \pm 2.64\) & \(38.50 \pm 1.69\) \\
                & Co-Dis          & \(64.02 \pm 1.37\)                  & \(54.55 \pm 2.06\)      & \(\bm{50.02 \pm 2.80}\) & \(59.15 \pm 1.92\) & \(43.38 \pm 1.25\) \\
                \cmidrule(r){2-7}
                & \textbf{Ours} & \(\bm{68.16 \pm 0.53}\) & \(\underline{\bm{60.78 \pm 0.46}}\) & \(\underline{\bm{60.31 \pm 0.37}}\) & \
                \(\bm{65.68 \pm 0.48}\) & \(\underline{\bm{57.21 \pm 0.60}}\)\\
                \bottomrule
            \end{tabular}
        }
        \label{table:cifar10}
    \end{table*}

    \subsection{Main Results}
    We evaluated our method on two synthetic datasets (CIFAR-10 and CIFAR-100) and one real-world dataset (CIFAR-10N). For CIFAR-10 and CIFAR-100, 90\% of the noisy-labeled data was used for training, 10\% for validation, and evaluation was performed on clean test samples. Several baseline results were obtained from previous work \cite{xia2023regularly}. As shown in Table~\ref{table:cifar10}, our method achieved state-of-the-art performance in most evaluated scenarios. Under challenging noise conditions (e.g., 50\% instance-dependent noise on CIFAR-10 and 45\% symmetric noise on CIFAR-100), our method significantly outperformed existing baselines, demonstrating its robustness against various types of noise. This robustness is particularly attributable to EchoMod's noise-independence, which enables the model to learn consistent features across different noise types and levels. Performance variations were mainly caused by differences in the number of clean samples in the datasets. On the real-world CIFAR-10N dataset, our method also outperformed state-of-the-art methods across all noise settings, exhibiting strong robustness with minimal variations in performance.

    \begin{table*}[!t]
        \centering
        \caption{Comparison of test accuracy (\%) with state-of-the-art methods on real-world datasets CIFAR-10N. The best three results are bolded and the best one is underlined.}
        \resizebox{\textwidth}{!}{
            \begin{tabular}{cccccc}
                \toprule
                Datasets                    & Methods         & Random 1                & Random 2                & Random 3                & Worst                   \\
                \midrule
                \multirow{13}{*}{CIFAR-10N} & CE              & \(83.17 \pm 0.48\)      & \(82.74 \pm 0.42\)      & \(82.90 \pm 0.28\) & \(76.57 \pm 0.23\) \\
                & APL             & \(84.40 \pm 0.26\)      & \(84.45 \pm 0.50\)      & \(84.35 \pm 0.43\)      & \(78.16 \pm 0.17\)      \\
                & PCE             & \(63.06 \pm 0.37\)      & \(62.26 \pm 0.36\)      & \(35.47 \pm 0.36\)      & \(33.80 \pm 0.33\)      \\
                & AUL             & \(76.26 \pm 0.28\)      & \(75.24 \pm 0.20\)      & \(75.48 \pm 0.40\)      & \(63.61 \pm 1.62\)      \\
                & CELC            & \(83.11 \pm 0.14\)      & \(83.09 \pm 0.22\)      & \(82.60 \pm 0.04\)      & \(73.49 \pm 0.50\)      \\
                & T-Revision      & \(80.99 \pm 0.26\)      & \(78.99 \pm 1.59\)      & \(78.80 \pm 1.87\)      & \(\bm{78.37 \pm 0.96}\) \\
                & Identifiability & \(82.52 \pm 0.87\)      & \(81.97 \pm 0.85\)      & \(82.09 \pm 0.73\)      & \(71.62 \pm 1.16\)      \\
                & Joint           & \(\bm{88.20 \pm 0.29}\) & \(\bm{87.54 \pm 0.33}\) & \(\bm{87.67 \pm 0.22}\) & \(\bm{84.29 \pm 0.40}\) \\
                & Co-teaching     & \(82.28\pm 0.13\)       & \(82.45 \pm 0.23\)      & \(82.09 \pm 0.24\)      & \(79.62 \pm 0.25\)      \\
                & SIGUA           & \(\bm{87.67 \pm 1.18}\) & \(\bm{89.01 \pm 0.34}\) & \(\bm{88.40 \pm 0.42}\) & \(80.65 \pm 1.29\)      \\
                & Co-Dis          & \(80.81 \pm 0.23\)      & \(80.36 \pm 0.20\)      & \(80.76 \pm 0.13\)      & \(78.12 \pm 0.25\)      \\
                \cmidrule(r){2-6}
                & \textbf{Ours} & \(\underline{\bm{89.42 \pm 0.12}}\) & \(\underline{\bm{89.31 \pm 0.06}}\) & \
                \(\underline{\bm{89.80 \pm 0.25}}\) & \(\underline{\bm{84.35 \pm 0.09}}\)\\
                \bottomrule
            \end{tabular}
        }
    \end{table*}

    \subsection{In-Depth Analyses}
    \label{analyses}

    \textbf{Ablation Analysis} \quad To assess the effectiveness of EchoMod and EchoSelect, we conducted ablation studies by systematically disabling these components. Specifically, we evaluated two configurations: “Instance Modification Only” and “EchoSelect Only,” and compared both against the standard Cross-Entropy Loss (CE) as a baseline. These experiments were carried out under several settings with high noise rates, presenting significant challenges for the model. The experimental results in Table~\ref{table:ablation} revealed that when using only Instance Modification, the model's accuracy did not exceed the baseline CE, and even decreased. This decline primarily stems from the data distribution shift caused by solely using modified instances, adversely affecting the model's generalization capability. In contrast, using only EchoSelect improved performance but still fell short of the combined EchoAlign approach. This indicates that although EchoSelect significantly reduces the impact of noise, its effectiveness is limited by the number of available samples.

    \begin{table*}[!t]
        \centering
        \caption{Comparison with state-of-the-art methods on CIFAR-10 and CIFAR-100 in accuracy (\%).}
        \resizebox{\textwidth}{!}{
            \begin{tabular}{ccccc}
                \toprule
                & CIFAR-10 Pair-flip-45\% & CIFAR-10 IDN-50\% & CIFAR-100 Pair-flip-45\% & CIFAR-100 IDN-50\% \\
                CE                         & 51.49                  & 49.20             & 37.03                   & 34.70              \\
                Instance Modification Only & 42.77                  & 44.98             & 15.69                   & 16.36              \\
                EchoSelect Only            & 79.46                  & 65.77             & 44.24                   & 41.24              \\
                \textbf{Ours}              & \textbf{87.42}         & \textbf{77.81}    & \textbf{60.31}          & \textbf{57.21}     \\
                \bottomrule
            \end{tabular}
        }
        \label{table:ablation}
    \end{table*}

    \begin{table*}[!t]
        \centering
        \caption{Comparison of sample selection quality under CIFAR-10 instance-dependent noise.}
        \footnotesize
        \begin{tabular}{cccc}
            \toprule
            \multirow{3}{*}{Noise rate} &              & BLTM                    & Ours                    \\
            \cmidrule(r){2-4}
            & select. acc. & \# of selected examples & \# of selected examples \\
            \midrule
            \multirow{2}{*}{IDN-30\%}   & 96\%         & 17673 / 50000           & \textbf{26524 / 50000}  \\
            & 99\%         & 10673 / 50000           & \textbf{19010 / 50000}  \\
            \midrule
            \multirow{2}{*}{IDN-50\%}   & 94\%         & 8029 / 50000            & \textbf{11660 / 50000}  \\
            & 98\%         & 5098 / 50000            & \textbf{6090 / 50000}   \\
            \bottomrule
        \end{tabular}
        \label{table:selection}
    \end{table*}

    \textbf{Sensitivity Analysis} \quad The performance of EchoSelect is influenced by the threshold value \(\tau\), which affects both the quantity and the quality of the samples selected from noisy datasets. Under the simplified assumptions used in EchoSelect, \(\tau\) is expected to be close to 0.5 and can serve as a practical starting point. To contextualize the selection behaviour, we compare against the state-of-the-art BLTM \cite{Yang2022bayeslabel}, with its reported results cited from the original publication. As shown in Table~\ref{table:selection}, EchoSelect selects substantially more samples than BLTM at comparable selection-accuracy levels. In particular, under 30\% instance-dependent noise, when the selection accuracy reaches 99\%, EchoSelect retains nearly twice as many samples as BLTM.
    \begin{figure*}[!t]
        \centering
        \begin{subfigure}{0.49\textwidth}
            \centering
            \includegraphics[width=\linewidth]{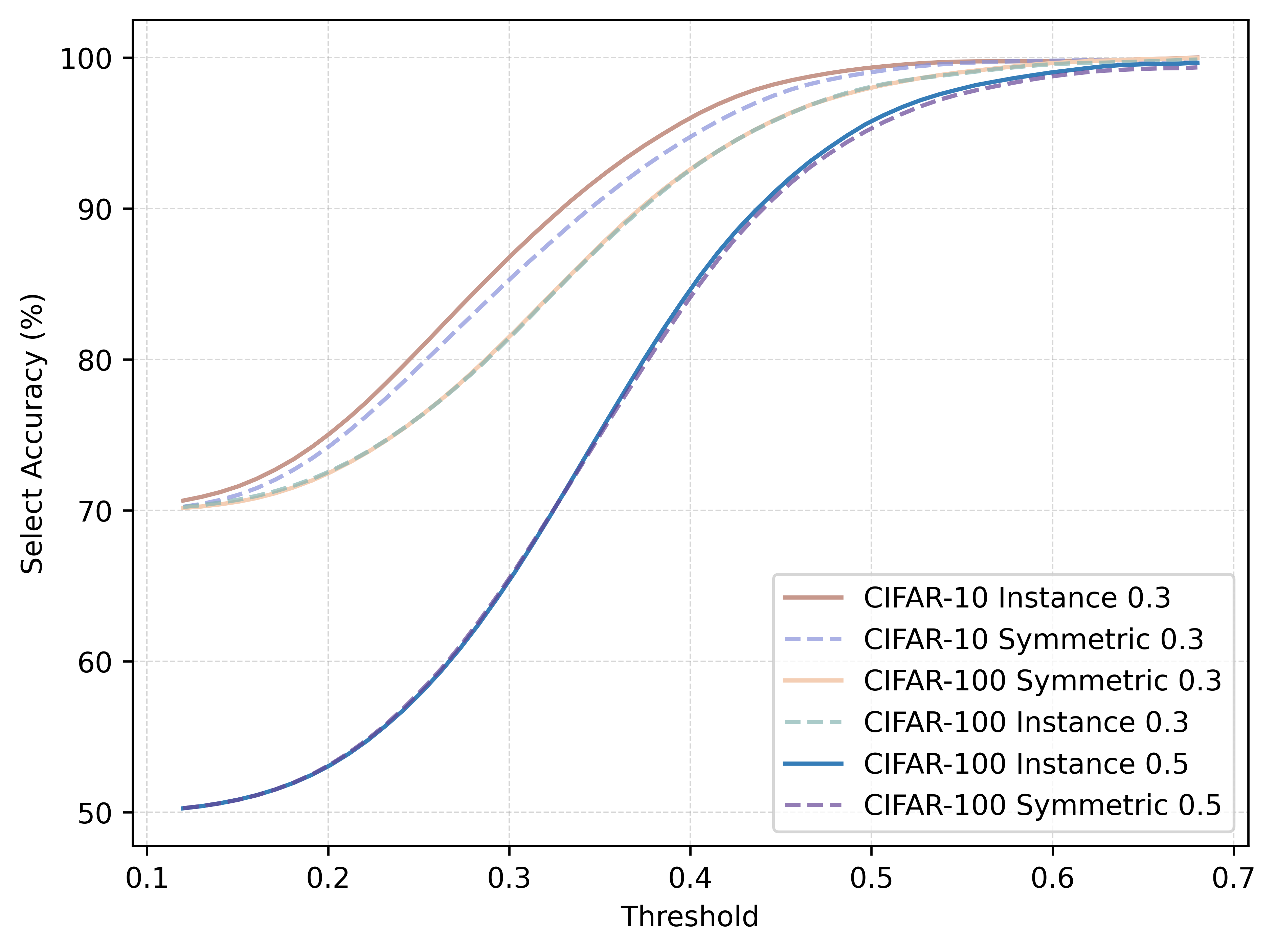}
            \caption{}
            \label{parameter_comparison}
        \end{subfigure}\hfill
        \begin{subfigure}{0.49\textwidth}
            \centering
            \includegraphics[width=\linewidth]{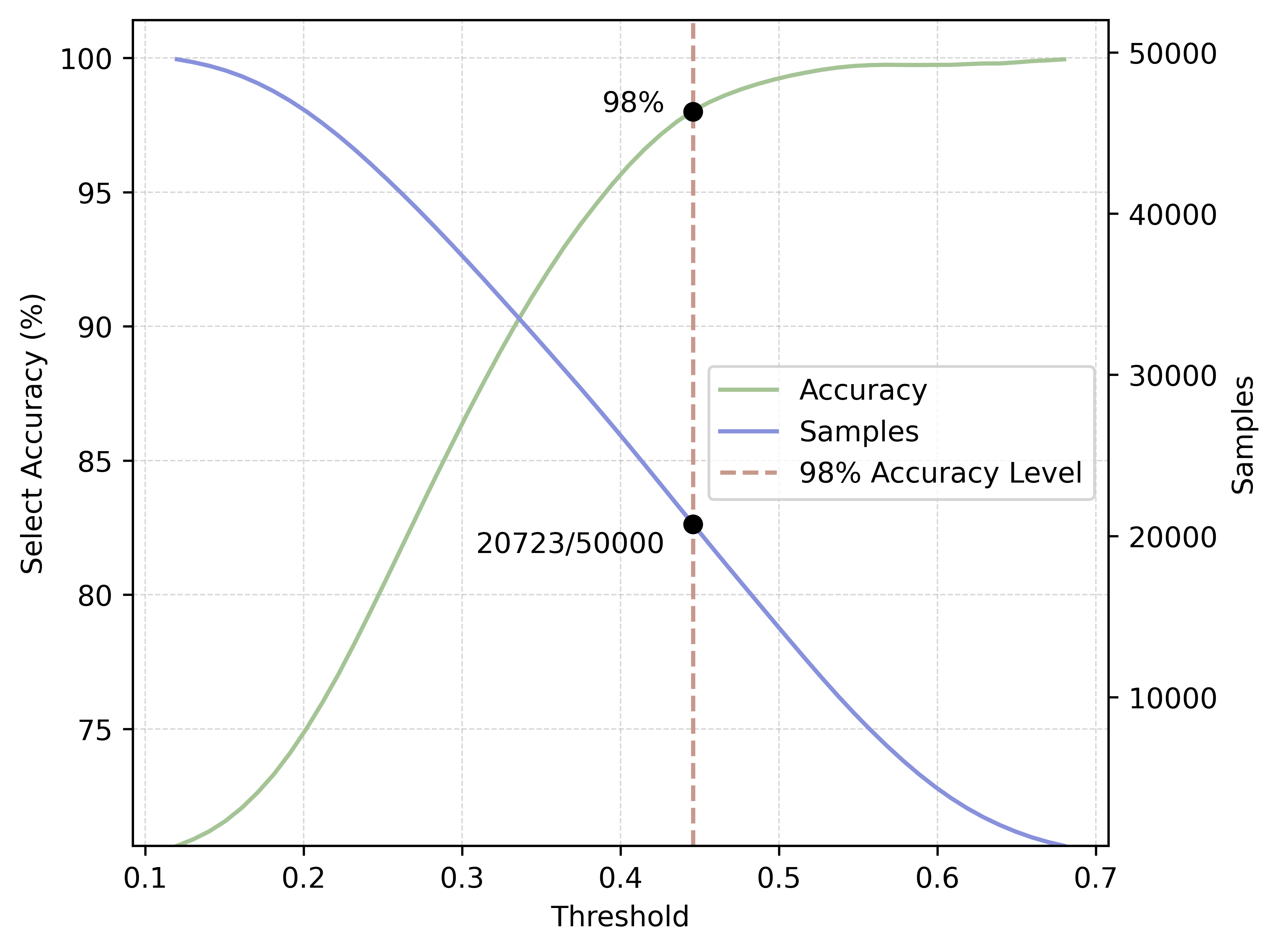}
            \caption{}
            \label{parameter_sensitivity}
        \end{subfigure}\hfill
        \caption{(a) Comparison of the effect of the threshold (\(\tau\)) on selection accuracy at different settings of 30\% noise rate. (b) Evaluation of thresholding effects on the quality and quantity of sample selection under 30\% instance-dependent noise on CIFAR-10.}
    \end{figure*}
    Figure~\ref{parameter_comparison} shows a consistent trend of selection accuracy as \(\tau\) varies across the evaluated datasets and noise settings, with the operating range of \(\tau\) primarily affected by the noise rate. The performance gap between CIFAR-10 and CIFAR-100 may be attributed to the higher intra-class similarity in CIFAR-100 due to its 20 superclasses, which increases the classification complexity. In practical use, an approximate noise-rate estimate (e.g., from a small validation set \cite{yu2018efficient}) can guide a lightweight calibration sweep to locate a plateau region where selection accuracy varies slowly with \(\tau\). Furthermore, Figure~\ref{parameter_sensitivity} illustrates the precision--coverage trade-off induced by thresholding: increasing \(\tau\) typically improves selection precision but retains fewer original instances. Since the downstream classifier is trained on the refined set, \(\tau\) shapes the learning signal through both the effective noise level and the available training data, and choosing \(\tau\) near the plateau provides a stable operating point under the evaluated settings.

    \subsection{Runtime Analysis}
    \label{runtime}

    \begin{table*}[ht]
        \centering
        \caption{Comparison of runtime at different settings using an NVIDIA V100-SXM2.}
        \footnotesize
        \begin{tabular}{ccccc}
            \toprule
            \multirow{3}{*}{Image resolution} & \multicolumn{3}{c}{CIFAR-10} & Clothing1M (est.) \\
            \cmidrule(r){2-5}
            & batch size-1 & batch size-8 & batch size-16 & batch size-16 \\
            \midrule
            \(256 \times 256\) & 31.5         & 5.5          & 4.5           & 129.5         \\
            \midrule
            \(512 \times 512\) & 35.1         & 18.5         & 17.2          & 504.5         \\
            \midrule
            \(768 \times 768\) & 68.5         & 51.9         & \cancel{}     & \cancel{}     \\
            \bottomrule
        \end{tabular}
        \label{v100}
    \end{table*}

    \begin{table}[ht]
        \centering
        \footnotesize
        \caption{Comparison of runtime at different computing performance.}
        \begin{tabular}{ccc}
            \toprule
            GPU       & CIFAR-10 & Clothing1M (est.) \\
            \midrule

            V100-SXM2 & 18.5     & 504.5             \\
            \midrule
            RTX 4090  & 8.5      & 338.2             \\
            \bottomrule
        \end{tabular}
        \label{4090}
    \end{table}

    The efficiency of EchoMod is significantly influenced by several factors, including the choice of the controllable generative model, the GPU's floating-point operations per second (FLOPS), the resolution of generated images, batch size, GPU memory capacity, floating-point precision, and the number of diffusion steps if a diffusion model is used. In this study, we employ an NVIDIA V100-SXM2 with 32GB of VRAM, using ControlNet as the benchmark generative model, and apply mixed precision to assess the impacts of image size and batch size on runtime. Runtime is measured in GPU hours, representing the computational time required to perform tasks on a single GPU. As the number of GPUs increases, wall-clock time may be reduced in practice, however, the scaling depends on system factors (e.g., I/O and batching) and is not necessarily linear. Our experiments are conducted on the CIFAR-10 dataset, and we also provide a rough runtime extrapolation for processing the Clothing1M dataset under the same generation setup and GPU configuration. Table~\ref{v100} demonstrates that increasing the batch size and reducing the image resolution both significantly impact runtime. We did not conduct tests with image resolution at \(768 \times 768\) and a batch size of 16 due to GPU memory constraints. Additionally, in Table~\ref{4090}, we compare the effects of different computing performance on runtime. We conducted tests on two different GPUs with an image resolution of \(512 \times 512\) and a batch size of 8. The NVIDIA V100-SXM2-32GB offers a half-precision compute capability of 125 Tensor TFLOPS and a single-precision capability of 15.7 TFLOPS. In contrast, the more powerful NVIDIA RTX 4090-24GB GPU provides 165.2 Tensor TFLOPS in half-precision and 82.58 TFLOPS in single-precision. We use ControlNet as the benchmark generative model for a standard and reproducible instantiation, and the flexibility of EchoMod allows the pipeline to benefit from established diffusion acceleration techniques and more efficient controllable-model variants. Efficient schedulers and memory-management techniques such as CPU offloading have been shown to improve throughput for diffusion pipelines under specific hardware and generation settings~\cite{ultra_fast_controlnet}. More efficient ControlNet architectures have also been proposed to reduce computation while retaining controllability~\cite{zavadski2023controlnet}. These results indicate that such optimizations are compatible with our pipeline design, however, we do not reproduce them in this paper and thus refrain from reporting a universal speedup factor. Accordingly, any runtime projection on Clothing1M should be interpreted as a rough estimate that depends on the GPU model, floating-point precision, image resolution, batch size, and the number of diffusion steps.

    \section{Conclusion}
    \label{conclusion}

    This work introduces a new perspective on learning with noisy labels by bridging generative and discriminative paradigms. Instead of correcting labels, EchoAlign treats noisy annotations as accurate and modifies instances to align with them, supported by theoretical motivation and empirical results on the evaluated benchmarks. By integrating controllable generative models with principled sample selection, EchoAlign addresses the dual challenges of characteristic and distribution shifts, enabling robust learning even under severe noise. Our results across diverse benchmarks demonstrate that generative--discriminative interplay can offer a fundamentally more resilient route to handling noisy supervision. While our evaluation focuses on these benchmarks and a single CGM instantiation, the results suggest that instance modification coupled with similarity-guided selection is a promising direction for noisy supervision. A systematic study across larger datasets, backbones, and alternative CGMs is an important next step.

    \section*{Acknowledgements}
    This work is supported by the National Natural Science Foundation of China [U23A20389, 62176139] and by the Qilu Young Scholars Program of Shandong University. The authors would also like to express sincere gratitude to Prof. Tongliang Liu for his invaluable guidance and support throughout this research.
    \appendix

    \section*{Appendices}

    \subsection*{Appendix A}
    \label{Proof}

    To provide a comprehensive proof of the theorem regarding the effectiveness of instance modification in learning from noisy labels, we will assume the definitions and setup described in the theorem statement. We will address each component of the theorem, demonstrating how the instance modification approach theoretically leads to improvements in alignment, error reduction, estimation stability, and generalization.

    \textbf{Notation and sufficient conditions.}

    Throughout Appendix A we treat Theorem~\ref{theorem1} as a sufficient-condition result. Let $X' = T(X,\tilde{Y};\theta)$ be the modified instance. We use the following sufficient conditions:
    \begin{itemize}
        \item A1. \emph{Label-alignment improvement:}
        $H(\tilde{Y}\mid X') \le H(\tilde{Y}\mid X)$.

        \item A2. \emph{Preservation of task-relevant signal (squared-loss form):}
        $\mathbb{E}[\operatorname{Var}(Y\mid X')] \allowbreak\le
        \mathbb{E}[\operatorname{Var}(Y\mid X)]$,
        i.e., the Bayes risk under squared loss using $X'$ is no larger than that using $X$.

        \item A3. \emph{Controlled shift (non-expansive radius):}
        for samples $\{(x_i,\tilde{y}_i)\}_{i=1}^n$ and $x_i' = T(x_i,\tilde{y}_i;\theta)$, the feature radius does not increase:%
        \par
        $\max_{1\le i\le n}\|x_i'\|_2 \allowbreak\le
        \max_{1\le i\le n}\|x_i\|_2$.
        Optionally, one sufficient explanation is that $T(\cdot,\tilde{Y};\theta)$ is $L$-Lipschitz in $X$ with $L\le 1$ together with an appropriate anchoring condition.

        \item A4. \emph{Stability-friendly conditioning (linear illustration):}
        in the linear illustration, let $X\in\mathbb{R}^{n\times d}$ and $X'\in\mathbb{R}^{n\times d}$ be the design matrices formed by stacking training samples row-wise (full column rank assumed). In the relevant subspace, the conditioning is no worse after modification, e.g.,%
        \linebreak[1]
        $\lambda_{\min}(X'^T X') \allowbreak\ge \lambda_{\min}(X^T X)$.
    \end{itemize}

    \textit{Proof.} We prove each component of Theorem~\ref{theorem1} regarding the effectiveness of instance modification as follows:

    \textbf{1. Alignment:}

    \textit{\textbf{Claim:}} The mutual information between $X'$ and $\tilde{Y}$, $ I(X'; \tilde{Y}) $, is no smaller than $ I(X; \tilde{Y}) $, indicating better alignment of modified instances with their noisy labels.

    \textit{\textbf{Definitions and Assumptions:}}

    Let $ X \in \mathbb{R}^d $ be the original instances with distribution $ P_X $ and $ \tilde{Y} \in \mathcal{Y} $ be the noisy labels, where $ \mathcal{Y} $ is the label space. The modified instances are defined as $ X' = T(X, \tilde{Y}; \theta) \in \mathbb{R}^d $, where $ T $ is a transformation function parameterized by $ \theta $, designed to improve alignment between $ X' $ and $ \tilde{Y} $.

    We make the following assumptions for this component:

    \begin{itemize}
        \item \textbf{S0.} \textbf{Identity is included}: There exists $\theta_{0}$ such that $T(X,\tilde{Y};\theta_{0}) = X$.
        \item \textbf{A1.} \textbf{Label-alignment improvement}: $H(\tilde{Y}\mid X') \le H(\tilde{Y}\mid X)$, which follows from choosing $\theta^{*}$ that minimizes $H(\tilde{Y}\mid T(X,\tilde{Y};\theta))$ over a hypothesis class that includes $\theta_{0}$.
    \end{itemize}

    \textit{\textbf{Goal:}}

    Our aim is to prove that the mutual information between $ X' $ and $ \tilde{Y} $ is greater than or equal to that between $ X $ and $ \tilde{Y} $: $$ I(X'; \tilde{Y}) \geq I(X; \tilde{Y}), $$ where the mutual information is defined as: $$ I(X; \tilde{Y}) = H(\tilde{Y}) - H(\tilde{Y} \mid X). $$

    \begin{proof}
        By definition, choosing $\theta^{*}$ to minimize \\$H(\tilde{Y}\mid T(X,\tilde{Y};\theta))$ is equivalent to maximizing \\$I(T(X,\tilde{Y};\theta);\tilde{Y}) = H(\tilde{Y})-H(\tilde{Y}\mid T(X,\tilde{Y};\theta))$ over $\theta$. We begin by expressing the mutual information between $ X $ (or $ X' $) and $ \tilde{Y} $:
        \begin{align*}
            I(X;\tilde{Y})  &= H(\tilde{Y}) - H(\tilde{Y}\mid X), \\
            I(X';\tilde{Y}) &= H(\tilde{Y}) - H(\tilde{Y}\mid X').
        \end{align*}
        The difference in mutual information is then:
        \begin{align*}
            \Delta I &= I(X'; \tilde{Y}) - I(X; \tilde{Y}) \\
            &= \left[ H(\tilde{Y}) - H(\tilde{Y} \mid X') \right] - \left[ H(\tilde{Y}) - H(\tilde{Y} \mid X) \right] \\
            &= H(\tilde{Y} \mid X) - H(\tilde{Y} \mid X').
        \end{align*}
        According to Assumption \textbf{A1}, the transformation $ T $ reduces the conditional entropy of $ \tilde{Y} $ given the features, so: $$ H(\tilde{Y} \mid X') \leq H(\tilde{Y} \mid X). $$ Therefore, the difference $ \Delta I $ is non-negative: $$ \Delta I = H(\tilde{Y} \mid X) - H(\tilde{Y} \mid X') \geq 0. $$ This implies that: $$ I(X'; \tilde{Y}) \geq I(X; \tilde{Y}). $$ Thus, the mutual information between the modified instances $ X' $ and the noisy labels $ \tilde{Y} $ is greater than or equal to that between the original instances $ X $ and $ \tilde{Y} $, indicating improved alignment between $ X' $ and $ \tilde{Y} $.

    \end{proof}

    \textbf{2. Error Reduction:}

    \textit{\textbf{Claim}}: Under squared loss, let \(f_{X'}^{*}(x')=\mathbb{E}[Y\mid X'=x']\) denote the population-optimal predictor on the modified instances. If the modification does not remove predictive information about \(Y\) and the induced distribution shift is controlled, then the achievable expected prediction error using \(X'\), \\\( \mathbb{E}_{X', Y}[(Y - f_{X'}^{*}(X'))^2] \), is no larger than that using \(X\).

    \textit{\textbf{Definitions and Assumptions:}}

    Let $ X \in \mathbb{R}^d $ be the original instances with distribution $ P_X $, $ \tilde{Y} \in \mathbb{R} $ be the noisy labels, and $ Y \in \mathbb{R} $ be the true labels. The modified instances are defined as $ X' = T(X, \tilde{Y}; \theta) \in \mathbb{R}^d $, where $ T $ is a transformation designed to improve alignment between $ X' $ and $ \tilde{Y} $ while preserving essential predictive information about $ Y $.

    We consider two models:

    \begin{itemize}
        \item $ f_X: \mathbb{R}^d \rightarrow \mathbb{R} $, trained on $ (X, \tilde{Y}) $.
        \item $ f_{X'}: \mathbb{R}^d \rightarrow \mathbb{R} $, trained on $ (X', \tilde{Y}) $.
    \end{itemize}
    The loss function $ L: \mathbb{R} \times \mathbb{R} \rightarrow \mathbb{R}_{\geq 0} $ is assumed to be convex and differentiable with respect to its second argument (e.g., squared loss $ L(y, \hat{y}) = (y - \hat{y})^2 $).

    We invoke the sufficient condition \textbf{A2} stated at the beginning of Appendix~A, i.e., $\mathbb{E}[\operatorname{Var}(Y\mid X')] \le \mathbb{E}[\operatorname{Var}(Y\mid X)]$. When needed, the controlled-shift intuition in \textbf{A3} is used to justify that finite-sample learning on $X'$ can transfer to test data.

    \textit{\textbf{Goal:}}

    Our aim is to show that using the modified representation does not increase the population risk under squared loss, i.e., $R_{X'}^{*} \le R_{X}^{*}$, where
    \begin{align*}
        R_{X}^{*} &\triangleq \inf_{f}\ \mathbb{E}_{X, Y}\!\left[(Y-f(X))^2\right], \\
        R_{X'}^{*} &\triangleq \inf_{f}\ \mathbb{E}_{X', Y}\!\left[(Y-f(X'))^2\right].
    \end{align*}

    \begin{proof}
        Under squared loss, for any representation $U$ (either $X$ or $X'$) and any predictor $f$, the risk admits the standard decomposition
        \[
            \mathbb{E}\big[(Y-f(U))^2\big]
            =
            \mathbb{E}\big[\operatorname{Var}(Y\mid U)\big]
            +
            \mathbb{E}\big[(\mathbb{E}[Y\mid U]-f(U))^2\big].
        \]
        Therefore, the population-optimal predictor is $f_U^{*}(u)=\mathbb{E}[Y\mid U=u]$, and the minimal achievable risk using $U$ equals
        \[
            R_U^{*}\triangleq \inf_f \mathbb{E}\big[(Y-f(U))^2\big] = \mathbb{E}\big[\operatorname{Var}(Y\mid U)\big].
        \]
        If instance modification makes $Y$ no harder to predict from the representation, i.e.,
        \[
            \mathbb{E}\big[\operatorname{Var}(Y\mid X')\big] \le \mathbb{E}\big[\operatorname{Var}(Y\mid X)\big],
        \]
        then it follows immediately that $R_{X'}^{*}\le R_X^{*}$, hence $\mathbb{E}[(Y-f_{X'}^{*}(X'))^2] \le \mathbb{E}[(Y-f_X^{*}(X))^2]$.
        Finally, when the induced distribution shift is controlled, standard generalization arguments imply that learning on $X'$ can inherit this advantage at finite sample sizes.
    \end{proof}

    \textbf{3. Estimation Stability:}

    \textit{\textbf{Claim:}} In a linear regression illustration, if the modified design is no worse conditioned in the relevant subspace (e.g., \(\lambda_{\min}(X'^T X') \ge \lambda_{\min}(X^T X)\)) and the modification is non-expansive in norm (e.g., \(\|x'\|_2 \le \|x\|_2\)), then the predictive variance using \(X'\) is reduced compared to using \(X\), leading to more stable predictions.

    \textit{Formulation (linear illustration).}
    Consider a fixed-design homoskedastic linear regression with response vector $y\in\mathbb{R}^n$ and design matrix $U\in\mathbb{R}^{n\times d}$, where $U=X$ for original features and $U=X'$ for modified features:
    $$ y = U\beta + \epsilon,\qquad \mathbb{E}[\epsilon]=0,\ \ \operatorname{Cov}(\epsilon)=\sigma^2 I. $$
    For a test input $u\in\mathbb{R}^d$, the prediction is $f_U(u)=u^T\hat{\beta}_U$.

    \textit{\textbf{Goal:}}
    To demonstrate that the variance of the estimator \( f_{X'} \) is lower than that of \( f_X \).

    \textit{Proof.}
    \begin{itemize}
        \item \textbf{Model Definitions and Assumptions}:
        Assume that both \( \beta_X \) and \( \beta_{X'} \) are obtained by ordinary least squares (OLS), implying that they minimize the respective mean squared errors. The variance of the estimator in OLS is inversely proportional to the Fisher information of the model, Fisher information matrix  \(I(\beta)\)  is represented as  \(X^T X\)  and  \(X'^T X'\), reflecting the variability of input features.

        \item \textbf{Variance of Estimators}:
        The covariance of the estimated coefficients under OLS can be expressed as:
        \begin{align*}
            \text{Cov}(\hat{\beta}_X) &= \sigma^2 (X^T X)^{-1} \\
            \text{Cov}(\hat{\beta}_{X'}) &= \sigma^2 (X'^T X')^{-1}
        \end{align*}

        where \( \sigma^2 \) is the variance of the error terms \( \epsilon_X \) and \( \epsilon_{X'} \), assumed equal for simplicity. The variance of the predicted values at any input \( x \) and its modified version \( x' \) is:
        \begin{align*}
            \text{Var}(f_X(x)) &= x^T \text{Cov}(\hat{\beta}_X) x = \sigma^2 x^T (X^T X)^{-1} x \\
            \text{Var}(f_{X'}(x')) &= x'^T \text{Cov}(\hat{\beta}_{X'}) x' = \sigma^2 x'^T (X'^T X')^{-1} x'
        \end{align*}

        \item \textbf{Comparative Analysis of Variance}:
        Under the homoskedastic linear illustration with full-column-rank design matrices, the OLS covariance satisfies
        \begin{align*}
            \mathrm{Cov}(\hat{\beta}_X \mid X)&=\sigma^{2}(X^{T}X)^{-1}, \\
            \mathrm{Cov}(\hat{\beta}_{X'} \mid X')&=\sigma^{2}(X'^{T}X')^{-1}.
        \end{align*}
        Therefore, for any test inputs $x$ and $x'$, the conditional prediction variances are
        \begin{align*}
            \mathrm{Var}(f_X(x)\mid X)&=\sigma^{2} x^{T}(X^{T}X)^{-1}x, \\
            \mathrm{Var}(f_{X'}(x')\mid X')&=\sigma^{2} x'^{T}(X'^{T}X')^{-1}x'.
        \end{align*}
        For any symmetric positive definite matrix $A$ and any vector $v$, we have the bound
        \[
            v^{T}A^{-1}v \le \frac{\|v\|_{2}^{2}}{\lambda_{\min}(A)}.
        \]
        Applying this inequality yields
        \begin{align*}
            \mathrm{Var}(f_X(x)\mid X)
            &\le \sigma^{2}\frac{\|x\|_{2}^{2}}{\lambda_{\min}(X^{T}X)}, \\
            \mathrm{Var}(f_{X'}(x')\mid X')
            &\le \sigma^{2}\frac{\|x'\|_{2}^{2}}{\lambda_{\min}(X'^{T}X')}.
        \end{align*}
        To obtain a rigorous stability comparison, we consider the worst-case conditional prediction variance over bounded test inputs. For any radius $r>0$, define
        \[
            V_U(r)\triangleq \sup_{\|u\|_2\le r}\ \mathrm{Var}(f_U(u)\mid U).
        \]
        Since
        \begin{align*}
            &\sup_{\|u\|_2\le r} u^T (U^T U)^{-1}u\\
            &= r^2 \lambda_{\max}((U^T U)^{-1})\\
            &= \frac{r^2}{\lambda_{\min}(U^T U)},
        \end{align*}
        we have
        \begin{align*}
            V_X(r)
            &= \sigma^2\frac{r^2}{\lambda_{\min}(X^T X)}, \\
            V_{X'}(r)
            &= \sigma^2\frac{r^2}{\lambda_{\min}(X'^T X')}.
        \end{align*}
        Under \textbf{A4}, $\lambda_{\min}(X'^T X') \ge \lambda_{\min}(X^T X)$, hence $V_{X'}(r)\le V_X(r)$ for the same radius $r$.

        \item \textbf{Estimation Stability}:
        The inequality above shows that, in the linear illustration, instance modification can reduce (or at least not increase) the worst-case conditional prediction variance over bounded test inputs when the effective design is not worsened. Concretely, this stability statement holds under the following sufficient conditions:
        \begin{enumerate}
            \item the homoskedastic linear model with full-column-rank design matrices so that $(X^{T}X)^{-1}$ and $(X'^{T}X')^{-1}$ are well-defined;
            \item $\lambda_{\min}(X'^{T}X') \ge \lambda_{\min}(X^{T}X)$ (A4), meaning the relevant conditioning is no worse after modification;
            \item $\|x'\|_{2}\le \|x\|_{2}$ (A3), meaning the modification does not increase the feature radius.
        \end{enumerate}
        These conditions formalize the non-expansive intuition used in our analysis and provide a sufficient explanation for why feature-level alignment can improve estimator stability in the presence of noisy supervision.
    \end{itemize}
    This linear analysis provides a sufficient-condition explanation that connects non-expansive modification and non-worsened conditioning to improved prediction stability, which motivates our design choice of controlling distribution shift while leveraging label-aligned instance modification.

    \hfill \qedsymbol

    \textbf{4. Generalization:}

    \textit{\textbf{Claim:}} Modifications in \(X'\) can lead to better generalization. In particular, under a non-expansive modification (e.g., \(\|x_i'\|_2 \le \|x_i\|_2\)), the complexity of common hypothesis classes on \(X'\) is no larger than that on \(X\), which yields a no-worse (and potentially tighter) generalization bound.

    \textit{\textbf{Setup:}}
    Let
    \begin{itemize}
        \item \( X \) denotes the original feature space and \( X' = T(X, \tilde{Y}; \theta) \) denotes the modified feature space, where \( T \) is a transformation (assumed to be Lipschitz continuous with Lipschitz constant \( L \leq 1 \)), and \( \theta \) is fixed, that optimizes some aspect of the data to better align with noisy labels \( \tilde{Y} \). In our EchoAlign framework, the transformation \( T \) is instantiated using controllable generative models. The Lipschitz condition \( L \leq 1 \) is adopted here as a sufficient analytical condition for bounding the drift induced by instance modification, in practice, we do not explicitly enforce a tight Lipschitz constant for the generative model implementation. This non-expansive intuition aligns with our design goal of avoiding amplification of noise, and EchoSelect further mitigates distribution shift by retaining original instances.
        \item \(\mathcal{F}\) be the class of functions \(f : \mathcal{X} \to \mathbb{R}\) considered by the learning algorithm, where \(\mathcal{X}\) is either the space of \(X\) or \(X'\).
    \end{itemize}

    \textit{\textbf{Rademacher Complexity:}}
    Rademacher complexity measures the ability of a function class to fit random noise. The Rademacher complexity for the class of functions \( \mathcal{F} \) applied to the original features \( X \) is the modified features \( X' \) are defined respectively as:
    \begin{align*}
        \mathfrak{R}_n(\mathcal{F}_X) &= \mathbb{E}_{\sigma, X}\left[\sup_{f \in \mathcal{F}_X} \frac{1}{n} \sum_{i=1}^n \sigma_i f(X_i)\right] \\
        \mathfrak{R}_n(\mathcal{F}_{X'}) &= \mathbb{E}_{\sigma, X'}\left[\sup_{f \in \mathcal{F}_{X'}} \frac{1}{n} \sum_{i=1}^n \sigma_i f(X'_i)\right]
    \end{align*}

    \textit{\textbf{Generalization Bounds:}}
    Using these definitions, the generalization bounds for a Lipschitz continuous loss function \( l \) can be expressed for both feature sets. Assuming the same hypothesis class \( \mathcal{F} \), the bounds are:
    \begin{align*}
        \mathbb{E}[l(f(X), Y)] &\leq \frac{1}{n} \sum_{i=1}^n l(f(X_i), Y_i) + 2 \mathfrak{R}_n(\mathcal{F}_X) + c \\
        \mathbb{E}[l(f(X'), Y)] &\leq \frac{1}{n} \sum_{i=1}^n l(f(X'_i), Y_i) + 2 \mathfrak{R}_n(\mathcal{F}_{X'}) + c
    \end{align*}
    where \( c \) is a constant that depends on the complexity of the loss function.

    \textit{Impact of Instance Modification on Feature Space:}
    The transformation \( T \) is designed to adjust features in \( X \) to more effectively align with \( \tilde{Y} \), potentially reducing the variability of \( X \) that is irrelevant to predicting \( Y \). This transformation can:
    \begin{itemize}
        \item Increase the signal-to-noise ratio in \( X' \) compared to \( X \).
        \item Focus the variability in \( X' \) on aspects that are more predictive of \( Y \), based on the information contained in \( \tilde{Y} \).
    \end{itemize}

    \begin{proof}
        We provide a rigorous illustration for a standard hypothesis class whose Rademacher complexity explicitly depends on the feature radius. Consider the bounded-norm linear predictor class $\mathcal{F}_\lambda \triangleq \{x\mapsto w^{T}x:\ \|w\|_{2}\le \lambda\}$. A standard empirical Rademacher bound states that for any sample $x_1,\ldots,x_n$,
        \begin{align*}
            \widehat{\mathfrak{R}}_n(\mathcal{F}_\lambda)
            &\le \frac{\lambda}{n}\sqrt{\sum_{i=1}^{n}\|x_i\|_2^{2}}
            \le \frac{\lambda\, C_X}{\sqrt{n}}, \\
            C_X
            &\triangleq \max_{1\le i\le n}\|x_i\|_2.
        \end{align*}
        Under the non-expansive condition in A3 (e.g., \\$\|x_i'\|_2\le \|x_i\|_2$ for all $i$), the feature radius does not increase, hence the Rademacher complexity bound for the modified sample is no larger than that for the original sample.

        Since standard symmetrization arguments bound the expected uniform deviation by (a constant multiple of) the expected Rademacher average, a no-larger Rademacher complexity implies a no-looser generalization control for models trained on the modified representation. Therefore, under A3, instance modification can improve generalization in this illustrative setting by not increasing, and potentially decreasing, the effective complexity term.
    \end{proof}

    \textbf{An illustrative sufficient construction for a non-expansive transformation:}

    We present an illustrative sufficient construction that guarantees a non-expansive transformation, which matches the regularity condition used in the generalization discussion above. This construction is for theoretical illustration and is not enforced by our ControlNet-based implementation.

    We define the transformation $ T $ as a convex combination of the original instance $ X $ and an adjustment function $ \phi(X, \tilde{Y};\theta) $ that incorporates the influence of the noisy label:

    $$ T(X, \tilde{Y}; \theta) = (1 - \alpha) X + \alpha \phi(X, \tilde{Y}; \theta), $$ where $ \alpha \in [0, 1] $ is a parameter controlling the degree of modification, and $ \phi(X, \tilde{Y}; \theta) $ is designed to adjust $ X $ based on $ \tilde{Y} $.

    To ensure that $ T $ is Lipschitz continuous with $ L \leq 1 $, we require that $ \phi $ itself is Lipschitz continuous with Lipschitz constant $ L_\phi \leq 1 $. Under this condition, for any two instances $ X_1, X_2 \in \mathcal{X} $, we have:
    \begin{align*}
        &\| T(X_1, \tilde{Y}; \theta) - T(X_2, \tilde{Y}; \theta) \| \\
        &= \left\| (1 - \alpha)(X_1 - X_2) + \alpha \left( \phi(X_1, \tilde{Y}; \theta) - \phi(X_2, \tilde{Y}; \theta) \right) \right\| \\ &\leq (1 - \alpha) \| X_1 - X_2 \| + \alpha \| \phi(X_1, \tilde{Y}; \theta) - \phi(X_2, \tilde{Y}; \theta) \| \\ &\leq (1 - \alpha) \| X_1 - X_2 \| + \alpha L_\phi \| X_1 - X_2 \| \\ &= \left( (1 - \alpha) + \alpha L_\phi \right) \| X_1 - X_2 \|.
    \end{align*}
    Since $ L_\phi \leq 1 $ and $ \alpha \in [0, 1] $, we have: \[ (1 - \alpha) + \alpha L_\phi \leq (1 - \alpha) + \alpha = 1, \] which means that the Lipschitz constant $ L $ of $ T $ satisfies $ L \leq 1 $.

    To ensure that $ \phi $ has Lipschitz constant $ L_\phi \leq 1 $, we can design $ \phi $ using various techniques:

    \begin{itemize}
        \item \textbf{Spectral Normalization:}
        Spectral normalization constrains the spectral norm (largest singular value) of each linear layer in the neural network implementing $ \phi $ to be at most 1~\cite{miyato2018spectralnormalizationgenerativeadversarial}. By normalizing the weight matrices $ W $ of the layers such that: $$ \| W \|_2 = \sigma_{\text{max}}(W) = 1, $$ we ensure that the Lipschitz constant of each layer does not exceed 1. Since the Lipschitz constant of a composition of functions is \\bounded by the product of the individual Lipschitz constants, and each layer has $ L_i \leq 1 $, the overall Lipschitz constant of $ \phi $ satisfies $ L_\phi \leq 1 $.
        \item \textbf{Gradient Penalties:}
        Incorporating gradient penalties into the training of $ \phi $ encourages the network to have controlled Lipschitz continuity~\cite{gulrajani2017improved}. We add a regularization term to the loss function: $$ \mathcal{L}_{\text{GP}} = \lambda \, \mathbb{E}_{X, \tilde{Y}} \left[ \left( \left\| \nabla_X \phi(X, \tilde{Y}; \theta) \right\|_2 - 1 \right)^2 \right], $$ where $ \lambda > 0 $ is a penalty coefficient. Minimizing $ \mathcal{L}_{\text{GP}} $ enforces the gradient norms of $ \phi $ to be close to 1, ensuring $ L_\phi \leq 1 $.
        \item \textbf{Contractive Autoencoders:}
        Designing $ \phi $ as a contractive autoencoder~\cite{Rifai2011ContractiveAE} involves adding a contraction penalty to the loss function: $$ \mathcal{L}_{\text{CAE}} = \mathbb{E}_{X} \left[ \| X - \phi(X, \tilde{Y}; \theta) \|_2^2 + \lambda\ \left\| \frac{\partial \phi(X, \tilde{Y}; \theta)}{\partial X} \right\|_F^2 \right], $$ where $ \| \cdot \|_F $ denotes the Frobenius norm, and $ \lambda > 0 $ controls the penalty strength. This penalizes large derivatives, encouraging $ \phi $ to be contractive and thus Lipschitz continuous with $ L_\phi \leq 1 $.
    \end{itemize}

    \subsection*{Appendix B: EchoMod generation setup checklist}
    \label{app:gen_checklist}

    \begin{table}[htbp]
        \centering
        \footnotesize
        \setlength{\tabcolsep}{4pt}
        \renewcommand{\arraystretch}{1.1}
        \caption{EchoMod generation setup checklist.}
        \begin{tabular}{@{}ll@{}}
            \hline
            \parbox[t]{0.32\linewidth}{\raggedright Component}          & \parbox[t]{0.62\linewidth}{\raggedright Setting}                                                                                                                                                 \\
            \hline
            \parbox[t]{0.32\linewidth}{\raggedright Model}              & \parbox[t]{0.62\linewidth}{\raggedright Stable Diffusion v1.5 + ControlNet v1.1 Canny.}                                                                                                          \\
            \parbox[t]{0.32\linewidth}{\raggedright Control signal}     & \parbox[t]{0.62\linewidth}{\raggedright Canny edges computed at \(R_{\mathrm{det}}=512\) with \((\tau_{\mathrm{low}},\tau_{\mathrm{high}})=(75,75)\), then resized to \(R_{\mathrm{img}}=512\).} \\
            \parbox[t]{0.32\linewidth}{\raggedright Prompting}          & \parbox[t]{0.62\linewidth}{\raggedright Prompt template: \texttt{a photo of \{\(\tilde{y}\)\}}.}                                                                                                 \\
            \parbox[t]{0.32\linewidth}{\raggedright Sampling}           & \parbox[t]{0.62\linewidth}{\raggedright DDIM with steps \(S=20\), \(\eta=1.0\), guidance scale \(g=9.0\).}                                                                                       \\
            \parbox[t]{0.32\linewidth}{\raggedright Control strength}   & \parbox[t]{0.62\linewidth}{\raggedright Default: constant \(s=1.0\) across 13 control layers; if guess mode: \(s\cdot 0.825^{(12-i)}\) for \(i=0,\dots,12\).}                                    \\
            \parbox[t]{0.32\linewidth}{\raggedright Randomness}         & \parbox[t]{0.62\linewidth}{\raggedright Fixed seed reported in experiments (if \texttt{seed}=-1, sampled uniformly from \([0,65535]\)).}                                                         \\
            \parbox[t]{0.32\linewidth}{\raggedright Output and scoring} & \parbox[t]{0.62\linewidth}{\raggedright Generate \texttt{num\_samples}=1 images per input.}                                                                                                      \\
            \hline
        \end{tabular}
    \end{table}

    \bibliographystyle{fcs}
    \bibliography{ref}

@inproceedings{Zhuang2023DyGenLF,
  title={DyGen: Learning from Noisy Labels via Dynamics-Enhanced Generative Modeling},
  author={Yuchen Zhuang and Yue Yu and Lingkai Kong and Xiang Chen and Chao Zhang},
  booktitle={Proceedings of the 29th ACM SIGKDD Conference on Knowledge Discovery and Data Mining},
  year={2023},
  doi={10.1145/3580305.3599318}
}

@article{Chen2023LabelRetrievalAugmentedDM,
  title={Label-Retrieval-Augmented Diffusion Models for Learning from Noisy Labels},
  author={Jian Chen and Ruiyi Zhang and Tong Yu and Rohan Sharma and Zhiqiang Xu and Tong Sun and Changyou Chen},
  journal={Proceedings of the Advances in Neural Information Processing Systems},
  year={2023}
}

@article{Du2023SequentialRW,
  title={Sequential Recommendation with Diffusion Models},
  author={Hanwen Du and Huanhuan Yuan and Zhen Huang and Pengpeng Zhao and Xiaofang Zhou},
  journal={arXiv preprint arXiv:2304.04541},
  year={2023}
}

@inproceedings{Wang2023ConditionalDD,
  title={Conditional Denoising Diffusion for Sequential Recommendation},
  author={Yu Wang and Zhiwei Liu and Liangwei Yang and Philip S. Yu},
  booktitle="Advances in Knowledge Discovery and Data Mining",
  year={2024}
}

@inproceedings{Franceschi2023UnifyingGA,
  author    = {Franceschi, Jean-Yves and Gartrell, Mike and Dos Santos, Ludovic and Issenhuth, Thibaut and de B{\'e}zenac, Emmanuel and Chen, Micka{\"e}l and Rakotomamonjy, Alain},
  title     = {Unifying GANs and Score-Based Diffusion as Generative Particle Models},
  booktitle = {Proceedings of the Advances in Neural Information Processing Systems},
  year      = {2023}
}

@article{han2022towards,
  title={Towards accurate and robust domain adaptation under multiple noisy environments},
  author={Han, Zhongyi and Gui, Xian-Jin and Sun, Haoliang and Yin, Yilong and Li, Shuo},
  journal={IEEE Transactions on Pattern Analysis and Machine Intelligence},
  volume={45},
  number={5},
  pages={6460--6479},
  year={2022},
  publisher={IEEE}
}

@article{han2022learning,
  title={Learning transferable parameters for unsupervised domain adaptation},
  author={Han, Zhongyi and Sun, Haoliang and Yin, Yilong},
  journal={IEEE Transactions on Image Processing},
  volume={31},
  pages={6424--6439},
  year={2022},
  publisher={IEEE}
}

@inproceedings{zhang2023adding,
  title={Adding conditional control to text-to-image diffusion models},
  author={Zhang, Lvmin and Rao, Anyi and Agrawala, Maneesh},
  booktitle={Proceedings of the IEEE/CVF International Conference on Computer Vision},
  pages={3836--3847},
  year={2023}
}

@article{kingma2021variational,
  title={Variational diffusion models},
  author={Kingma, Diederik and Salimans, Tim and Poole, Ben and Ho, Jonathan},
  journal={Proceedings of the Advances in Neural Information Processing Systems},
  volume={34},
  pages={21696--21707},
  year={2021}
}

@article{chen2023improving,
  title={Improving In-Context Learning in Diffusion Models with Visual Context-Modulated Prompts},
  author={Chen, Tianqi and Liu, Yongfei and Wang, Zhendong and Yuan, Jianbo and You, Quanzeng and Yang, Hongxia and Zhou, Mingyuan},
  journal={arXiv preprint arXiv:2312.01408},
  year={2023}
}

@inproceedings{tanaka2018joint,
  title={Joint optimization framework for learning with noisy labels},
  author={Tanaka, Daiki and Ikami, Daiki and Yamasaki, Toshihiko and Aizawa, Kiyoharu},
  booktitle={Proceedings of the IEEE/CVF Conference on Computer Vision and Pattern Recognition},
  pages={5552--5560},
  year={2018}
}

@inproceedings{Wu2020TopoFilter,
  author    = {Pengxiang Wu and Songzhu Zheng and
               Mayank Goswami and Dimitris N. Metaxas and Chao Chen},
  title     = {A Topological Filter for Learning with Label Noise},
  booktitle = {Proceedings of the Advances in Neural Information Processing Systems},
  pages={21382--21393},
  year      = {2020},
}

@inproceedings{xia2021CDR,
  title={Robust early-learning: Hindering the memorization of noisy labels},
  author={Xiaobo Xia and Tongliang Liu and Bo Han and Chen Gong and Nannan Wang and Zongyuan Ge and Yi Chang},
  booktitle={Proceedings of the International Conference on Learning Representations},
  year={2021}
}

@inproceedings{Yang2022bayeslabel,
  author    = {Shuo Yang and Erkun Yang and Bo Han and Yang Liu and
               Min Xu and Gang Niu and Tongliang Liu},
  title     = {Estimating Instance-dependent Bayes-label Transition Matrix using a Deep Neural Network},
  booktitle = {Proceedings of the International Conference on Machine Learning},
  pages     = {25302--25312},
  year      = {2022},
}

@inproceedings{Zhang2017MixUp,
  author    = {Hongyi Zhang and
               Moustapha Ciss{\'{e}} and
               Yann N. Dauphin and
               David Lopez{-}Paz},
  title     = {mixup: Beyond Empirical Risk Minimization},
  booktitle = {Proceedings of the International Conference on Learning Representations},
  year      = {2018},
}

@inproceedings{Li2020DivideMix,
  title     = {DivideMix: Learning with Noisy Labels as Semi-supervised Learning},
  author    = {Junnan Li and Richard Socher and Steven C. H. Hoi},
  booktitle = {Proceedings of the International Conference on Learning Representations},
  year      = {2020},
}

@inproceedings{Kim2021FINE,
  author    = {Taehyeon Kim and Jongwoo Ko and Sangwook Cho and
               Jinhwan Choi and Se{-}Young Yun},
  title     = {{FINE} Samples for Learning with Noisy Labels},
  booktitle = {Proceedings of the Advances in Neural Information Processing Systems},
  pages     = {24137--24149},
  year      = {2021},
}

@inproceedings{Liu2020ELR,
  author    = {Sheng Liu and Jonathan Niles{-}Weed and Narges Razavian and Carlos Fernandez{-}Granda},
  title     = {Early-Learning Regularization Prevents Memorization of Noisy Labels},
  booktitle = {Proceedings of the Advances in Neural Information Processing Systems},
  pages     = {20331--20342},
  year      = {2020},
}

@inproceedings{Bai2021PES,
  title={Understanding and improving early stopping for learning with noisy labels},
  author={Bai, Yingbin and Yang, Erkun and Han, Bo and Yang, Yanhua and Li, Jiatong and Mao, Yinian and Niu, Gang and Liu, Tongliang},
  booktitle={Proceedings of the Advances in Neural Information Processing Systems},
  pages={24392--24403},
  year={2021}
}

@article{xia2023regularly,
  title={Regularly truncated m-estimators for learning with noisy labels},
  author={Xia, Xiaobo and Lu, Pengqian and Gong, Chen and Han, Bo and Yu, Jun and Liu, Tongliang},
  journal={IEEE Transactions on Pattern Analysis and Machine Intelligence},
  year={2023},
  publisher={IEEE}
}

@inproceedings{ma2020normalized,
  title={Normalized Loss Functions for Deep Learning with Noisy Labels},
  author={Ma, Xingjun and Huang, Hanxun and Wang, Yisen and Romano, Simone and Erfani, Sarah and Bailey, James},
  booktitle={Proceedings of the International Conference on Machine Learning},
  year={2020}
}

@inproceedings{menon2019can,
  title={Can gradient clipping mitigate label noise?},
  author={Menon, Aditya Krishna and Rawat, Ankit Singh and Reddi, Sashank J and Kumar, Sanjiv},
  booktitle={Proceedings of the International Conference on Learning Representations},
  year={2019}
}

@article{zhou2023asymmetric,
  title={Asymmetric loss functions for noise-tolerant learning: Theory and applications},
  author={Zhou, Xiong and Liu, Xianming and Zhai, Deming and Jiang, Junjun and Ji, Xiangyang},
  journal={IEEE Transactions on Pattern Analysis and Machine Intelligence},
  year={2023},
  publisher={IEEE}
}

@inproceedings{radford2021learning,
  title={Learning transferable visual models from natural language supervision},
  author={Radford, Alec and Kim, Jong Wook and Hallacy, Chris and Ramesh, Aditya and Goh, Gabriel and Agarwal, Sandhini and Sastry, Girish and Askell, Amanda and Mishkin, Pamela and Clark, Jack and others},
  booktitle={International conference on machine learning},
  pages={8748--8763},
  year={2021},
  organization={PMLR}
}

@inproceedings{wei2023logitclip,
  title={Mitigating Memorization of Noisy Labels by Clipping the Model Prediction},
  author={Wei, Hongxin and Zhuang, Huiping and Xie, Renchunzi and Feng, Lei and Niu, Gang and An, Bo and Li, Yixuan},
  booktitle={International Conference on Machine Learning},
  year={2023},
  organization={PMLR}
}

@InProceedings{pmlr-v119-han20c,
  title = 	 {{SIGUA}: Forgetting May Make Learning with Noisy Labels More Robust},
  author =       {Han, Bo and Niu, Gang and Yu, Xingrui and Yao, Quanming and Xu, Miao and Tsang, Ivor and Sugiyama, Masashi},
  booktitle = 	 {International Conference on Machine Learning},
  pages = 	 {4006--4016},
  year = 	 {2020}
}

@inproceedings{xia2023combating,
  title={Combating Noisy Labels with Sample Selection by Mining High-Discrepancy Examples},
  author={Xia, Xiaobo and Han, Bo and Zhan, Yibing and Yu, Jun and Gong, Mingming and Gong, Chen and Liu, Tongliang},
  booktitle={Proceedings of the IEEE/CVF International Conference on Computer Vision},
  pages={1833--1843},
  year={2023}
}

@inproceedings{ICML2019_UnsupervisedLabelNoise,
  title={Unsupervised label noise modeling and loss correction},
  author={Arazo, Eric and Ortego, Diego and Albert, Paul and O’Connor, Noel and McGuinness, Kevin},
  booktitle={International conference on machine learning},
  pages={312--321},
  year={2019},
  organization={PMLR}
}

@inproceedings{chen2024catastrophic,
  title={On Catastrophic Inheritance of Large Foundation Models},
  author={Chen, Hao and Raj, Bhiksha and Xie, Xing and Wang, Jindong},
  booktitle={Data-centric Machine Learning Research},
  year={2024}
}

@inproceedings{Nguyen2020SELF,
  author    = {Duc Tam Nguyen and
               Chaithanya Kumar Mummadi and
               Thi{-}Phuong{-}Nhung Ngo and
               Thi Hoai Phuong Nguyen and
               Laura Beggel and
               Thomas Brox},
  title     = {{SELF:} Learning to Filter Noisy Labels with Self-Ensembling},
  booktitle = {Proceedings of the International Conference on Learning Representations},
  year      = {2020},
}

@inproceedings{han2018co,
  title={Co-teaching: Robust training of deep neural networks with extremely noisy labels},
  author={Han, Bo and Yao, Quanming and Yu, Xingrui and Niu, Gang and Xu, Miao and Hu, Weihua and Tsang, Ivor and Sugiyama, Masashi},
  booktitle={Proceedings of the Advances in Neural Information Processing Systems},
  pages={8527--8537},
  year={2018}
}

@inproceedings{xia2019anchor,
  title={Are Anchor Points Really Indispensable in Label-Noise Learning?},
  author={Xia, Xiaobo and Liu, Tongliang and Wang, Nannan and Han, Bo and Gong, Chen and Niu, Gang and Sugiyama, Masashi},
  booktitle={Proceedings of the Advances in Neural Information Processing Systems},
  pages={6835--6846},
  year={2019}
}

@inproceedings{Berthon2021ConfidenceScores,
  author    = {Antonin Berthon and
               Bo Han and
               Gang Niu and
               Tongliang Liu and
               Masashi Sugiyama},
  title     = {Confidence Scores Make Instance-dependent Label-noise Learning Possible},
  booktitle = {Proceedings of the International Conference on Machine Learning},
  series    = {Proceedings of Machine Learning Research},
  pages     = {825--836},
  year      = {2021},
}

@inproceedings{Lu2022NoiseAttention,
  author       = {Yangdi Lu and
                  Yang Bo and
                  Wenbo He},
  title        = {Noise Attention Learning: Enhancing Noise Robustness by Gradient Scaling},
  booktitle    = {Proceedings of the Advances in Neural Information Processing Systems},
  year         = {2022},
}

@inproceedings{Tan2019EfficientNet,
  author    = {Mingxing Tan and Quoc V. Le},
  title     = {EfficientNet: Rethinking Model Scaling for Convolutional Neural Networks},
  booktitle = {Proceedings of the International Conference on Machine Learning},
  pages     = {6105--6114},
  year      = {2019},
}

@inproceedings{Dosovitskiy2021AnImageWorth,
  author    = {Alexey Dosovitskiy and Lucas Beyer and Alexander Kolesnikov and Dirk Weissenborn and
               Xiaohua Zhai and Thomas Unterthiner and Mostafa Dehghani and Matthias Minderer and
               Georg Heigold and Sylvain Gelly and Jakob Uszkoreit and Neil Houlsby},
  title     = {An Image is Worth 16x16 Words: Transformers for Image Recognition at Scale},
  booktitle = {Proceedings of the International Conference on Learning Representations},
  year      = {2021},
}

@inproceedings{Yu2018WebCrawling,
  author    = {Xiyu Yu and Tongliang Liu and Mingming Gong and Dacheng Tao},
  title     = {Learning with Biased Complementary Labels},
  booktitle = {Proceedings of the European Conference on Computer Vision},
  pages     = {69--85},
  year      = {2018},
}

@inproceedings{Welinder2010crowdsourcing,
  author    = {Peter Welinder and Steve Branson and Serge J. Belongie and Pietro Perona},
  title     = {The Multidimensional Wisdom of Crowds},
  booktitle={Proceedings of the Advances in Neural Information Processing Systems},
  pages     = {2424--2432},
  year      = {2010},
}

@inproceedings{Arpit2017Look,
  author    = {Devansh Arpit and
               Stanislaw Jastrzebski and
               Nicolas Ballas and
               David Krueger and
               Emmanuel Bengio and
               Maxinder S. Kanwal and
               Tegan Maharaj and
               Asja Fischer and
               Aaron C. Courville and
               Yoshua Bengio and
               Simon Lacoste{-}Julien},
  title     = {A Closer Look at Memorization in Deep Networks},
  booktitle = {Proceedings of the International Conference on Machine Learning},
  pages     = {233--242},
  year      = {2017},
}

@inproceedings{
wei2022learning,
title={Learning with Noisy Labels Revisited: A Study Using Real-World Human Annotations},
author={Jiaheng Wei and Zhaowei Zhu and Hao Cheng and Tongliang Liu and Gang Niu and Yang Liu},
booktitle={Proceedings of the International Conference on Learning Representations},
year={2022}
}

@inproceedings{He2016ResNet,
  author    = {Kaiming He and Xiangyu Zhang and Shaoqing Ren and Jian Sun},
  title     = {Deep Residual Learning for Image Recognition},
  booktitle = {Proceedings of the IEEE/CVF Conference on Computer Vision and Pattern Recognition},
  pages     = {770--778},
  year      = {2016},
}

@article{krizhevsky2009CIFAR,
  title={Learning multiple layers of features from tiny images},
  author={Krizhevsky, Alex and Hinton, Geoffrey and others},
  journal={Technical report},
  year={2009},
  publisher={Toronto, ON, Canada}
}

@inproceedings{Xiao2015Clothing,
  title={Learning from massive noisy labeled data for image classification},
  author={Xiao, Tong and Xia, Tian and Yang, Yi and Huang, Chang and Wang, Xiaogang},
  booktitle={Proceedings of the IEEE/CVF Conference on Computer Vision and Pattern Recognition},
  pages={2691--2699},
  year={2015}
}

@article{bose2022controllable,
  title={Controllable generative modeling via causal reasoning},
  author={Bose, Joey and Monti, Ricardo Pio and Grover, Aditya},
  journal={Transactions on Machine Learning Research},
  year={2022}
}

@inproceedings{Stiennon20HumanFeedback,
 title = {Learning to summarize with human feedback},
 author = {Stiennon, Nisan and Ouyang, Long and Wu, Jeffrey and Ziegler, Daniel and Lowe, Ryan and Voss, Chelsea and Radford, Alec and Amodei, Dario and Christiano, Paul F},
 booktitle = {Proceedings of the Advances in Neural Information Processing Systems},
 pages = {3008--3021},
 year = {2020}
}

@inproceedings{chen2023understanding,
  title={Understanding and Mitigating the Label Noise in Pre-training on Downstream Tasks},
  author={Chen, Hao and Wang, Jindong and Shah, Ankit and Tao, Ran and Wei, Hongxin and Xie, Xing and Sugiyama, Masashi and Raj, Bhiksha},
  booktitle={Proceedings of the International Conference on Learning Representations},
  year={2024}
}

@article{li2017webvision,
  title={Webvision database: Visual learning and understanding from web data},
  author={Li, Wen and Wang, Limin and Li, Wei and Agustsson, Eirikur and Van Gool, Luc},
  journal={arXiv preprint arXiv:1708.02862},
  year={2017}
}

@article{zhang2017improving,
  title={Improving crowdsourced label quality using noise correction},
  author={Zhang, Jing and Sheng, Victor S and Li, Tao and Wu, Xindong},
  journal={IEEE transactions on neural networks and learning systems},
  volume={29},
  number={5},
  pages={1675--1688},
  year={2017},
  publisher={IEEE}
}

@article{natarajan2013learning,
  title={Learning with noisy labels},
  author={Natarajan, Nagarajan and Dhillon, Inderjit S and Ravikumar, Pradeep K and Tewari, Ambuj},
  journal={Proceedings of the Advances in Neural Information Processing Systems},
  volume={26},
  year={2013}
}

@article{gu2023instance,
  title={An instance-dependent simulation framework for learning with label noise},
  author={Gu, Keren and Masotto, Xander and Bachani, Vandana and Lakshminarayanan, Balaji and Nikodem, Jack and Yin, Dong},
  journal={Machine Learning},
  volume={112},
  number={6},
  pages={1871--1896},
  year={2023},
  publisher={Springer}
}

@inproceedings{yao2023better,
  title={Which is better for learning with noisy labels: the semi-supervised method or modeling label noise?},
  author={Yao, Yu and Gong, Mingming and Du, Yuxuan and Yu, Jun and Han, Bo and Zhang, Kun and Liu, Tongliang},
  booktitle={International Conference on Machine Learning},
  pages={39660--39673},
  year={2023},
  organization={PMLR}
}

@article{xia2020part,
  title={Part-dependent label noise: Towards instance-dependent label noise},
  author={Xia, Xiaobo and Liu, Tongliang and Han, Bo and Wang, Nannan and Gong, Mingming and Liu, Haifeng and Niu, Gang and Tao, Dacheng and Sugiyama, Masashi},
  journal={Proceedings of the Advances in Neural Information Processing Systems},
  volume={33},
  pages={7597--7610},
  year={2020}
}

@article{menon2018learning,
  title={Learning from binary labels with instance-dependent noise},
  author={Menon, Aditya Krishna and Van Rooyen, Brendan and Natarajan, Nagarajan},
  journal={Machine Learning},
  volume={107},
  pages={1561--1595},
  year={2018},
  publisher={Springer}
}

@inproceedings{bai2023subclass,
  title={Subclass-Dominant Label Noise: A Counterexample for the Success of Early Stopping},
  author={Bai, Yingbin and Han, Zhongyi and Yang, Erkun and Yu, Jun and Han, Bo and Wang, Dadong and Liu, Tongliang},
  booktitle={Thirty-seventh Conference on Neural Information Processing Systems},
  year={2023}
}

@inproceedings{yu2019does,
  title={How does disagreement help generalization against label corruption?},
  author={Yu, Xingrui and Han, Bo and Yao, Jiangchao and Niu, Gang and Tsang, Ivor and Sugiyama, Masashi},
  booktitle={International Conference on Machine Learning},
  pages={7164--7173},
  year={2019},
  organization={PMLR}
}

@inproceedings{wang2019co,
  title={Co-mining: Deep face recognition with noisy labels},
  author={Wang, Xiaobo and Wang, Shuo and Wang, Jun and Shi, Hailin and Mei, Tao},
  booktitle={Proceedings of the IEEE/CVF international conference on computer vision},
  pages={9358--9367},
  year={2019}
}

@inproceedings{scott2013classification,
  title={Classification with asymmetric label noise: Consistency and maximal denoising},
  author={Scott, Clayton and Blanchard, Gilles and Handy, Gregory},
  booktitle={Conference on learning theory},
  pages={489--511},
  year={2013},
  organization={PMLR}
}

@inproceedings{goldberger2016training,
  title={Training deep neural-networks using a noise adaptation layer},
  author={Goldberger, Jacob and Ben-Reuven, Ehud},
  booktitle={Proceedings of the International Conference on Learning Representations},
  year={2016}
}

@inproceedings{scott2015rate,
  title={A rate of convergence for mixture proportion estimation, with application to learning from noisy labels},
  author={Scott, Clayton},
  booktitle={Artificial Intelligence and Statistics},
  pages={838--846},
  year={2015},
  organization={PMLR}
}

@article{liu2015classification,
  title={Classification with noisy labels by importance reweighting},
  author={Liu, Tongliang and Tao, Dacheng},
  journal={IEEE Transactions on Pattern Analysis and Machine Intelligence},
  volume={38},
  number={3},
  pages={447--461},
  year={2015},
  publisher={IEEE}
}

@inproceedings{reed2014training,
  title={Training deep neural networks on noisy labels with bootstrapping},
  author={Reed, Scott and Lee, Honglak and Anguelov, Dragomir and Szegedy, Christian and Erhan, Dumitru and Rabinovich, Andrew},
  booktitle={Proceedings of the International Conference on Learning Representations},
  year={2015}
}

@incollection{yao2023causality,
  title={Causality Encourages the Identifiability of Instance-Dependent Label Noise},
  author={Yao, Yu and Liu, Tongliang and Gong, Mingming and Han, Bo and Niu, Gang and Zhang, Kun},
  booktitle={Machine Learning for Causal Inference},
  pages={247--264},
  year={2023},
  publisher={Springer}
}

@inproceedings{liu2023identifiability,
  title={Identifiability of label noise transition matrix},
  author={Liu, Yang and Cheng, Hao and Zhang, Kun},
  booktitle={International Conference on Machine Learning},
  pages={21475--21496},
  year={2023},
  organization={PMLR}
}

@inproceedings{cheng2020learning,
  title={Learning with bounded instance and label-dependent label noise},
  author={Cheng, Jiacheng and Liu, Tongliang and Ramamohanarao, Kotagiri and Tao, Dacheng},
  booktitle={International conference on machine learning},
  pages={1789--1799},
  year={2020},
  organization={PMLR}
}

@article{neuberg2003causality,
  title={Causality: models, reasoning, and inference, by judea pearl, cambridge university press, 2000},
  author={Neuberg, Leland Gerson},
  journal={Econometric Theory},
  volume={19},
  number={4},
  pages={675--685},
  year={2003},
  publisher={cambridge university press}
}

@book{peters2017elements,
  title={Elements of causal inference: foundations and learning algorithms},
  author={Peters, Jonas and Janzing, Dominik and Sch{\"o}lkopf, Bernhard},
  year={2017},
  publisher={The MIT Press}
}

@article{yao2021instance,
  title={Instance-dependent label-noise learning under a structural causal model},
  author={Yao, Yu and Liu, Tongliang and Gong, Mingming and Han, Bo and Niu, Gang and Zhang, Kun},
  journal={Proceedings of the Advances in Neural Information Processing Systems},
  volume={34},
  pages={4409--4420},
  year={2021}
}

@inproceedings{yu2018efficient,
  title={An efficient and provable approach for mixture proportion estimation using linear independence assumption},
  author={Yu, Xiyu and Liu, Tongliang and Gong, Mingming and Batmanghelich, Kayhan and Tao, Dacheng},
  booktitle={Proceedings of the IEEE Conference on Computer Vision and Pattern Recognition},
  pages={4480--4489},
  year={2018}
}

@misc{ultra_fast_controlnet,
  author = {Paul, Sayak and Xu, Yiyi and Platen, Patrick von},
  title = {Ultra fast ControlNet with Diffusers},
  year = {2023},
  publisher = {Hugging Face}
}

@inproceedings{zavadski2023controlnet,
  author    = {Zavadski, Denis and Feiden, Johann-Friedrich and Rother, Carsten},
  title     = {ControlNet-XS: Rethinking the Control of Text-to-Image Diffusion Models as Feedback-Control Systems},
  booktitle = {Proceedings of the European Conference on Computer Vision},
  year      = {2024}
}

@article{covid2022,
	author = {Mirza, Muhammad Waqar and Siddiq, Asif and Khan, Ishtiaq Rasool},
	date = {2023/06/01},
	doi = {10.1007/s11760-022-02214-2},
	id = {Mirza2023},
	isbn = {1863-1711},
	journal = {Signal, Image and Video Processing},
	number = {4},
	pages = {915--924},
	title = {A comparative study of medical image enhancement algorithms and quality assessment metrics on COVID-19 CT images},
	volume = {17},
	year = {2023}
}

@inproceedings{miyato2018spectralnormalizationgenerativeadversarial,
  author    = {Miyato, Takeru and Kataoka, Toshiki and Koyama, Masanori and Yoshida, Yuichi},
  title     = {Spectral Normalization for Generative Adversarial Networks},
  booktitle = {International Conference on Learning Representations},
  year      = {2018}
}

@article{gulrajani2017improved,
  title={Improved training of wasserstein gans},
  author={Gulrajani, Ishaan and Ahmed, Faruk and Arjovsky, Martin and Dumoulin, Vincent and Courville, Aaron C},
  journal={Proceedings of the Advances in Neural Information Processing Systems},
  volume={30},
  year={2017}
}

@inproceedings{Rifai2011ContractiveAE,
  title={Contractive Auto-Encoders: Explicit Invariance During Feature Extraction},
  author={Salah Rifai and Pascal Vincent and Xavier Muller and Xavier Glorot and Yoshua Bengio},
  booktitle={International Conference on Machine Learning},
  year={2011}
}

    \begin{biography}{figures/kevin}
        \textbf{Yuxiang Zheng} is a Ph.D. student at the University of Technology Sydney, working with Dr. Marian-Andrei Rizoiu in the Behavioral Data Science Lab. He received his Bachelor’s degree from the University of Sydney, where he was awarded a University Medal. His research interests include information diffusion, social data science, robust machine learning, and artificial intelligence.
    \end{biography}

    \begin{biography}{figures/zhongyi}
        \textbf{Zhongyi Han} is a Professor and Ph.D. supervisor at the School of Software, Shandong University, China.
        He received his Ph.D. degree from Shandong University in 2023.
        He had been a Postdoctoral Fellow at King Abdullah University of Science and Technology (KAUST) in 2024 and at MBZUAI in 2023, and a visiting student at the University of Western Ontario.
        His research interests include robust machine learning, trustworthy AI, and AI for Science.
    \end{biography}

    \begin{biography}{figures/yilong}
        \textbf{Yilong Yin} is a Professor and Ph.D. supervisor at the School of Software, Shandong University, China, where he also serves as the Director of the Artificial Intelligence Research Center. He received his Ph.D. degree from Jilin University, Changchun, China, in 2000, and was a Postdoctoral Fellow at Nanjing University from 2000 to 2002. His research interests include machine learning, data mining, and artificial intelligence. He has published over 100 papers in leading international journals and conferences.
    \end{biography}

\end{document}